\documentclass[runningheads]{llncs}
\usepackage{graphicx}
\usepackage{hyperref}
\usepackage{caption}
\usepackage{subcaption}
\usepackage{float}
\usepackage[section]{placeins}
\usepackage{amsfonts}
\usepackage{amsmath}
\begin{document}
\title{Scalable Concept Extraction in Industry 4.0}
%
%
\author{
Andrés Felipe Posada-Moreno\inst{1}\orcidID{0000-0003-3751-0680} \and
Kai Müller\inst{2} \and
Florian Brillowski\inst{2}\orcidID{0000-0003-4805-9573} \and
Friedrich Solowjow\inst{1}\orcidID{0000-0003-2623-5652} \and
Thomas Gries\inst{2}\orcidID{0000-0002-2480-8333} \and
Sebastian Trimpe\inst{1}\orcidID{0000-0002-2785-2487}
}
\authorrunning{A. Posada-Moreno et al.}
%
\institute{
Institute for Data Science in Mechanical Engineering, RWTH Aachen University, Dennewartstr. 27, 52068 Aachen, Germany \\
\email{\{andres.posada,solowjow,trimpe\}@dsme.rwth-aachen.de} \and
Institut für Textiltechnik of RWTH Aachen University, Otto-Blumenthal-Straße 1, 52074 Aachen, Germany \\
\email{\{kai.mueller,florian.brillowski,thomas.gries\}@ita.rwth-aachen.de}
}
\maketitle              
%
\begin{abstract}
The industry 4.0 is leveraging digital technologies and machine learning techniques to connect and optimize manufacturing processes.
Central to this idea is the ability to transform raw data into human understandable knowledge for reliable data-driven decision-making.
Convolutional Neural Networks (CNNs) have been instrumental in processing image data, yet, their ``black box'' nature complicates the understanding of their prediction process.
In this context, recent advances in the field of eXplainable Artificial Intelligence (XAI) have proposed the extraction and localization of concepts, or which visual cues intervene on the prediction process of CNNs.
This paper tackles the application of concept extraction (CE) methods to industry 4.0 scenarios.
To this end, we modify a recently developed  technique, ``Extracting Concepts with Local Aggregated Descriptors'' (ECLAD), improving its scalability.
Specifically, we propose a novel procedure for calculating concept importance, utilizing a wrapper function designed for CNNs. This process is aimed at decreasing the number of times each image needs to be evaluated.
Subsequently, we demonstrate the potential of CE methods, by applying them in three industrial use cases.
We selected three representative use cases in the context of quality control for material design (tailored textiles), manufacturing (carbon fiber reinforcement), and maintenance (photovoltaic module inspection).
In these examples, CE was able to successfully extract and locate concepts directly related to each task.
This is, the visual cues related to each concept, coincided with what human experts would use to perform the task themselves, even when the visual cues were entangled between multiple classes.
Through empirical results, we show that CE can be applied for understanding CNNs in an industrial context, giving useful insights that can relate to domain knowledge.

\end{abstract}
%
%
%
\section{Introduction} \label{sec:Introduction}

Industry 4.0 represents a change in how industries operate through the integration of digital technologies, interconnectivity, and artificial intelligence.
This paradigm shift is made possible through three key factors. 
First, the maturity of sensor and communication technology has facilitated seamless monitoring of machines and communication through integrated systems, evolving into the concept of the \textit{Internet of Things} (IoT) \cite{DBLP:journals/candie/WangLWT21}.
Second, the increase in available computational resources has enabled pervasive and complex data processing in real-time decision-making.
Third, the advancements of big data and artificial intelligence technologies allows industries to extract valuable insights from large amounts of complex data \cite{DBLP:journals/ijinfoman/DuanED19}.
Together, these factors are the foundations of the Industry 4.0, where data, interconnected processes, and information processing algorithms are the cornerstone of decision-making in a more efficient, productive, and sustainable future.


In this context, convolutional neural networks (CNNs) have been instrumental for making sense out of image data. CNNs have found numerous applications due to their ability to automatically learn relevant features and patterns from data, enabling applications in visual inspection, quality control, predictive maintenance, and robot vision.
However, one of the challenges of implementing CNNs in industrial applications is their ``black box'' nature, referring to the lack of understanding how their prediction process works.
This issue is highlighted when CNNs behave unexpectedly, due to issues such biases, shortcut learning, and data leakages \cite{DBLP:journals/natmi/GeirhosJMZBBW20}.
This situation underscores the challenge of automatically generating valuable insights from complex data while ensuring the reasons behind them align with expert knowledge and comply with industry standards. 

As a practical example, let us consider a textile quality control process, where a CNN is trained to classify images into normal textiles, folded textiles, or textiles with gaps \cite{Brillowski:848836}.
This quality control step can be part of the product lifecycle of a carbon-fiber process, where the folds or gaps in the used textiles can indicate a detriment in the mechanical properties of a final product.
During operations, the model will classify whether new textile parts contain gaps, folds, or they are normal textiles.
Nonetheless, biases can exist in the dataset, such as different lighting conditions for the classes, specific markers, or even spurious correlations.
Thus, the model can learn to classify the samples based on the unintended bias, as it's an effective mechanism learned during training.
Eventually, it will cause unexpected behaviors, generating issues in the production process, and lead to a general loss of trust in the CNN models.

To address this issue, the field of eXplainable Artificial Intelligence (XAI) focuses on developing techniques to increase the interpretability and understandability of model's prediction processes \cite{DBLP:journals/corr/abs-2006-11371,DBLP:journals/inffus/ArrietaRSBTBGGM20,DBLP:journals/jair/BurkartH21}.
Depending on the context, these techniques can be classified as local or global. 
On one hand, local explanations focus on understanding the reasoning behind a single image prediction, e.g., by computing the contribution of each feature for the current prediction. This type of explanations can be used for detecting critical or non-important factors, providing insights on a single prediction \cite{DBLP:journals/corr/abs-2006-11371,DBLP:journals/inffus/ArrietaRSBTBGGM20,DBLP:journals/jair/BurkartH21}.
In the context of CNNs, these types of explanations are known as saliency maps, but have been found to be noisy and unreliable \cite{DBLP:journals/corr/abs-2110-14297,DBLP:conf/nips/AdebayoGMGHK18}.
On the other hand, global explanations aim to provide a general understanding of how a model functions across multiple datapoints, e.g., by automatically extracting the patterns learned by a model, providing a set of visual cues that a model has learned to differentiate, and how important they are in the prediction process of the model. In the case of CNNs, \textit{concept extraction} (CE) techniques provide sets of patterns learned by CNNs named \textit{concepts}, which refer to visual cues that the models have learned to differentiate in their prediction process \cite{DBLP:conf/icml/KimWGCWVS18,DBLP:conf/nips/GhorbaniWZK19,DBLP:journals/corr/abs-2206-04531}.
These types of explanations have been used for the early detection of biases, and for ensuring the alignment of expert knowledge. Yet, their computational requirements can be an obstacle for scaling their usage to larger industrial datasets.
Overall, the usage of local and global explanations can fulfill a critical role in making CNNs more transparent, understandable, and useful in industrial applications. 

The core process of CE aims at identifying a set of high-level features or representations learned by a CNN during training. It is based on the assumption that through the layers of a CNN, the network learns to represent increasingly complex and abstract features which are functional to the task being learned.
In this process, each high-level feature or concept is represented by three components. First, a vector in the latent space of the CNN, used to determine if a concept is present in an image (and sometimes where). Second, an importance score, describing how relevant is the concept for the prediction process of the network. Third, a set of example images containing the concept, which serves as a human understandable medium for making sense of which visual cues are associated with the concept.
These methods aim to understand and visualize these learned representations, to gain insights in the functioning of the CNN, and to assess how aligned they are compared to experts knowledge.

Current research on applications of XAI methods for industrial applications has focused on using local explanation techniques, such as SHAP \cite{DBLP:conf/nips/LundbergL17} to explain timeseries or tabular data based models.
In contrast, global explanation techniques have been primarily used in the medical domain to search for global biases and artifacts \cite{DBLP:journals/corr/abs-1904-04520,gamble2021determining}.
However, global explanation techniques have rarely been applied in the context of the industry 4.0. This represents a relevant gap because global explanations can provide a general overview of what a model learns, allowing a clear comparison with domain knowledge, this makes them relevant for the industry 4.0. 
Furthermore, how such explanations relate in the context of a digital shadow or digital twin has not been discussed.

In this work, we extend a recent CE method for better scalability and apply it on three representative datasets.
We select datasets in the context of material design, product manufacturing and maintenance.
First, we select a dataset related to material production, this is, a dataset on quality control during textile/weave production of 2D semifinished composites. 
Second, we select a dataset one step further in a product manufacturing process, this is, a dataset on carbon fiber three-dimensional form reinforcement.
Finally, we choose a dataset on visual inspection during maintenance, this is, a dataset on photovoltaic modules defect detection.  
In addition to demonstrating how CE helps for the validation of the trained models/CNNs, we also discuss how the concepts can be an integral part of model development and quality control.
There are three key contributions in this work. 
First, we modify the concept extraction technique \textit{Extracting Concepts with Local Aggregated Descriptors} (ECLAD) to reduce its computational requirements by reformulating its concept extraction step.
Second, we provide three use cases of industrial applications where concept-based explanations can be used, providing empirical results significant in the context of the industry 4.0.
Third, we discuss the relation between global explanations with the notion of digital shadows. We highlight how the usage of global explanations can relate to expert knowledge, and how this allows a more informative usage in operations. 
\section{Background} \label{sec:Background}

Our work explores the intersection of two topics within the realm of artificial intelligence. 
The first topic focuses on the field of explainable artificial intelligence, where we further develop a type of concept-based explanation technique based on local aggregated descriptors.
The second topic relates to the applications of XAI techniques in an industrial context, specifically, exploring the connection between explanations, industry 4.0, and knowledge generation from data.

In this section we briefly explore the current literature of both topics, providing an overview of the state-of-the-art research related to both concept-based explanations, and XAI techniques in the industry 4.0.

\subsection{Concept-Based Explanations} \label{subsec:concept-based explanations}

In the field of XAI, \textbf{post-hoc explanations} refer to methods used to provide insights into the inner workings of a model after it has been trained, and without modifying its architecture or training process.
Post-hoc explanations are generally divided in two categories, local explanations (concerning a single data point), or in global explanations (concerning the general behavior of the model) \cite{DBLP:journals/corr/abs-2006-11371,DBLP:journals/inffus/ArrietaRSBTBGGM20,DBLP:journals/jair/BurkartH21}.
Concerning of CNNs, local explanations are known as saliency maps. These explanations are able to increase the transparency of a single prediction, but are unable of providing any direct insight regarding the future behavior of a model, nor it's generalization capabilities to other datapoints.
In contrast, global explanations, and specifically concept extraction techniques, provide a series of references of how a model encodes information, which types of visual cues it learned to detect, and how important are these patterns in the prediction process.
In this context, we have based our work in concept-based explanations, as these provide high-level abstractions, serving as bits of generated knowledge which human experts can assess and confirm.

As a generality, \textbf{concept extraction} refers to techniques aiming to identify and represent in human understandable visualizations, high level abstraction within the latent space of a CNN.
This family of post-hoc techniques, have been developed following the work on``Testing with Concept Activation Vectors'' (TCAV) \cite{DBLP:conf/icml/KimWGCWVS18}, where Kim et al. propose a method for testing whether a human defined concept was learned by a CNN by statistically testing latent representations (flattened activation maps) of the concept's examples.

A representative technique is ``\textbf{Automatic Concept-based Explanations}'' (ACE) \cite{DBLP:conf/nips/GhorbaniWZK19}, complimenting TCAV by automatically extracting concepts. The base mechanism of ACE consisted on extracting and clustering representations of multiple patches of each image, obtained using a superpixel segmentation technique.
In contrast, \textbf{Concept-Shap} \cite{DBLP:conf/nips/YehKALPR20} proposes the learning of a concept representation, by inserting a bottleneck layer of lower dimensions between two layers of a CNN. This method also suggested the usage of Shapley values \cite{DBLP:conf/nips/LundbergL17} instead of TCAV for scoring the importance of each concept. Other similar techniques such as PACE \cite{DBLP:conf/ijcnn/KamakshiGK21}, MACE \cite{DBLP:journals/tai/KumarSGKK21} and CACE \cite{DBLP:journals/corr/abs-1907-07165} have proposed similar approaches of training an encoder towards a lower dimensional space. In comparison, these techniques are more computationally expensive, as the learning of this representation space requires training through multiple epochs of the complete analyzed dataset.
In comparison, ``\textbf{Extracting Concepts with Local Aggregated Descriptors}'' (ECLAD) \cite{DBLP:journals/corr/abs-2206-04531}, proposed the identification of concepts through pixel-wise descriptors obtained from the latent space of models named ``Local Aggregated Descriptors''. This technique introduced the possibility of not only extracting concepts, but also localize them in new sample images.

Nonetheless, the current techniques are computationally expensive, requiring a significant number of evaluations of the CNN for each image in a dataset, as indicated in the complexity analysis by Posada-Moreno et al. \cite{DBLP:journals/corr/abs-2206-04531}. In the case of ACE, this number scales proportional to the number of concepts and superpixels (20-50). For Concept-Shap, MACE, and PACE, this number scales with the number of epochs required for learning the concept representation. As well as the number of samples used to estimate the concepts importance, e.g., Concept-Shap uses Shapley values (Monte Carlo approximation \cite{DBLP:journals/kais/StrumbeljK14}), depending also on the number of concepts. Similarly, for ECLAD, these evaluations scale with the number of classes of the dataset and the number of concepts (to compute their relative importance measure).
In our work, we aim to conserve the benefit of concept extraction and localization of ECLAD, while diminishing its required computational cost. We achieve this replacing the importance measurements of the concepts, requiring only one evaluation per image during concept identification, and one evaluation for concept scoring.

\subsection{Industrial Applications of XAI Techniques} \label{subsec:Industrial Applications of XAI Techniques}

On a practical side, most industrial applications of XAI techniques center around local explanations, which provide insights into individual predictions made by AI models. These focus has been observed in previous works \cite{saranya2023systematic,DBLP:journals/tii/AhmedJP22,islam2022systematic,DBLP:journals/access/ZhangHDYT22}, where a significant ration of the cited examples on each survey relate to local explanation techniques such as SHAP \cite{DBLP:conf/nips/LundbergL17}, LIME \cite{DBLP:conf/kdd/Ribeiro0G16}, and Grad-CAM \cite{DBLP:conf/iccv/SelvarajuCDVPB17}. To our knowledge, there is a gap in the literature on the usage of global explanations in the context of industrial applications.

The primary applications of XAI techniques relate to the understanding and validation of AI models though their development process. More specifically, current works have focused on the interpretability and transparency of these models as a mean to debug models and detect biases. Previous works have explored the usage of XAI techniques in the context of quality control \cite{DBLP:journals/mansci/SenonerNF22,Brillowski:848836}, anomaly detection \cite{DBLP:journals/corr/abs-2102-11848,chowdhury2022xai,DBLP:journals/sensors/MeasMKTLLPB22}, or maintenance \cite{DBLP:conf/ickii/HongLLKH20,DBLP:conf/icmla/MouchawehR22,DBLP:conf/fuzzIEEE/SerradillaZCAOZ20}. These works have focused mainly in analyzing models based on tabular data or timeseries. In our work, we focus not only on using XAI methods as a tool for testing and detecting biases, but also on the capability of generating knowledge, validating it with human experts and reusing said knowledge to provide valuable insights during operations.

A subset of previous publications has tackled the usage of XAI methods for analyzing models based on image data in industrial applications.
These studies employ XAI methods to understand the main features used by models for making predictions, with the goal of enhancing interpretability and detecting possible biases.
Notably, Brillowski et al. \cite{Brillowski:848836} explored the usage of Grad-CAM \cite{DBLP:conf/iccv/SelvarajuCDVPB17} to study a CNNs decision logic, focusing on enhancing interpretability, and detecting possible misalignment with human experts.
Similarly, Ho Sun et al. \cite{DBLP:journals/access/SunHTLJL20} tackle the usage of CAM \cite{DBLP:conf/cvpr/ZhouKLOT16} methods, for enhancing vibration video based fault diagnostics.
Nonetheless, these applications focus on local explanations, which limits significantly the scope of the insights, exposing any hypothesis to possible confirmation biases.
These approaches seek to enhance predictions, but leave aside any knowledge that can be abstracted on a global scale, such as the identification of the visual cues learned by the model.
In our work, we seek not only to enhance the predictions of the original model by extending them through explanations, but also to profit from the validation process of the model to generate coherent knowledge about the visual cues related to experts knowledge.

Researchers have designed multiple frameworks on the concept of industry 4.0 \cite{becker2021conceptual}. Specially, the term Digital Twin (DT) refers to the models related to physical entities, and the term Digital shadow refers to the data trace generated by the sensors \cite{DBLP:conf/caise/BibowDHMRSSW20}. In our work, we propose including explanation methods not only as part of these entities, but also as tools for the direct generation of knowledge from data. This is, predictive models and explanation methods become part of the digital twins, whereas the predictions and explanations become part of digital shadows. Additionally, the interaction with human experts provides the assurance that concepts generate domain knowledge during operations.

\section{Concept Extraction} \label{sec:Concept Extraction}

The term \textit{concept-extraction} refers to a process of analyzing a trained model $y = f(x)$ and a dataset $E = \{ (x_i, y_i)\}_{i=1}^{N}$ to identify and score a set of high-level features or concepts that the model learned to differentiate during training. As a result, a set of concepts $C = \{ c_j\}_{j=1}^{n_c}$ is extracted, where each concept $c_j$, is described by a latent representation $v_{c_j}$ within the analyzed CNN, an importance score $I_{c_j}$ denoting its relevance towards the prediction process, and a set of examples $\varepsilon_{c_j}$ containing the visual cues related to the concept. The process of concept extraction, highly depends on (i) the choice of latent representation to use, (ii) the process of concept identification, and (iii) the metric used for concept importance scoring. In this section, we will introduce our choices for providing a scalable and effective concept extraction technique.

In this work, we use ECLAD \cite{DBLP:journals/corr/abs-2206-04531} as a method for concept extraction.
We then modify its concept importance scoring phase, to diminish its computational cost. thus, we start by describing the latent representation (Subsection \ref{subsec:latent representation}) and the concept identification (Subsection \ref{subsec:concept identification}) as introduced in ECLAD. Finally, we introduce a novel improved importance score (Subsection \ref{subsec:improved importance score}), which we propose to diminish the computational cost of assessing the relevance of concepts.

\subsection{Latent representations}\label{subsec:latent representation}
\paragraph{The latent representations} in concept extraction are a choice which has direct influence on how the concepts are abstracted, identified, and scored. We use the notion of Local Aggregated Descriptors (LADs) introduced in ECLAD \cite{DBLP:journals/corr/abs-2206-04531}. This descriptor $d_{(a,b)}$ refers to each pixel-wise element of the tensor $d$, which can be obtained by concatenating the upscaled activation maps of a set of arbitrary layers $L$. in contrast with other latent representations (e.g. flatten activation maps for TCAV and ACE \cite{DBLP:conf/icml/KimWGCWVS18,DBLP:conf/nips/GhorbaniWZK19}), LADs provide information on how the CNN encodes a region of the images in multiple levels of abstraction, while maintaining the positional information of each spatial region. As LADs are computed for each pixel in an image, they allow for an intrinsic concept localization, informing users not only about what is in an image, but also where. This advantage over other methods enables the posterior usage of concept vectors to generate local explanations (for a single instance), through concept localization masks.

\subsection{Concept identification}\label{subsec:concept identification}
\paragraph{The concept identification} process, requires the partial or total computation of latent representations for a dataset and the usage of a pattern mining method. In contrast with other types of latent representations, LADs are computed on a per-pixel basis for each image. Thus, the computation of LADs for all images in a dataset would pose a significant memory requirement. To mitigate this requirement, the process of concept identification is performed by computing the LADs of batches of images, and using the clustering algorithm minibatch k-means \cite{DBLP:conf/www/Sculley10} to obtain a set $\Gamma=\{ \gamma_j \}_{j=1}^{n_{c}}$ of $n_{c}$ centroids defining each concept.

In a practical sense, the concept identification step proposed in ECLAD \cite{DBLP:journals/corr/abs-2206-04531}, consists on performing minibatch k-means over the set $D$ of all descriptors $d_{x_i,(a,b)}$ of all images of the dataset $E$,

\begin{equation}
    \Gamma =\{ \gamma_j \}_{j=1}^{n_{c}} = \text{minibatch k-means}(D, n_{c}).
\end{equation}

As a highlight, we can obtain a binary mask $m_{x_i}^{\gamma_j}$ ($m_{x_i}^{c_j}$) by extracting the LADs of the image $x_i$, and comparing which pixel-wise representations are closer to the centroid $\gamma_j$. Similarly, the example set of which visual features in the images correspond to a concept can be computed, by attenuating unrelated regions of the images by a factor $\lambda$ as introduced in ECLAD, 

\begin{equation}
\varepsilon_{c_j} = \{(1-\lambda) \, m_{x_i}^{c_j} \odot x_i + \lambda x_i \mid x_i \in E \}.
\label{eq:example sets}
\end{equation}

Where $\odot$ denotes the element-wise product between matrices.

\subsection{Improved importance score}\label{subsec:improved importance score}
\paragraph{The importance score} of a concept is a numerical value quantifying the relevance of its related visual cues towards the prediction of the analyzed model $f$. This score helps understand which visual cues have the most significant impact on the model's output. Nonetheless, the relative importance score $RI_{c_j}$ proposed in ECLAD, is computationally expensive to compute, requiring $n_c \times n_k$ evaluations of the model for each analyzed image, where $n_c$ denotes the number of concepts, and $n_k$ the number of output classes of the model $f$. 
\textbf{We propose a new importance metric}, which can be computed through a single evaluation of the model $f$ for each analyzed image. To this end, we introduce the function $g(y)$,
\begin{equation} \label{eq:wrapper}
    g(y) = \| y \cdot \boldsymbol{1}^{\top} - \boldsymbol{1} \cdot y^{\top} \|_2. 
\end{equation}
Where $y \in \mathbb{R}^{n_\mathrm{k}}$ is the output of the model $f$, and $\boldsymbol{1}$ is a vector of ones the of the same size as $y$. $g(y)$ reduces the final output of the model $f$, to a single value related to the difference between the logits of each output dimension.

The function $g(y)$ or $g(f(x))$ aims to highlight how distinctive an image is with respect to all classes in a single score. 
In this context the result of $y \cdot \boldsymbol{1}^{\top} - \boldsymbol{1} \cdot y^{\top}$ yields a matrix where each item represents the difference between the logits of two classes. Thus, the elements will be zero in the diagonal, and will represent the difference between classes in the rest of positions. Thus, $\nabla_{x}g(f(x))$ gives a quantitative measure, how how much each pixel in $x$ contributes to differentiating one class from another.

To estimate the relevance of the concept $c_j$ per image, we aggregate the gradient $g(f(x))$ of the pixels related to said concept, as seen in Equation \ref{eq:concept relevance}. We obtain the mean relevance of the concept for all images containing the concept in the dataset, as seen in Equation \ref{eq:mean relevance}.

\begin{equation} \label{eq:concept relevance}
    r_{x_i}^{c_j} = \| \nabla_{x}g(f(x_i)) \odot m_{x_i}^{c_j} \|_{1}.
\end{equation}
\begin{equation} \label{eq:mean relevance}
    \overline{r^{c_j}} = \frac{1}{n_{c_j}} \sum_{x_i \in E}  r_{x_i}^{c_j}  .
\end{equation}
Where  $n_{c_j} = \| \{ \|m_{x_i}^{c_j}\|>0 \mid x_i \in E \} \|$ is the number of images in the dataset $E$ containing pixels related to $c_j$;
And $\odot$ denotes the element-wise product between matrices.
As an intuition, $r_{x_i}^{c_j}$ refers to how much the pixels belonging to the concept $c_j$ contribute towards the difference in logits, in other terms, towards the prediction of the model. Similarly, $\overline{r^{c_j}}$ refers to the average contribution of a concept, along a complete dataset.
Finally, we introduce the importance score $I_{c_j}$ as the scaled absolute value of the mean relevance of each concept,
\begin{equation} \label{eq:mean relevance}
    I_{c_j} = \frac{\lvert \overline{r^{c_j}} \rvert}{ \underset{c_j}{\mathrm{max}}(\lvert \overline{r^{c_j}} \rvert)}.
\end{equation}
After the execution of our method, a set $C = \{ c_j\}_{j=1}^{n_c}$ of $n_c$ concepts are extracted. Each concept is defined by its latent representation, $\gamma_j$ defining the pattern learned by the CNN; The importance score $I_{c_j}$, defining the relevance of the concepts in the prediction process of the model; and the example set $\varepsilon_{c_j}$, containing masked images, showing which visual cues relate to each concept.
Our method retains the extraction and localization capabilities from ECLAD, while reducing significantly its computational. Specifically, we reduce the number of evaluations of the model for computing the importance scores of the concept, from $n_c \times n_k$ per image for ECLAD, to once per image for ours.

\section{Use Cases} \label{sec:Use Cases}

In this section, we present three representative industrial use cases where we apply our concept extraction method.
We first present a case where CE is used in the context of fiberglass textiles, where the analyzed model was trained to detect common material errors within samples during a materials design process.
Then, we explore an application of carbon fiber reinforcement quality control, where models are used during production, as a mean for detecting placement errors which can directly impact the mechanical properties of a resulting product.
Finally, we study a case in the context of maintenance, where we analyze a dataset of photovoltaic modules, depicting an automatic inspection process in large solar plants.
As a highlight, the choice of use cases, relates to three key industrial topics, material design and production, product manufacturing, and maintenance processes for asset management. 

For each use case, we start with a brief introduction, explaining the relevance of the applications. Then, we introduce the dataset used as a representative problem in each context. Research on image classification has advanced significantly over the past decades, allowing the training of high accuracy models without significant issues. We solve the tasks of images classification through a standard Densenet121 \cite{DBLP:conf/cvpr/HuangLMW17} architecture, training each model until convergence.

In the three cases, the classification problem can be solved through the used architecture and training scheme. Yet, for the usage of these models, it is important to understand how the predictions are being made. It is only through the explanation processes that we can ensure whether the models are performing the tasks as intended. Thus, for each use case, we follow the training with the execution of our concept extraction technique, and obtain explanations on how each models' prediction process work. Finally, we discuss each set of results, performing visual inspection and comparing them with human experts knowledge.

\subsection{Tailored Textiles} \label{subsec:Tailored Textiles}

The tailored textiles (TT) use case concerns a quality control problem during the development of custom reinforcement composite materials.
The value of TT arises from the increase of the global fiber-reinforced plastics (FRP) market, which was approximately 8\% in 2021 \cite{Witten.2022}.
In this market, the high cost of producing fiber reinforced polymers (FRPs) has led to efforts to find ways to reduce manufacturing costs while maintaining structural performance.
In these market needs, TT are a promising solution for achieving weight reduction, as they can be locally reinforced with additional material, reducing weight while maintaining strength. 
Reinforcements are incorporated during production, leading to cost savings during preforming. 
Due to their properties, TT have the potential to improve the cost and weight efficiency of FRPs \cite{Uthemann.2017}. Quality controls are necessary to ensure that TTs meet load-path-compliant reinforcement requirements, such as fiber angle, freedom from defects, and proper setting \cite{DINDeutschesInstitutfurNormierunge.V..1999,DINDeutschesInstitutfurNormierunge.V..1987}.
This issue can be tackled through the usage of automatic visual inspection systems, which enable non-destructive and fast inline testing \cite{Duboust.2017,Gholizadeh.2016,Mueller.2022}.

\begin{figure}[!htb]
    \centering
    \begin{subfigure}{\textwidth}
        \centering
        \includegraphics[width=0.24\linewidth]{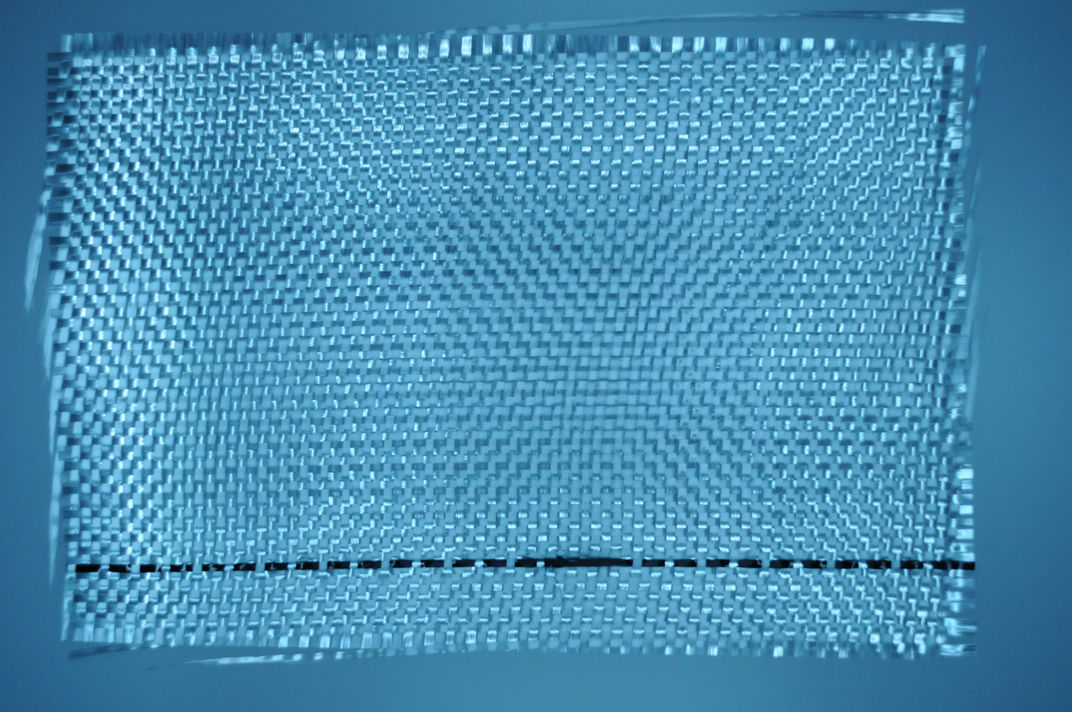}
        \includegraphics[width=0.24\linewidth]{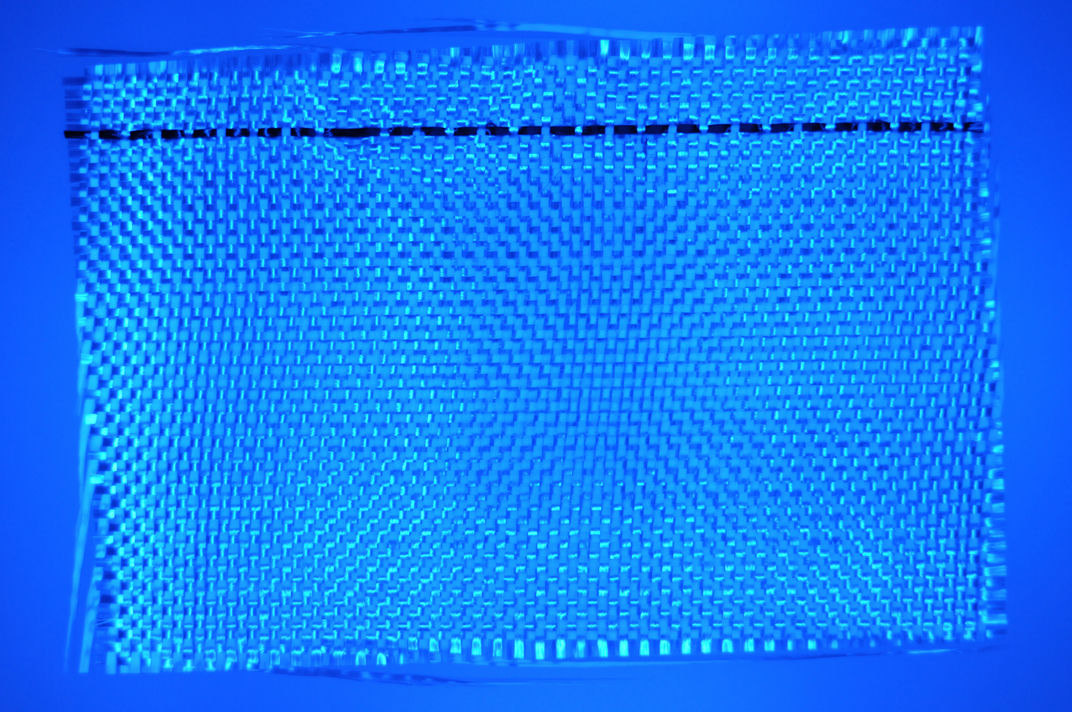}
        \includegraphics[width=0.24\linewidth]{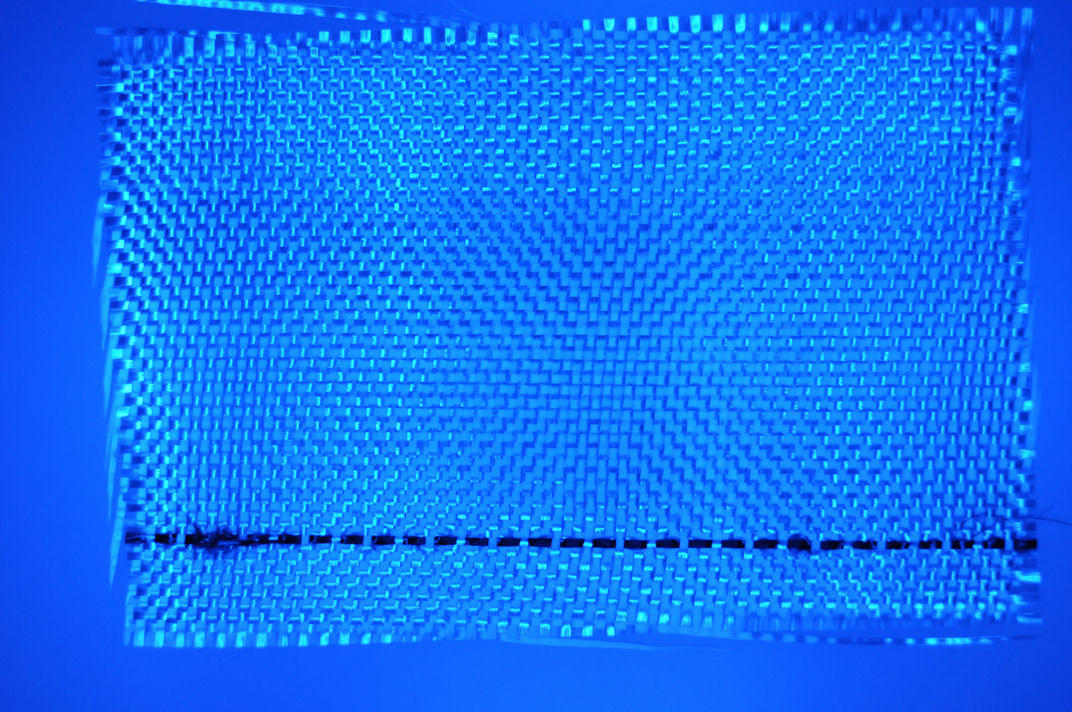}
        \includegraphics[width=0.24\linewidth]{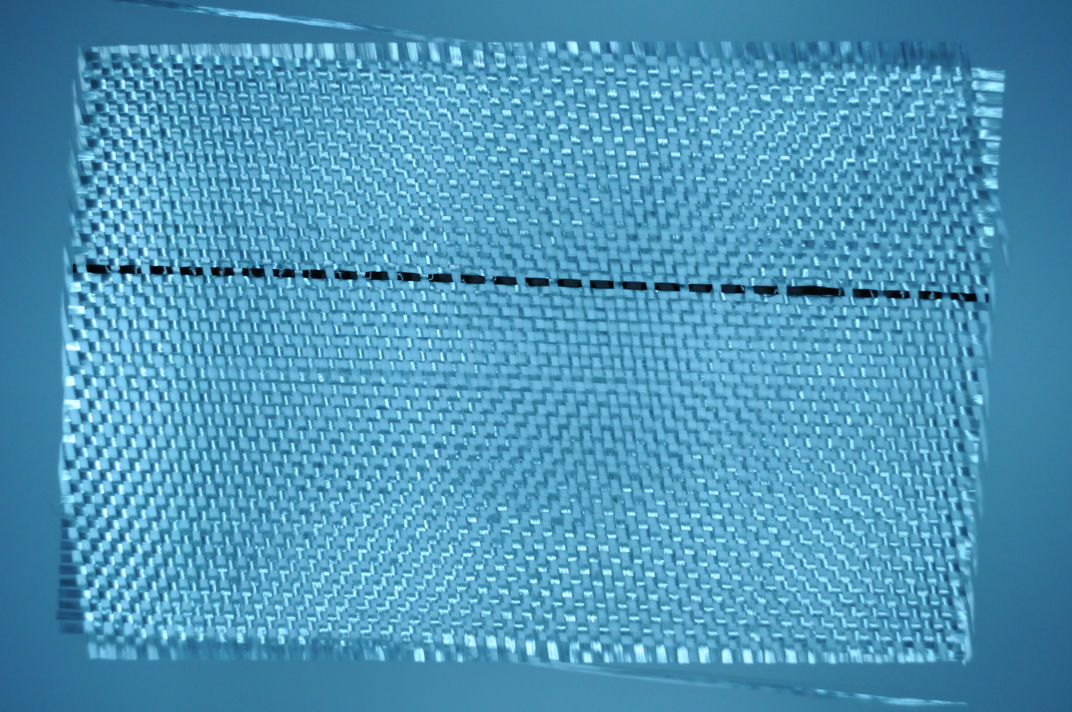}
        \caption{Class 0: ``Binding error''.}
        \label{subfig:TT_0}
    \end{subfigure}

    \begin{subfigure}{\textwidth}
        \centering
        \includegraphics[width=0.24\linewidth]{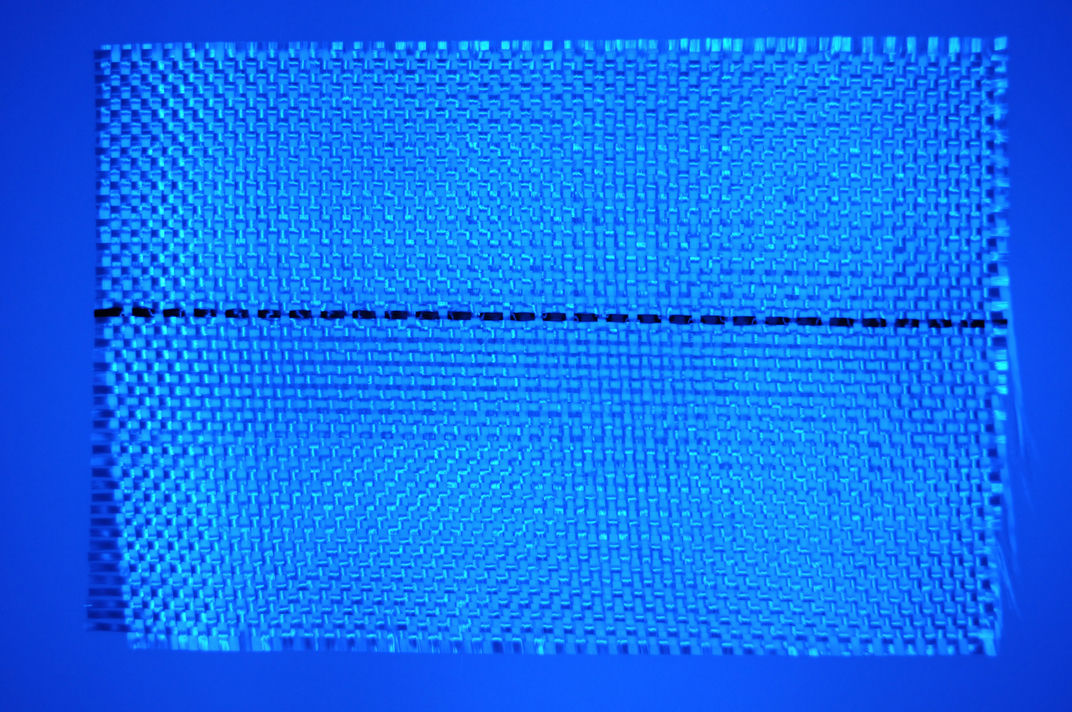}
        \includegraphics[width=0.24\linewidth]{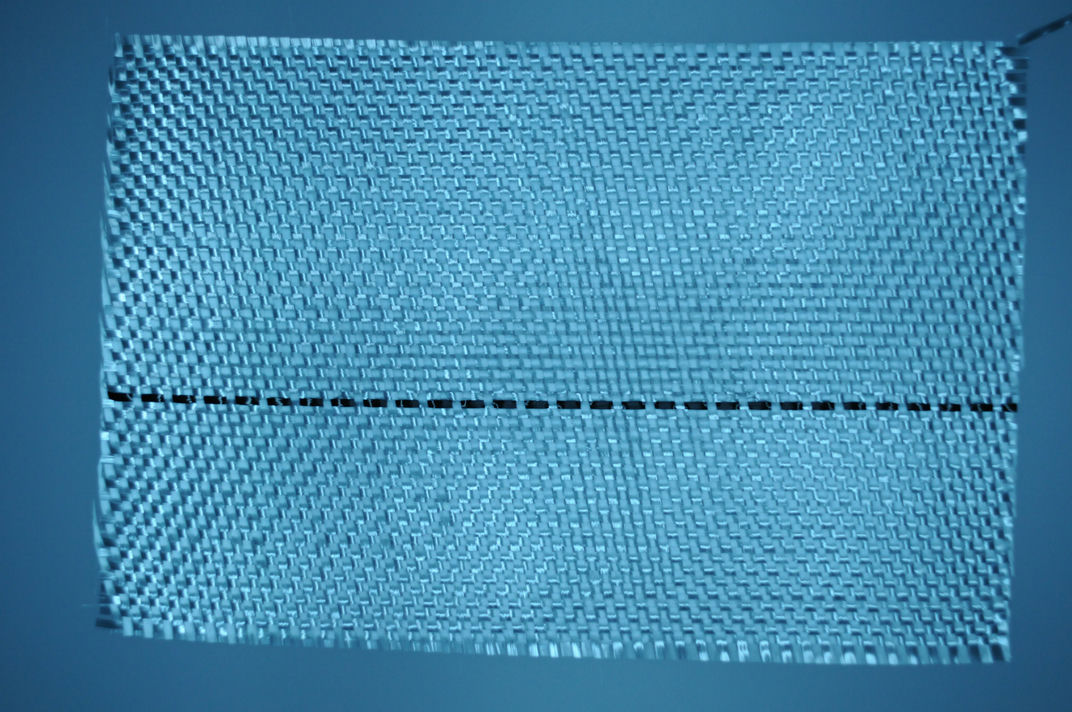}
        \includegraphics[width=0.24\linewidth]{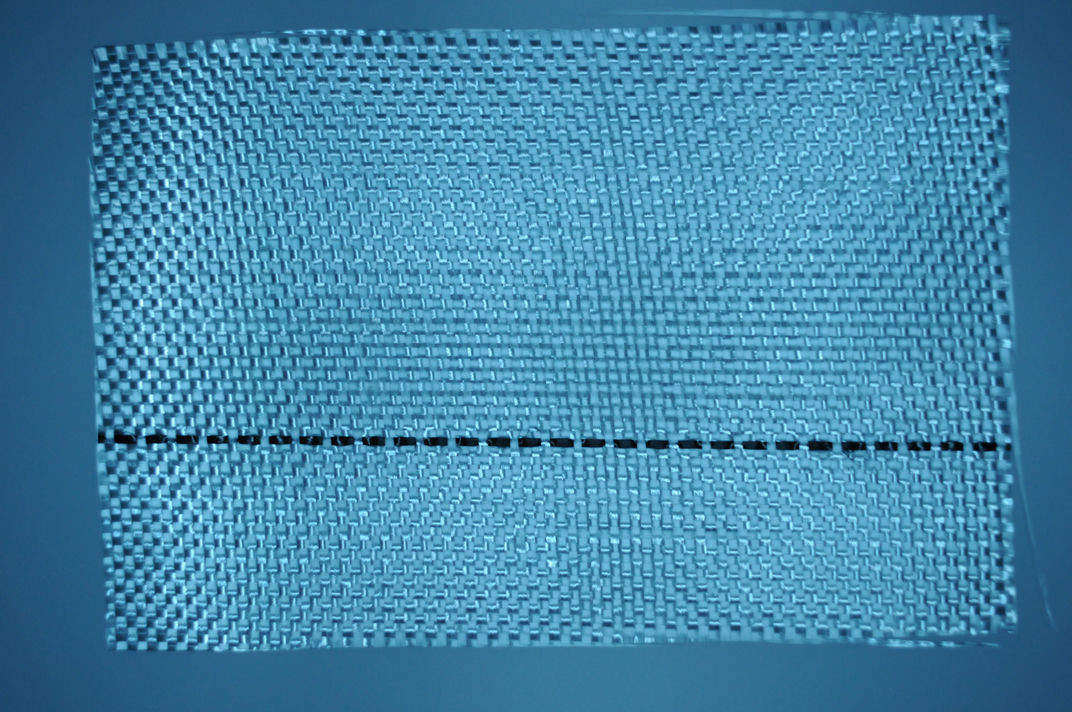}
        \includegraphics[width=0.24\linewidth]{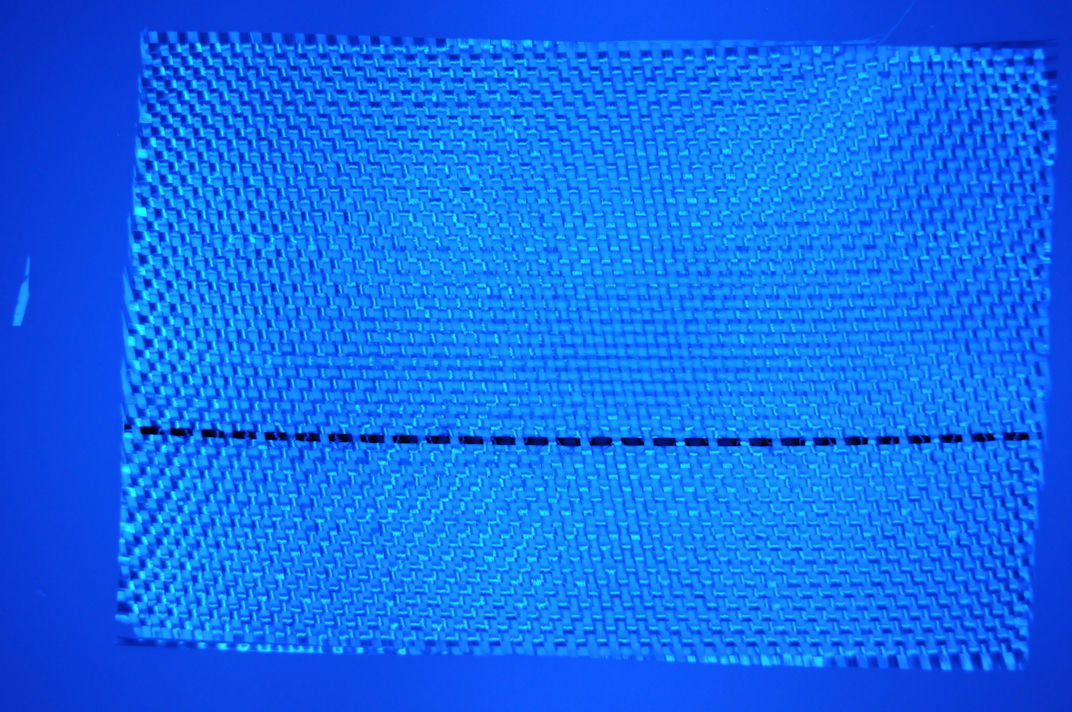}
        \caption{Class 1: ``11 mm binding'', C-fiber reinforced textile tied every 11 mm.}
        \label{subfig:TT_1}
    \end{subfigure}

    \begin{subfigure}{\textwidth}
        \centering
        \includegraphics[width=0.24\linewidth]{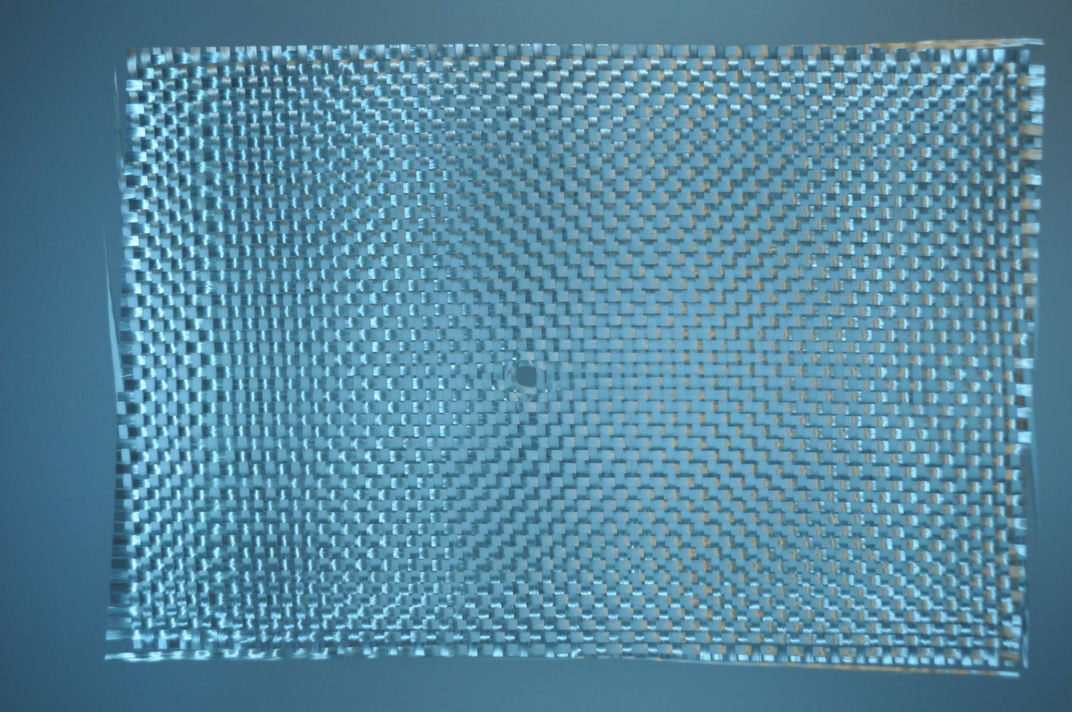}
        \includegraphics[width=0.24\linewidth]{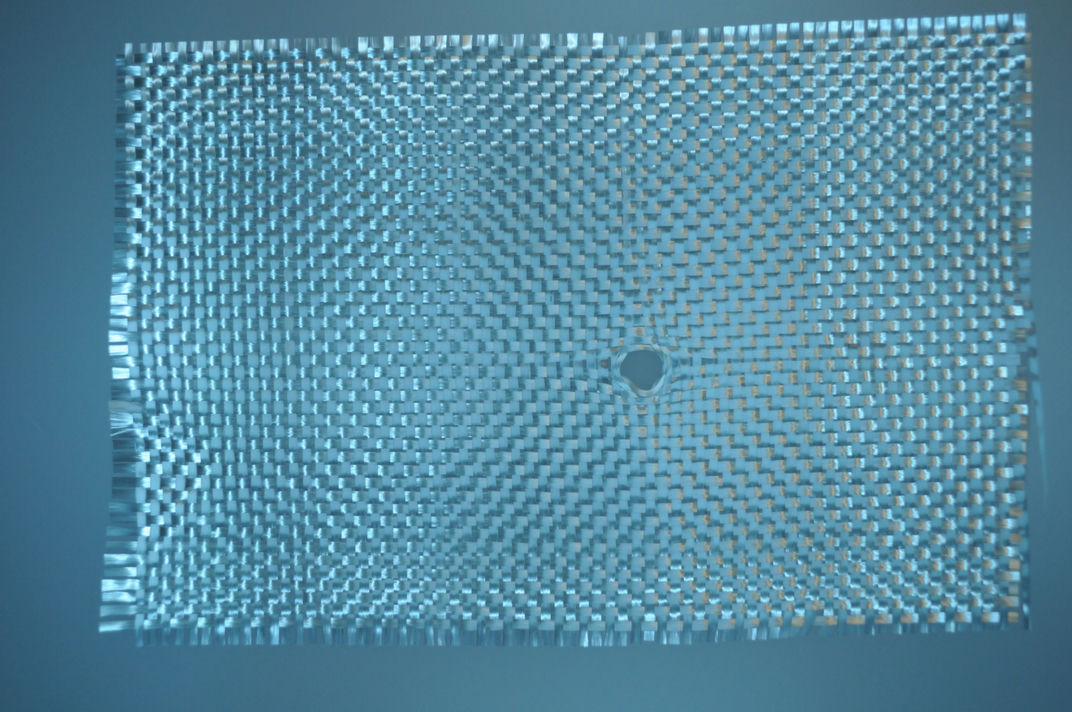}
        \includegraphics[width=0.24\linewidth]{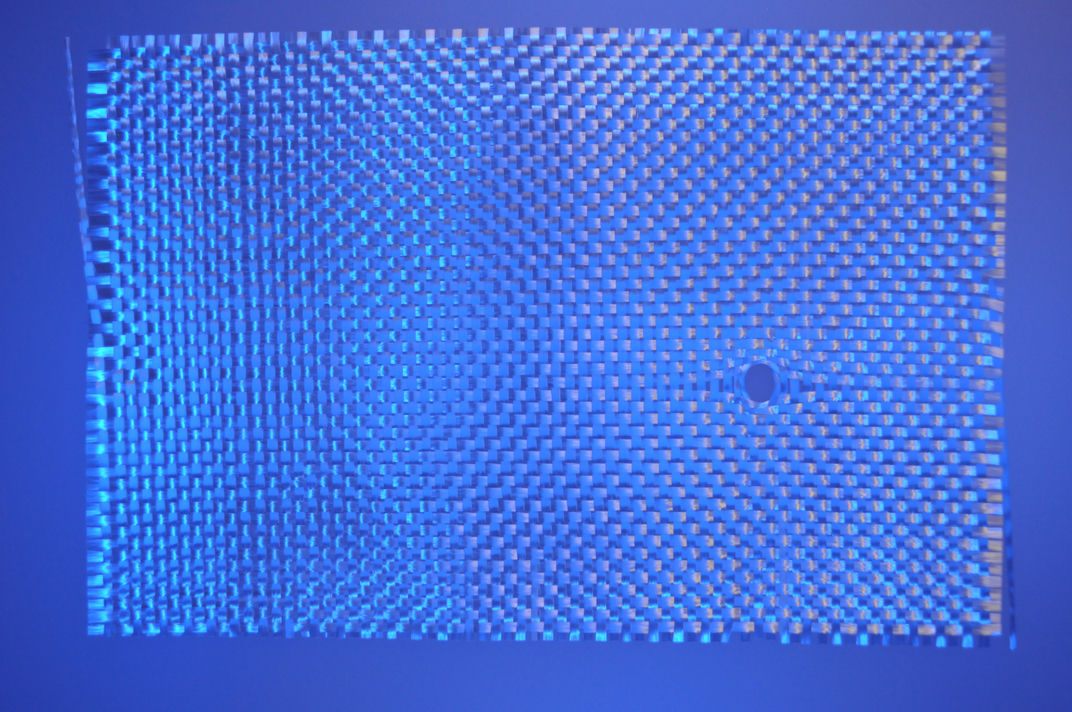}
        \includegraphics[width=0.24\linewidth]{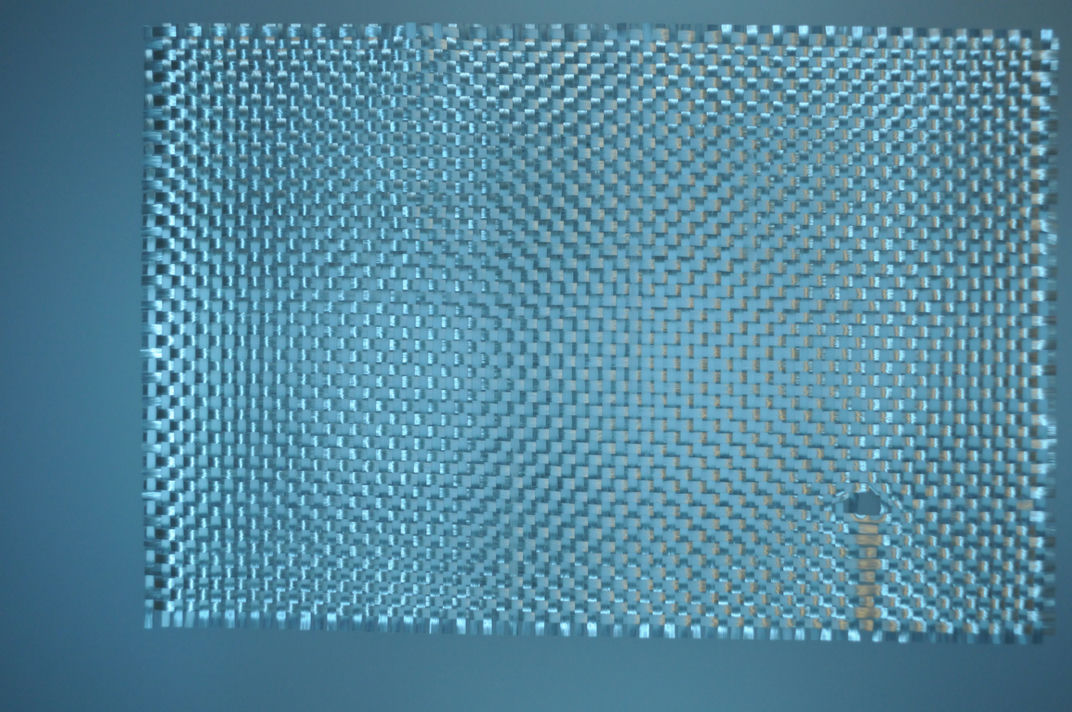}
        \caption{Class 2: ``Hole defect'', fabric pierced with a pin.}
        \label{subfig:TT_2}
    \end{subfigure}

    \begin{subfigure}{\textwidth}
        \centering
        \includegraphics[width=0.24\linewidth]{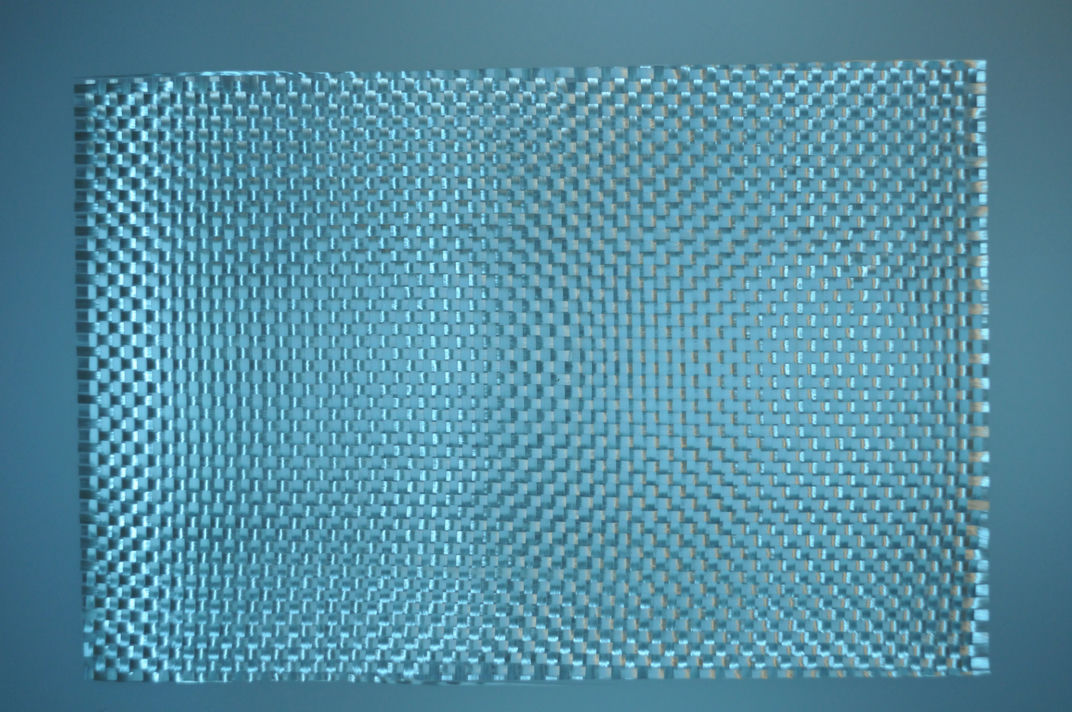}
        \includegraphics[width=0.24\linewidth]{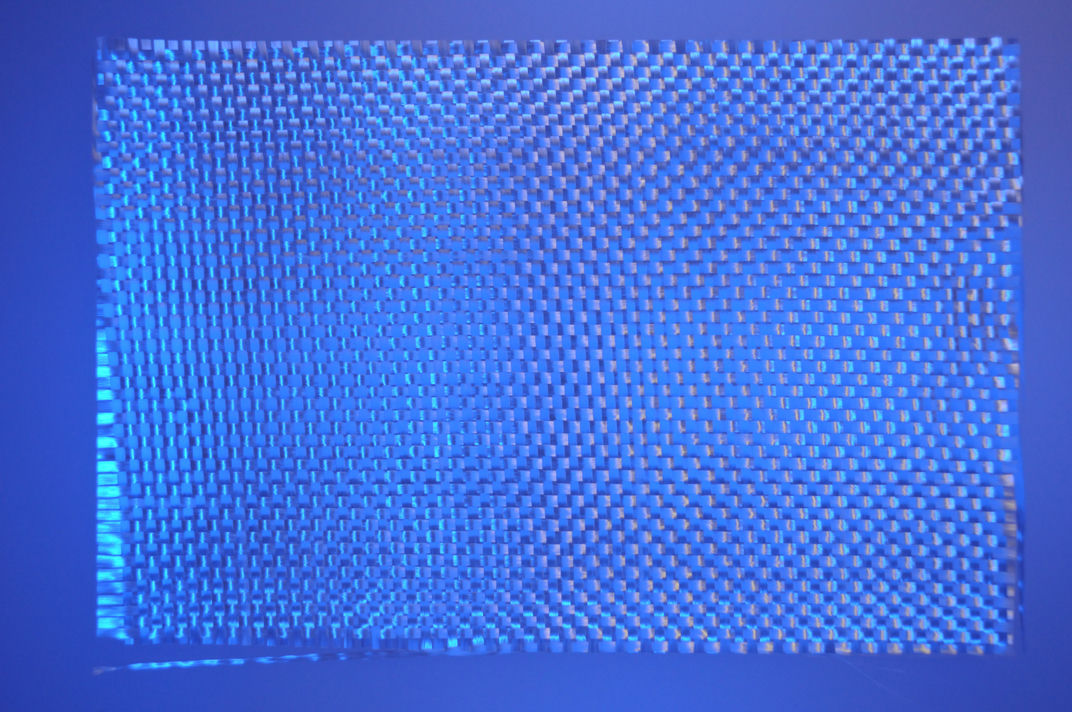}
        \includegraphics[width=0.24\linewidth]{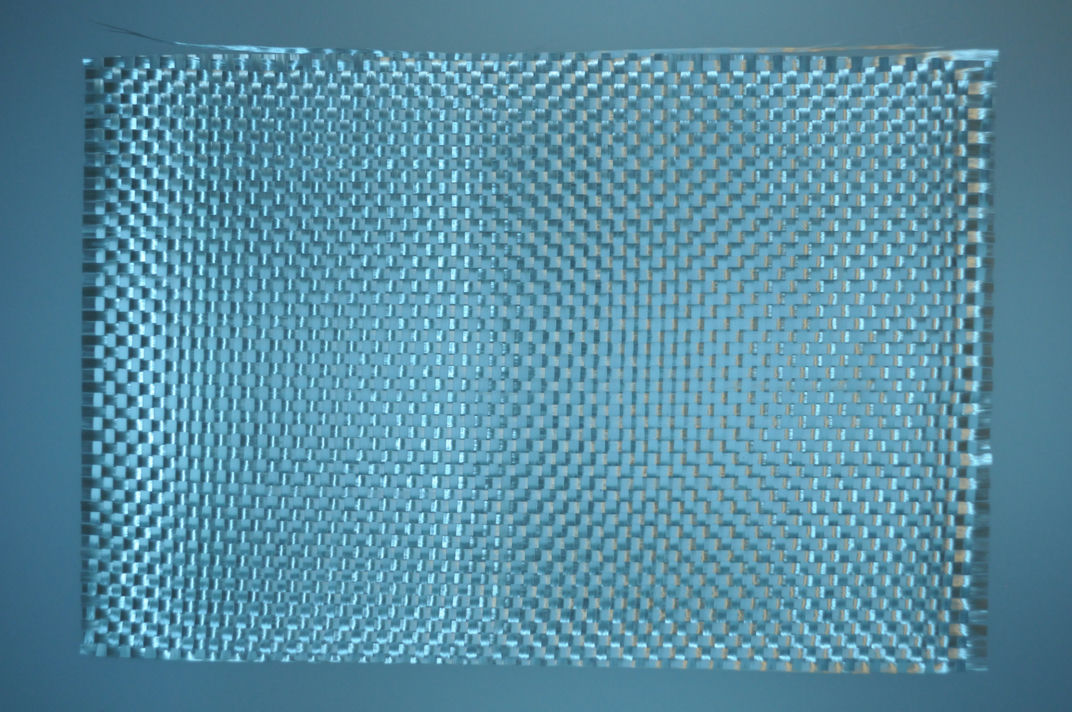}
        \includegraphics[width=0.24\linewidth]{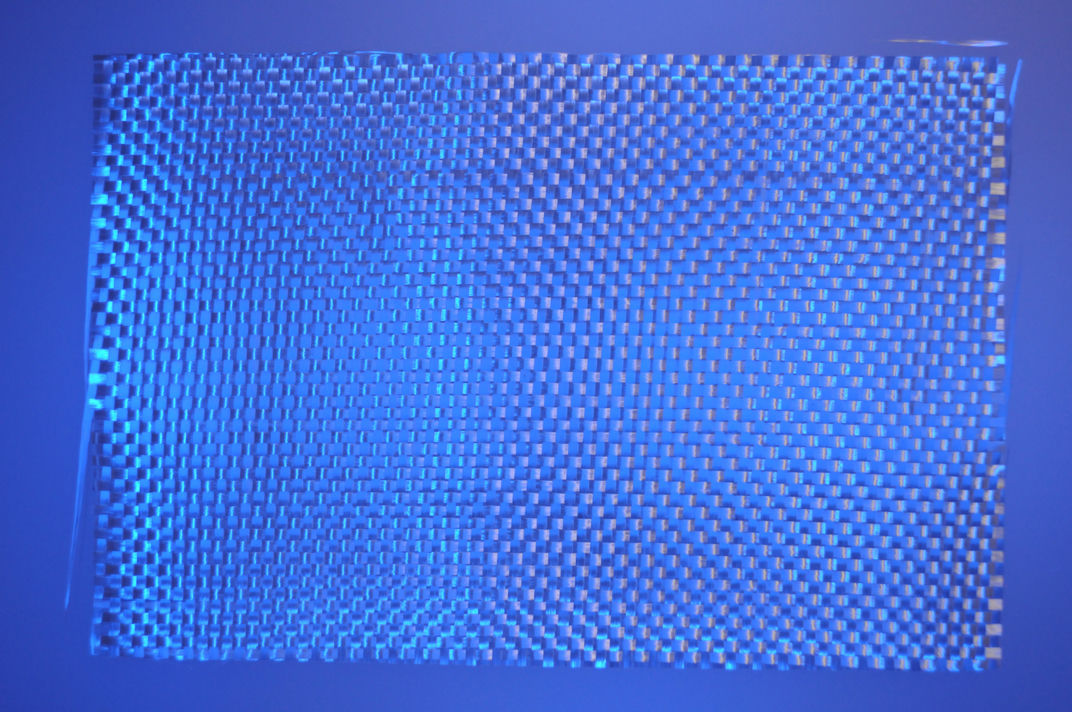}
        \caption{Class 3: ``Error-free'', fabric with plain weave without special features.}
        \label{subfig:TT_3}
    \end{subfigure}

    \begin{subfigure}{\textwidth}
        \centering
        \includegraphics[width=0.24\linewidth]{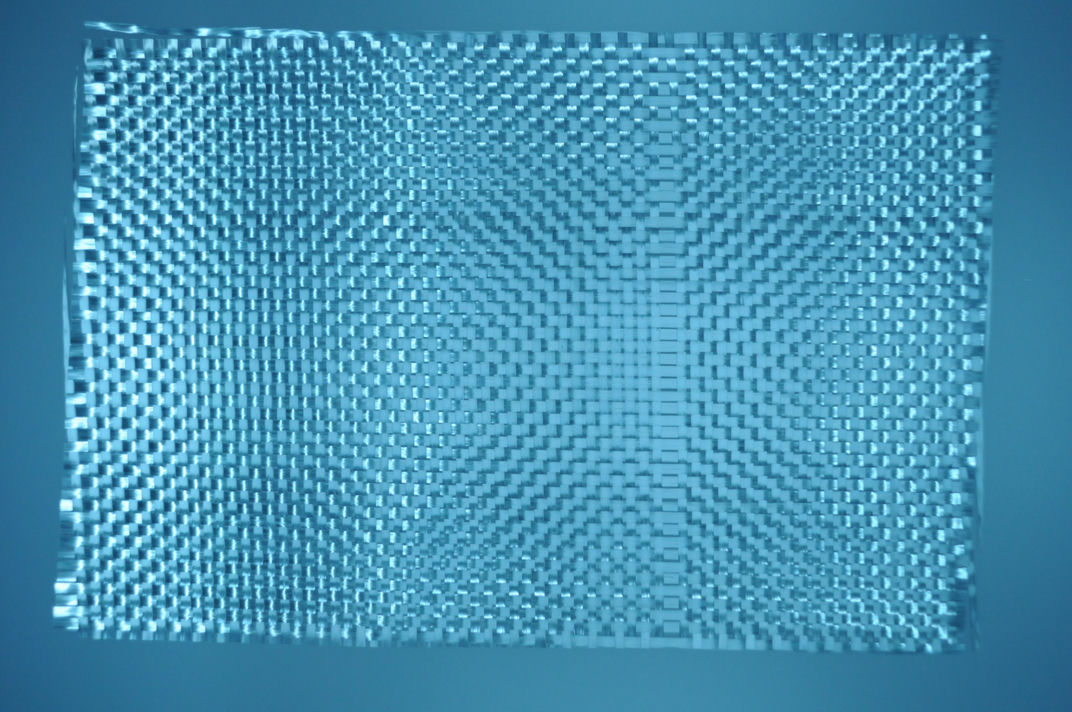}
        \includegraphics[width=0.24\linewidth]{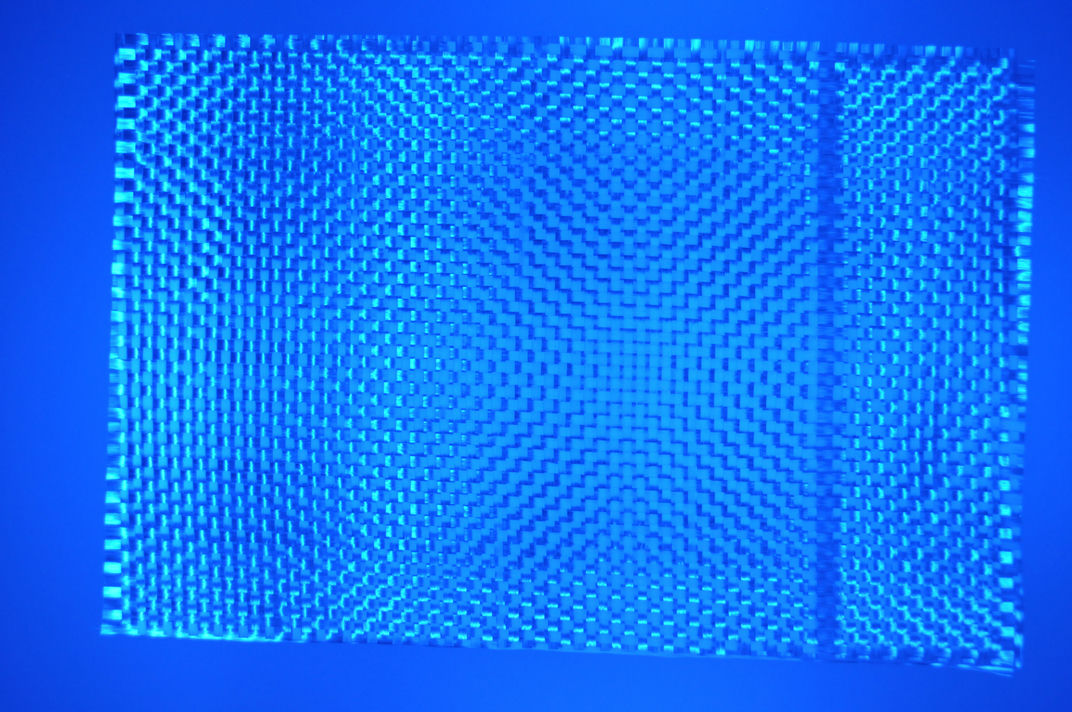}
        \includegraphics[width=0.24\linewidth]{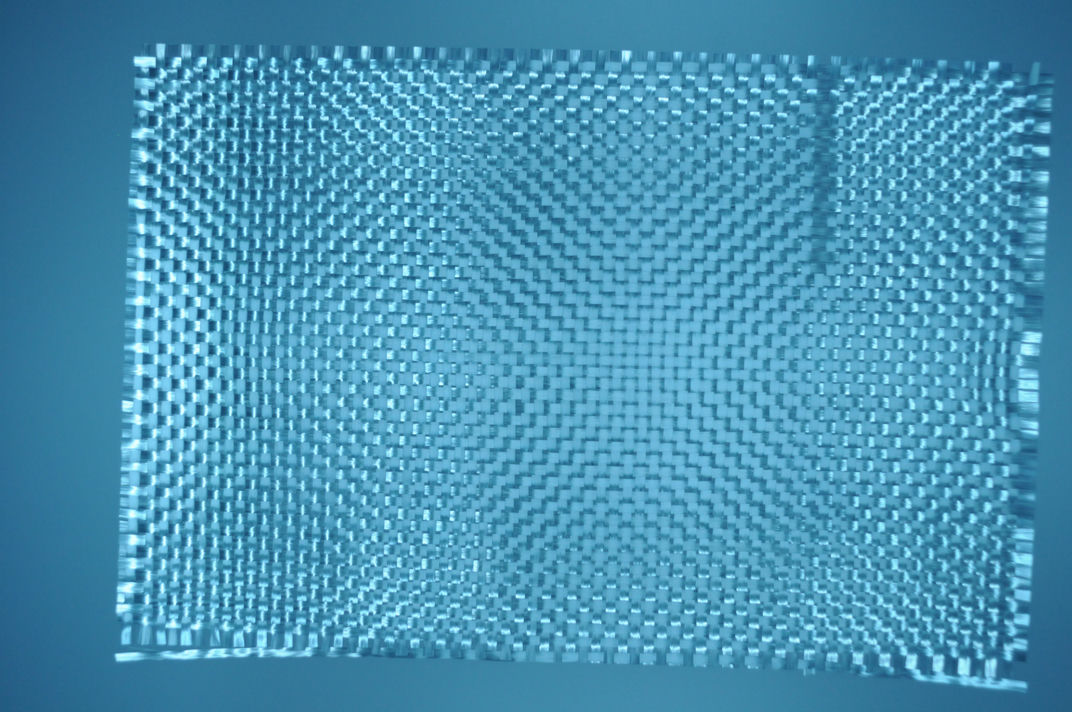}
        \includegraphics[width=0.24\linewidth]{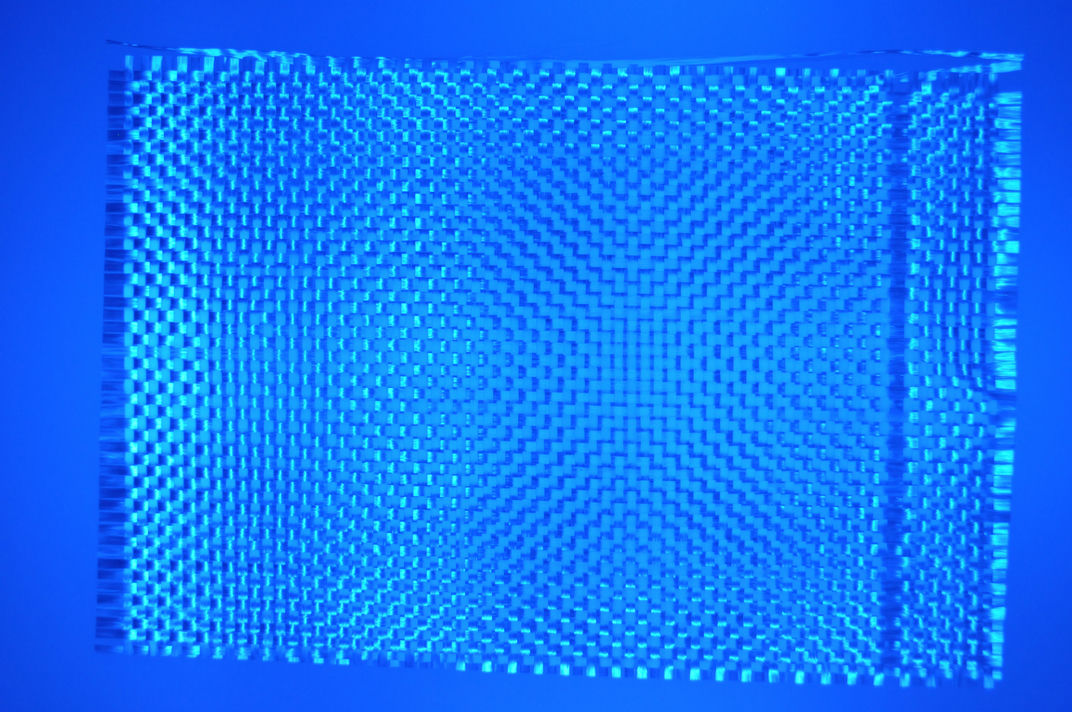}
        \caption{Class 4: ``Thread breakage'', plain weave with (partially) broken weft thread.}
        \label{subfig:TT_4}
    \end{subfigure}

    \begin{subfigure}{\textwidth}
        \centering
        \includegraphics[width=0.24\linewidth]{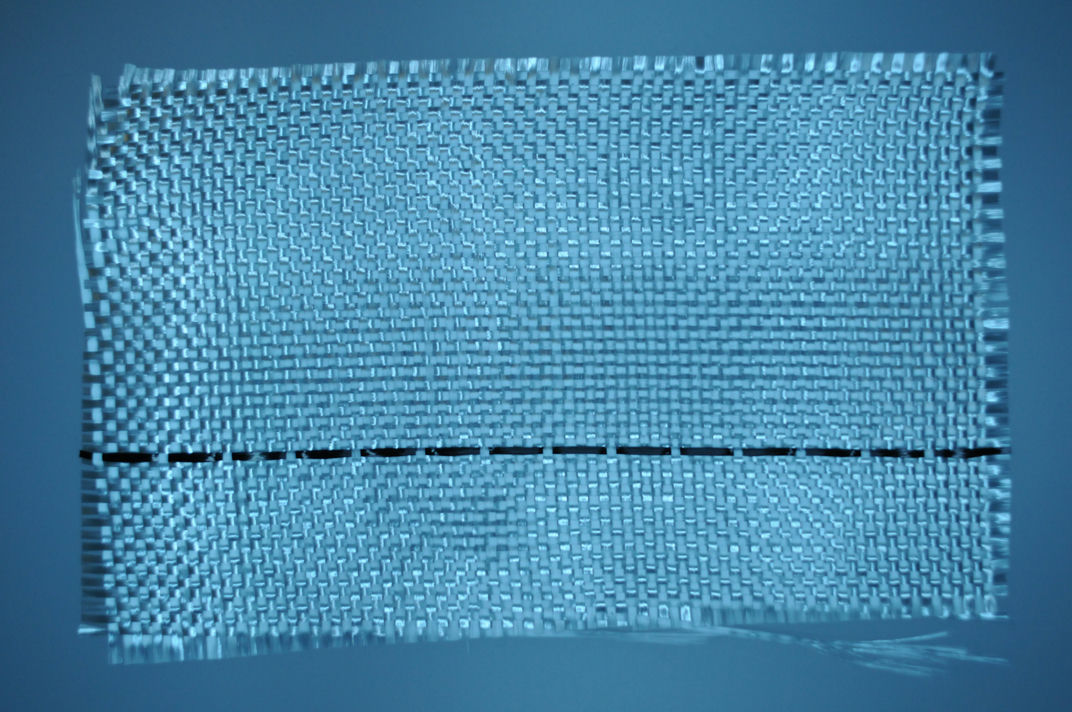}
        \includegraphics[width=0.24\linewidth]{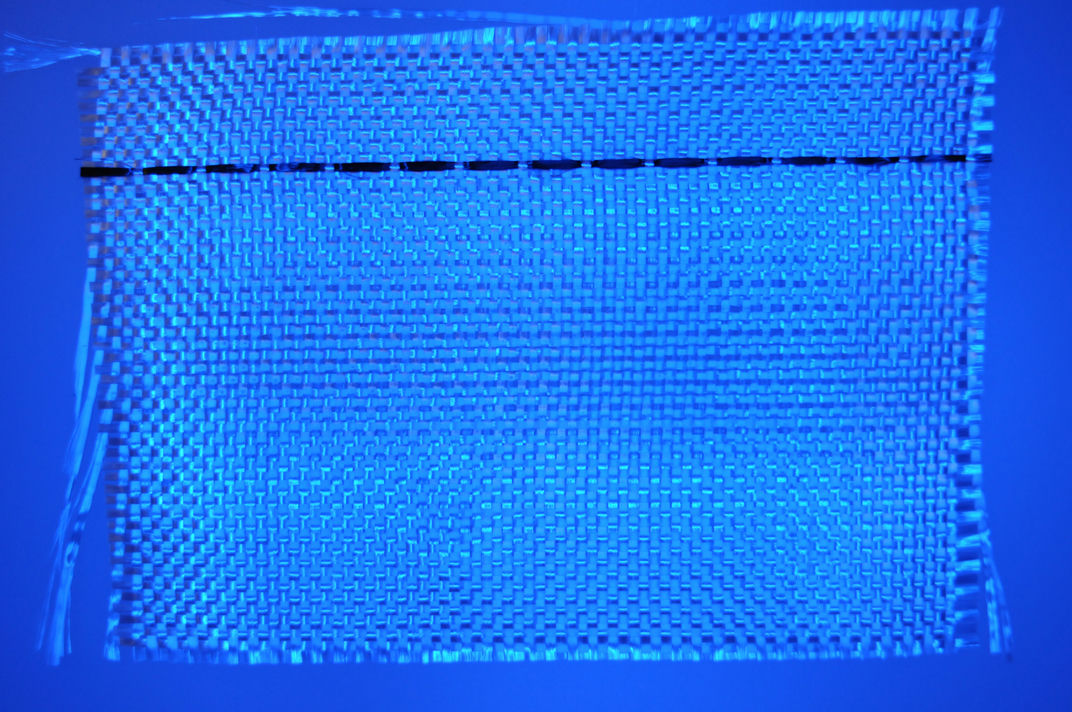}
        \includegraphics[width=0.24\linewidth]{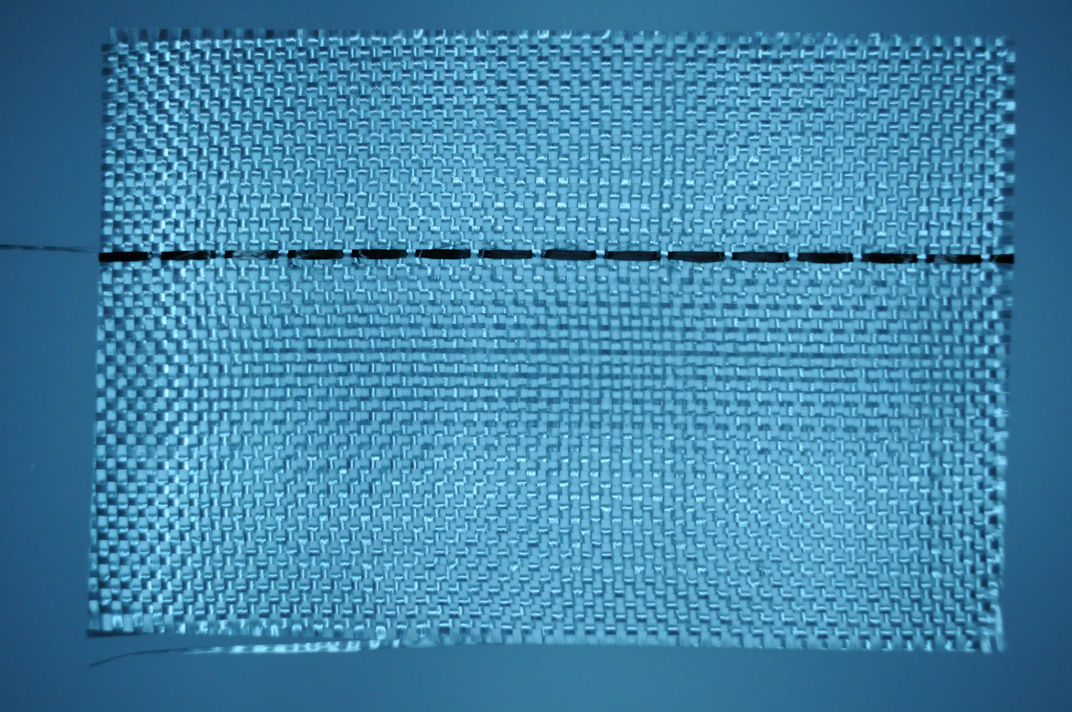}
        \includegraphics[width=0.24\linewidth]{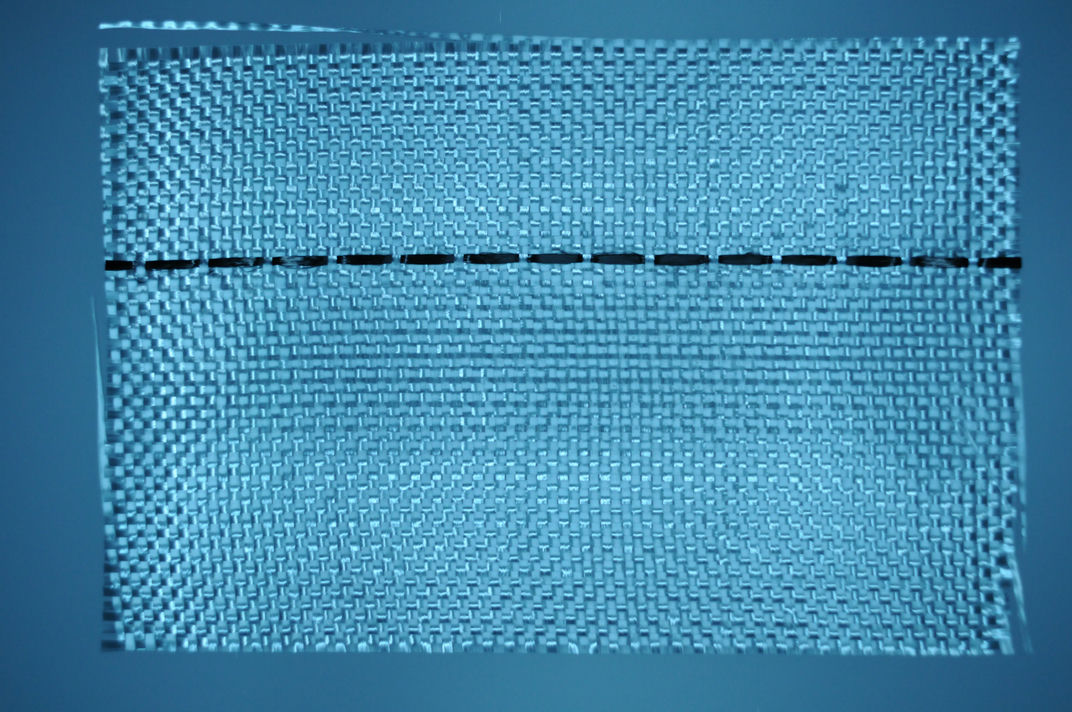}
        \caption{Class 5: ``22 mm binding'', C-fiber reinforced textile tied every 22 mm.}
        \label{subfig:TT_5}
    \end{subfigure}

    \caption{Samples images from the fiber glass errors dataset. Figures above depict the different types of error in fiberglass production. Figure \ref{subfig:TT_3} depicts error free fiber glass textiles. Figure \ref{subfig:TT_4} contains textiles with a thread partially broken. Figure \ref{subfig:TT_2} depicts textiles with a puncture hole. Whereas Figures \ref{subfig:TT_0},\ref{subfig:TT_1},\ref{subfig:TT_5} contain three different types of binding errors.}
    \label{fig:TT}
\end{figure}

In this context, a TT dataset was generated to represent a practical quality control scenario. In this experiment, continuous material rolls of glass fiber fabric with plain weave were cut into sizes of 300$\times$200 mm. Among these samples, half were reinforced with a single carbon fiber. The samples were classified into six classes of 300 images of 4288$\times$2848 pixels each, based on the presence of common defects or error free textiles, as seen in Figure \ref{fig:TT}. \footnote{The dataset is available under the link: \url{https://doi.org/10.5281/zenodo.7970596}.}

The problem detecting quality issues in the fiber glass images was formalized as an image classification problem. In a practical sense, this allows for the usage of different kinds of CNN architecture, training procedures, and data augmentations, which finally allow the training of a model attaining a high accuracy.
In this use case, we used a Densenet121 \cite{DBLP:conf/cvpr/HuangLMW17}, training it with a train-test split of 0.8 until convergence. 
We used an initial learning rate of 0.1, a reduce on plateau scheduler, and weight decay.
During training, the images were resized to a shape of 512$\times$512, making them small enough for an efficient evaluation, yet big enough not to lose the details of the fiber glass errors. The final model achieved perfect accuracy, yet, the question rises of whether it learned the task as intended, or there was an unintentional bias on the provided data.

\begin{figure}[!htb]
    \centering
    \begin{subfigure}{\textwidth}
        \centering
        \includegraphics[width=0.24\linewidth]{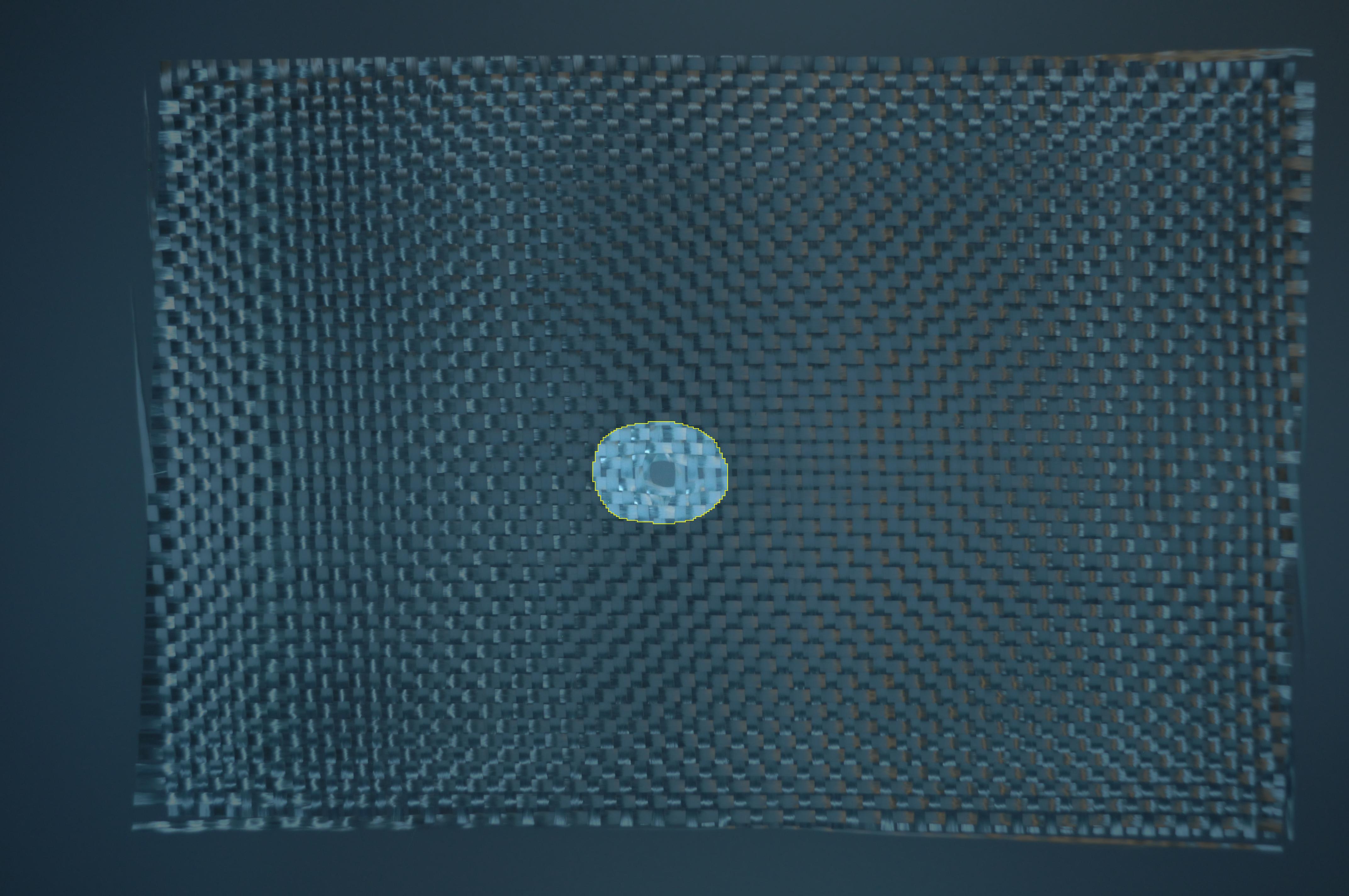}
        \includegraphics[width=0.24\linewidth]{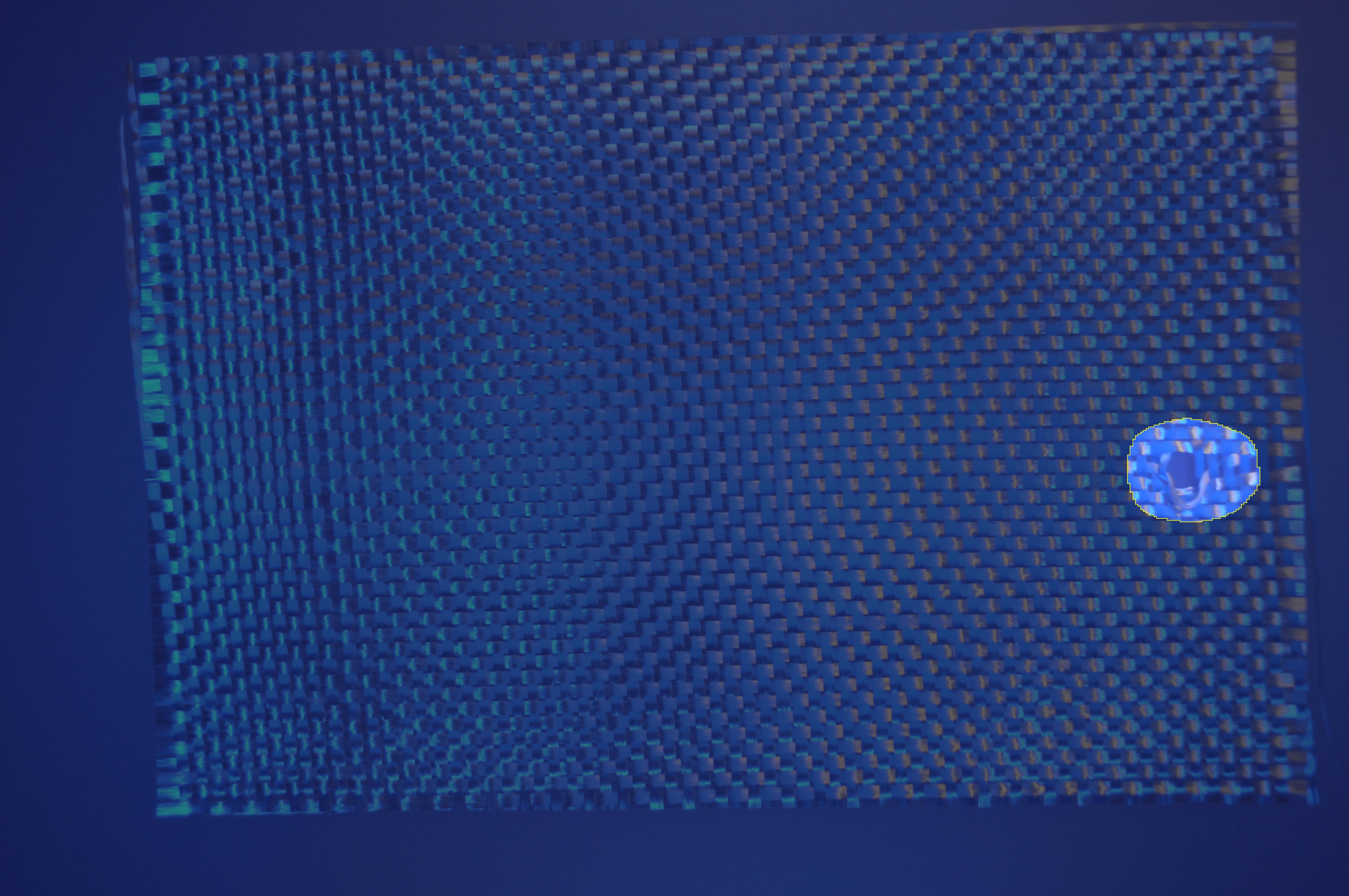}
        \includegraphics[width=0.24\linewidth]{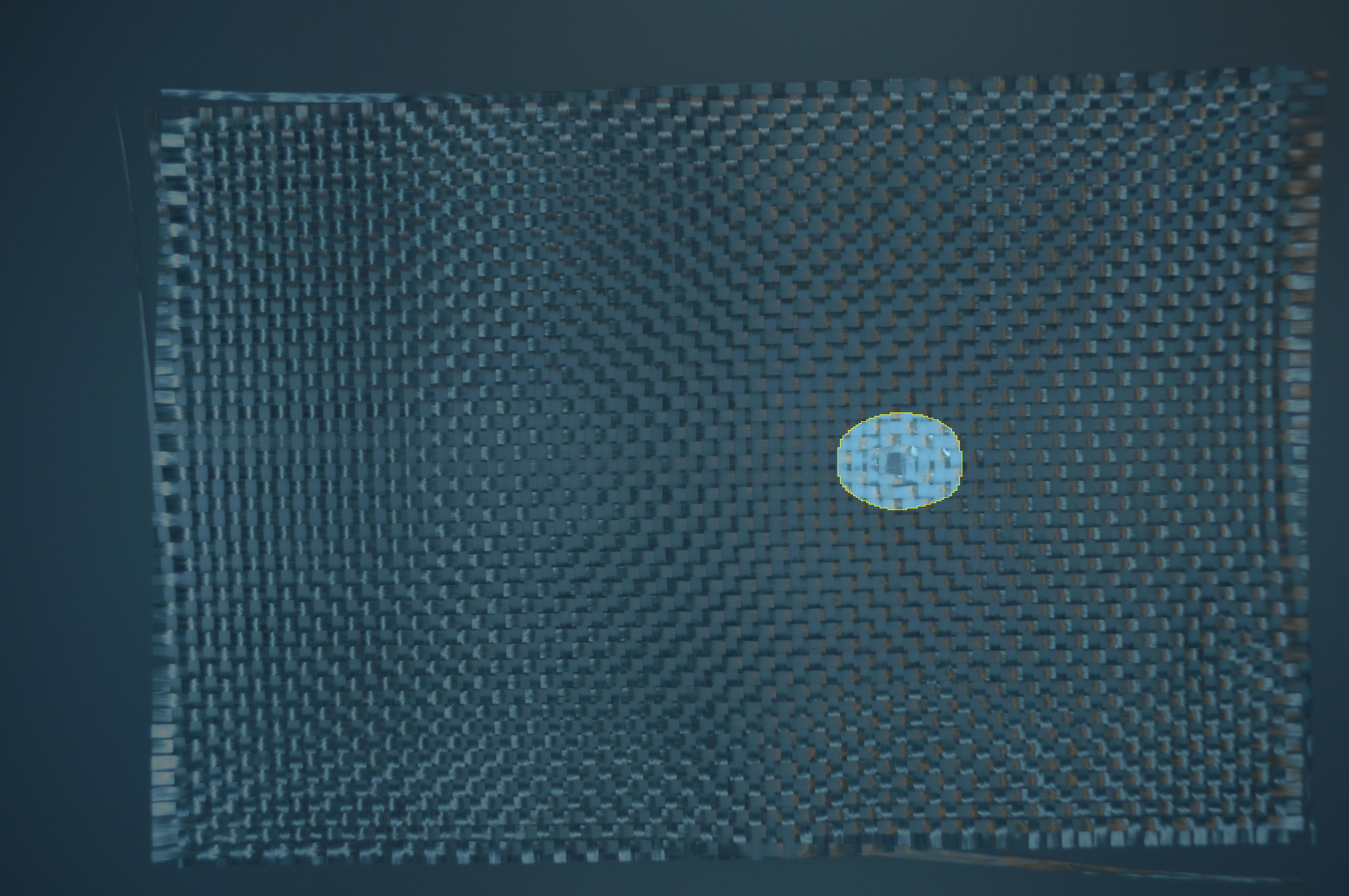}
        \includegraphics[width=0.24\linewidth]{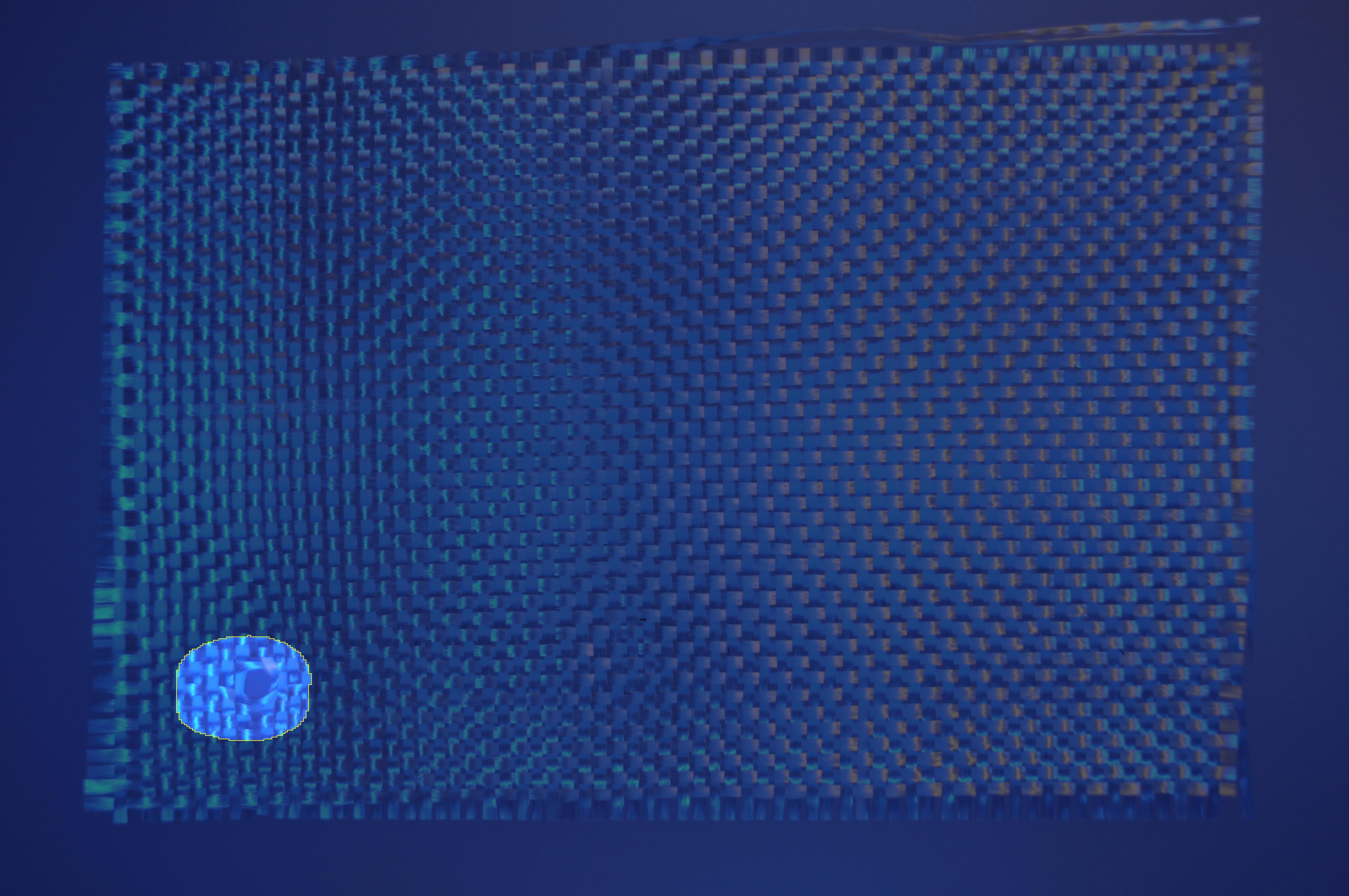}
        \caption{Concept 13, with a scaled importance score of 0.20.}
        \label{subfig:TT_C13}
    \end{subfigure}
    
    \begin{subfigure}{\textwidth}
        \centering
        \includegraphics[width=0.24\linewidth]{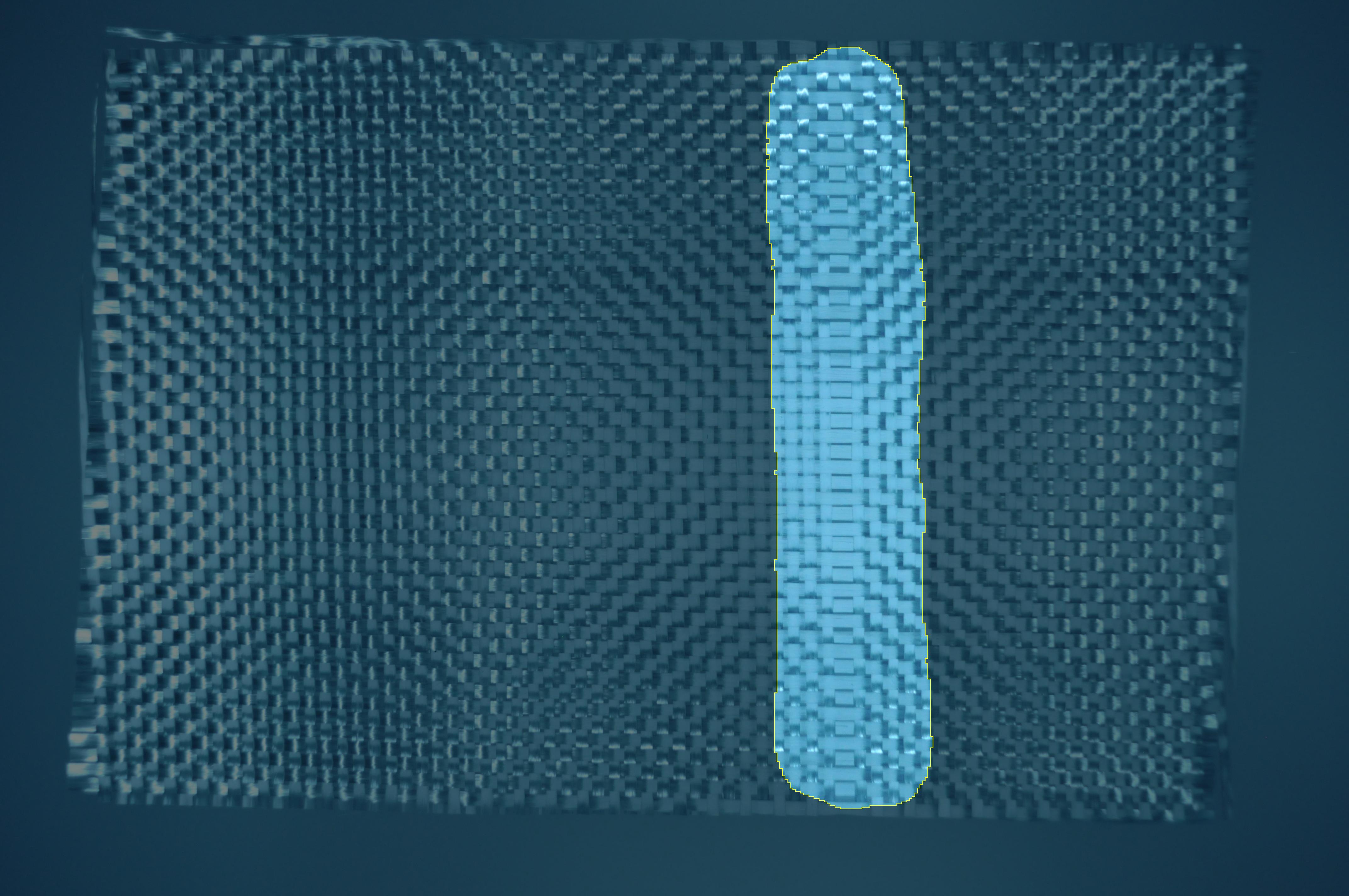}
        \includegraphics[width=0.24\linewidth]{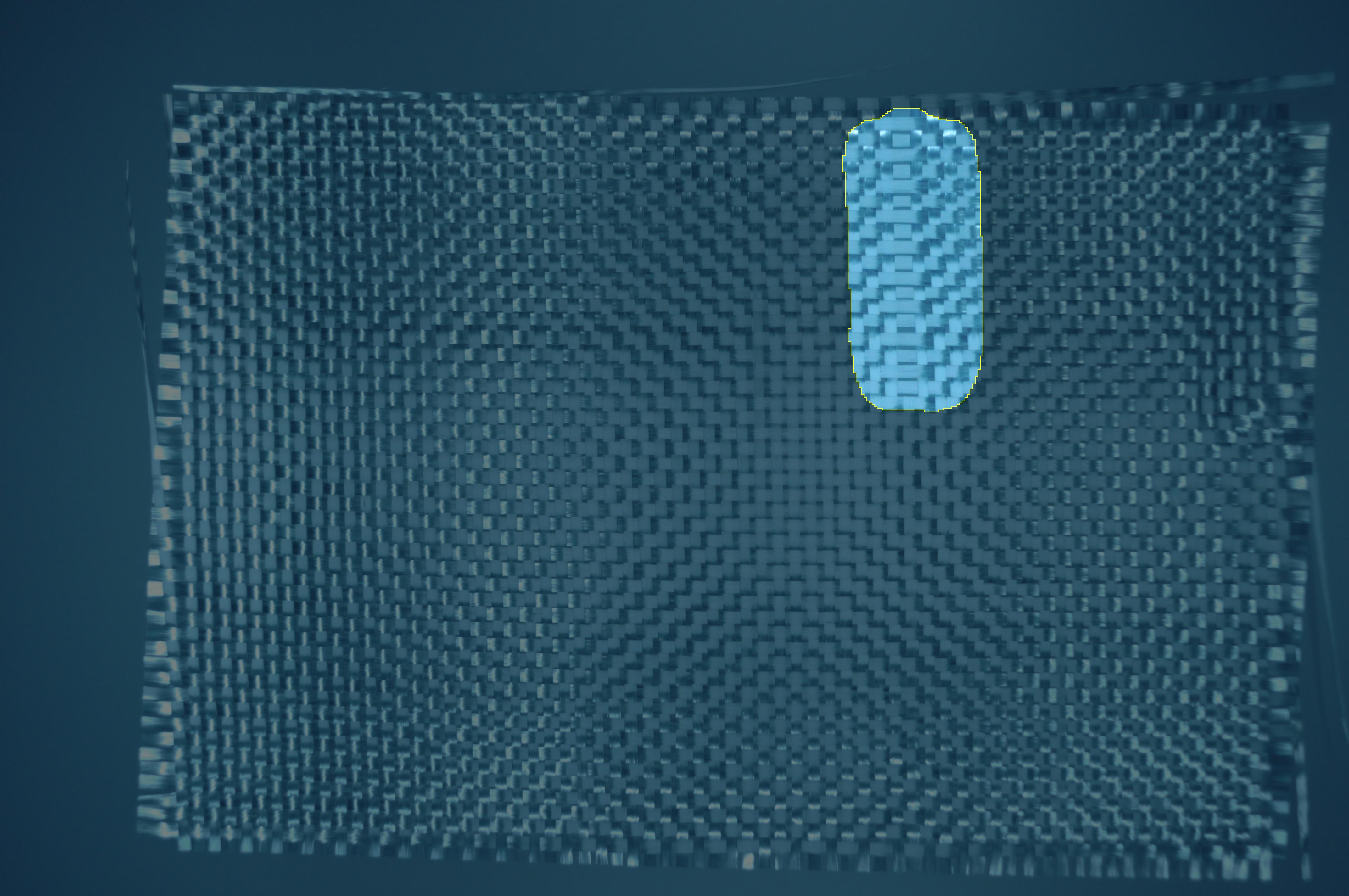}
        \includegraphics[width=0.24\linewidth]{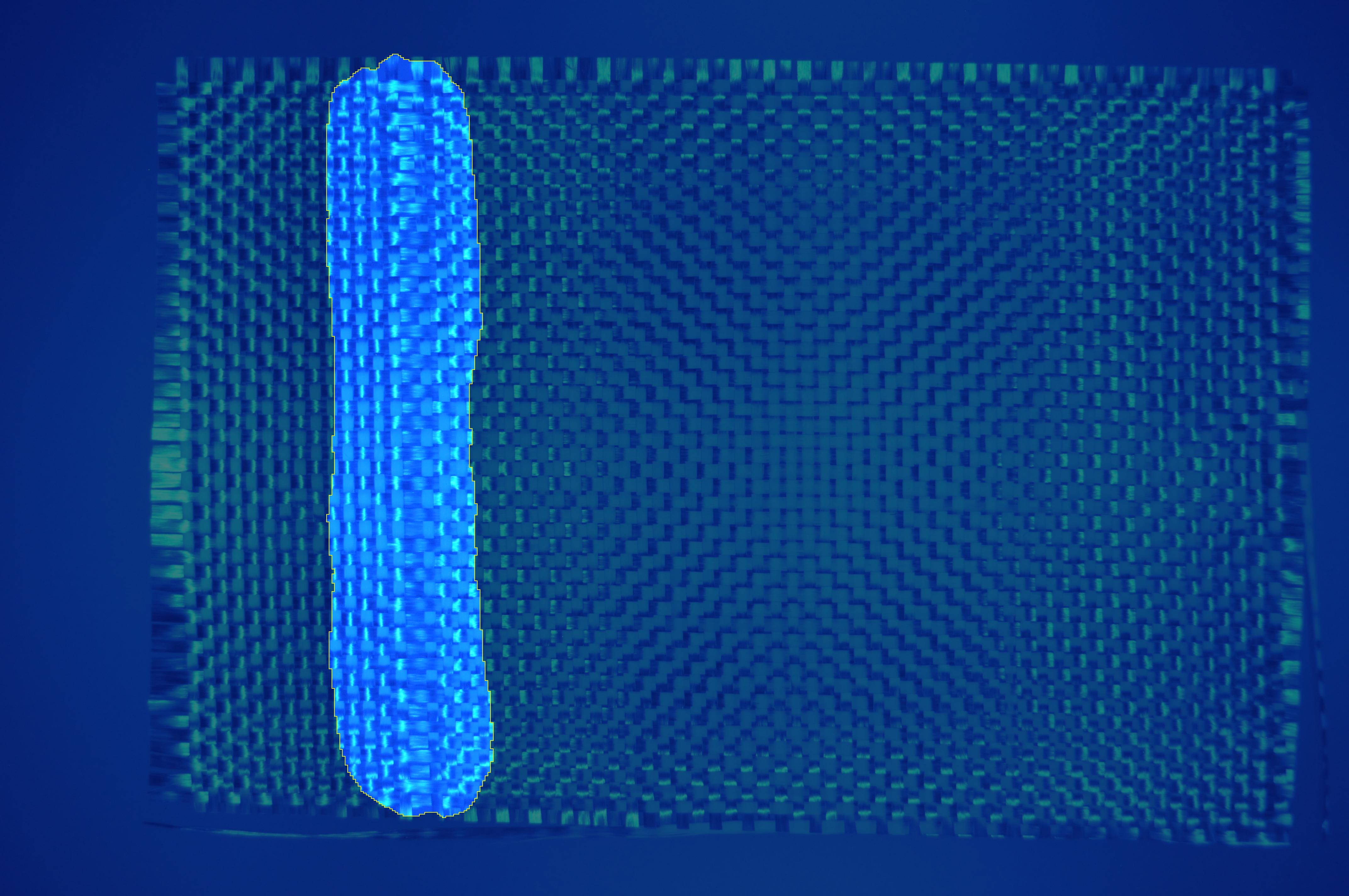}
        \includegraphics[width=0.24\linewidth]{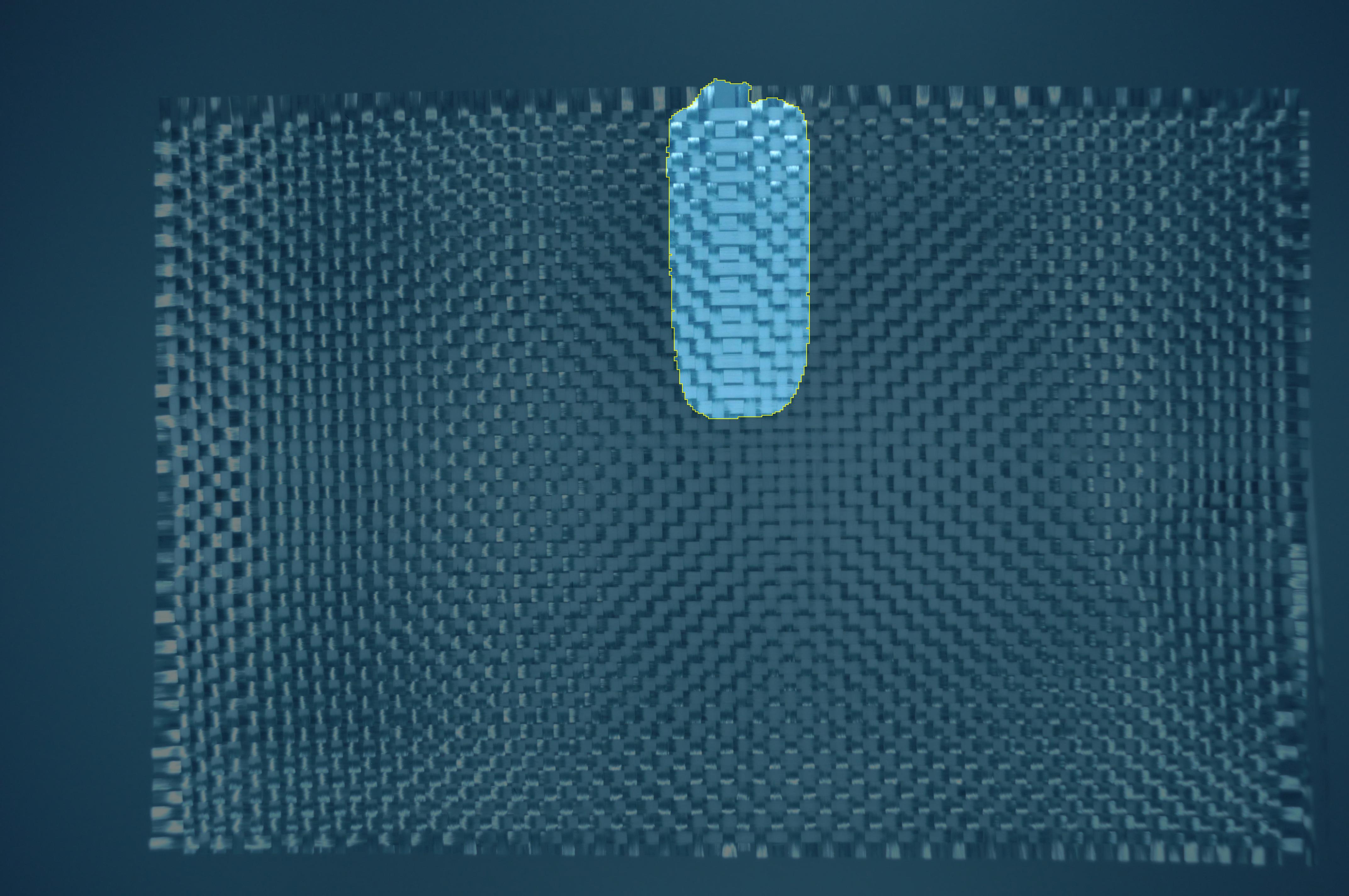}
        \caption{Concept 12, with a scaled importance score of 0.01.}
        \label{subfig:TT_C12}
    \end{subfigure}
    
    \begin{subfigure}{\textwidth}
        \centering
        \includegraphics[width=0.24\linewidth]{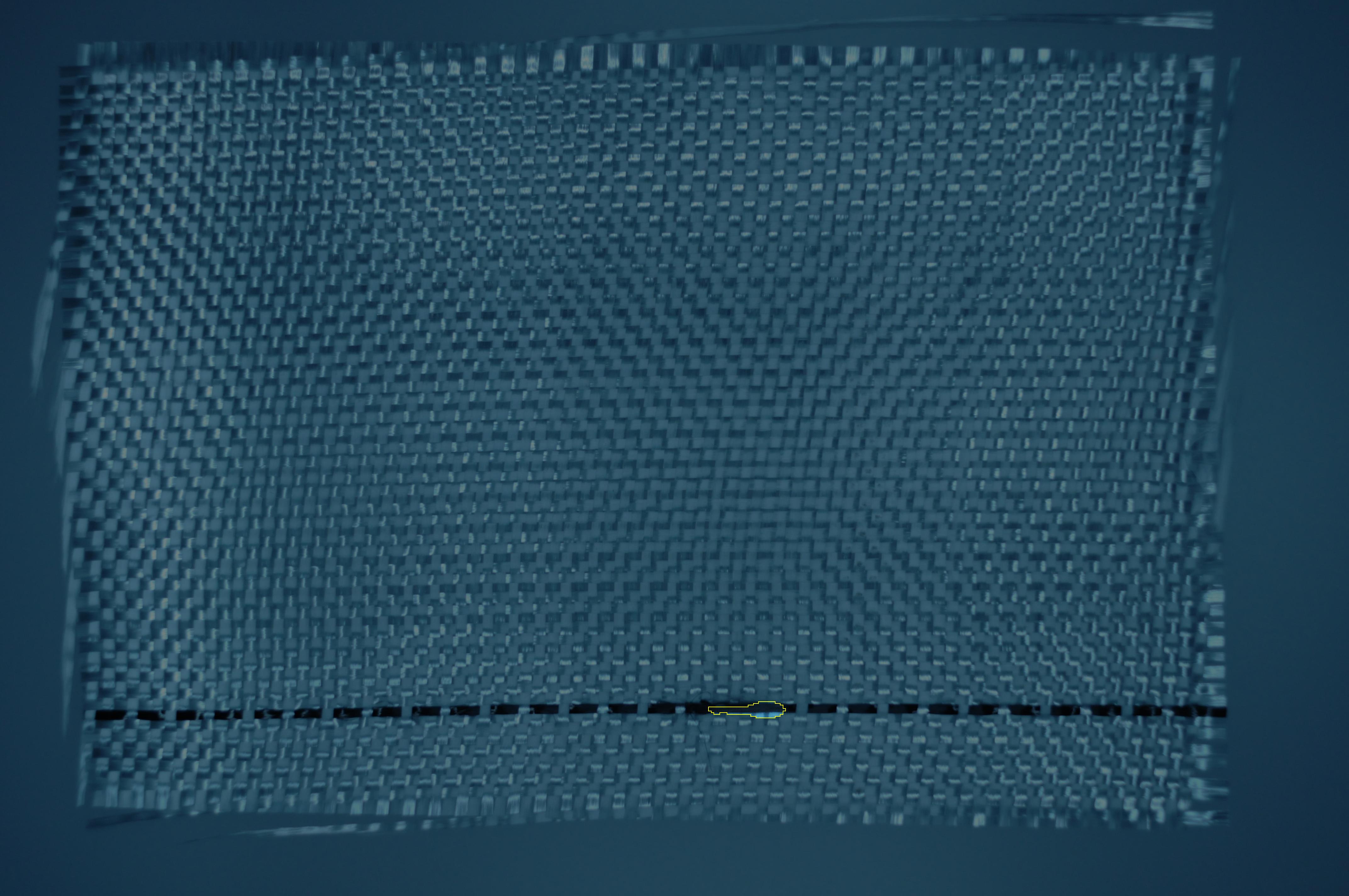}
        \includegraphics[width=0.24\linewidth]{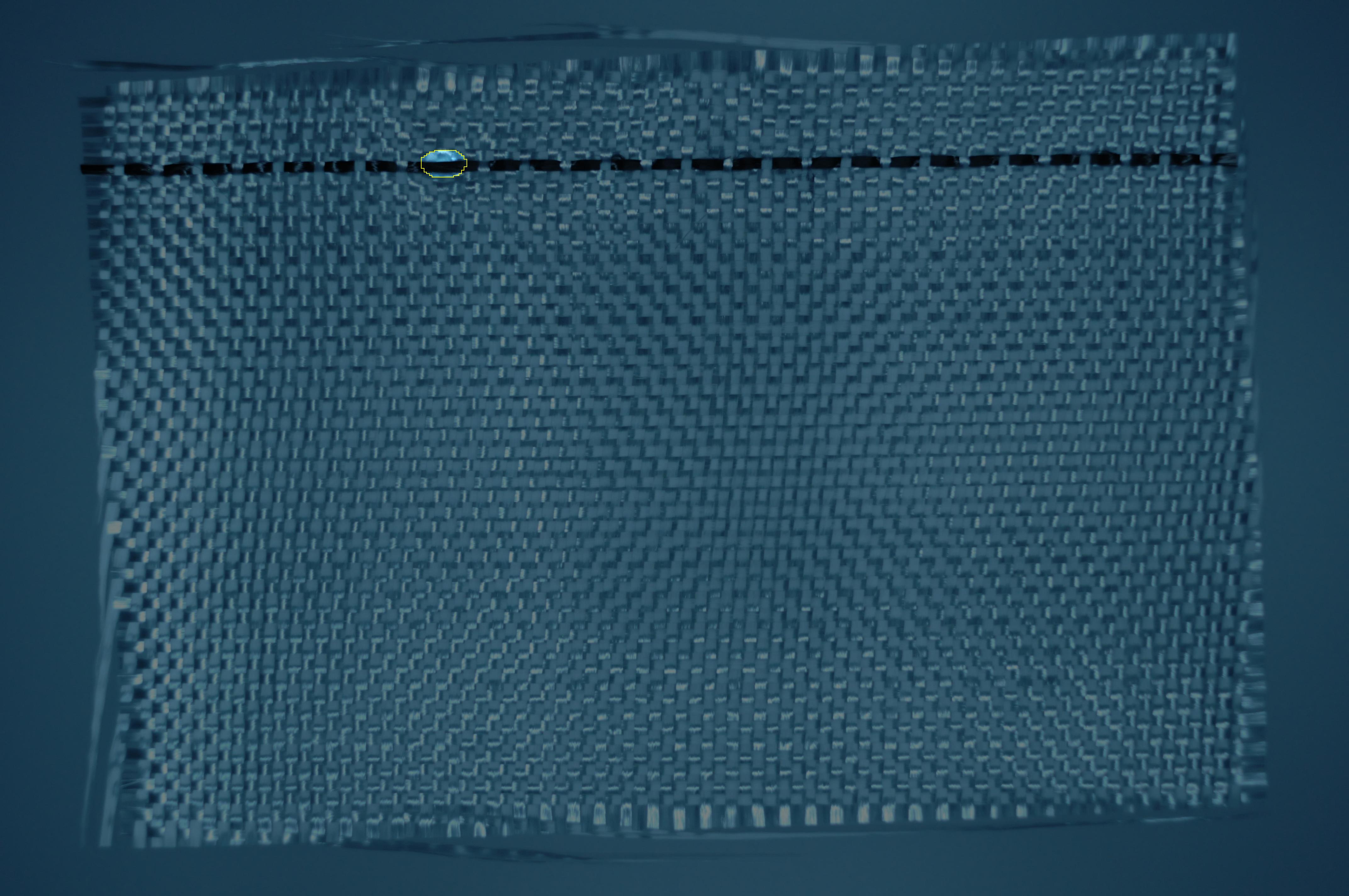}
        \includegraphics[width=0.24\linewidth]{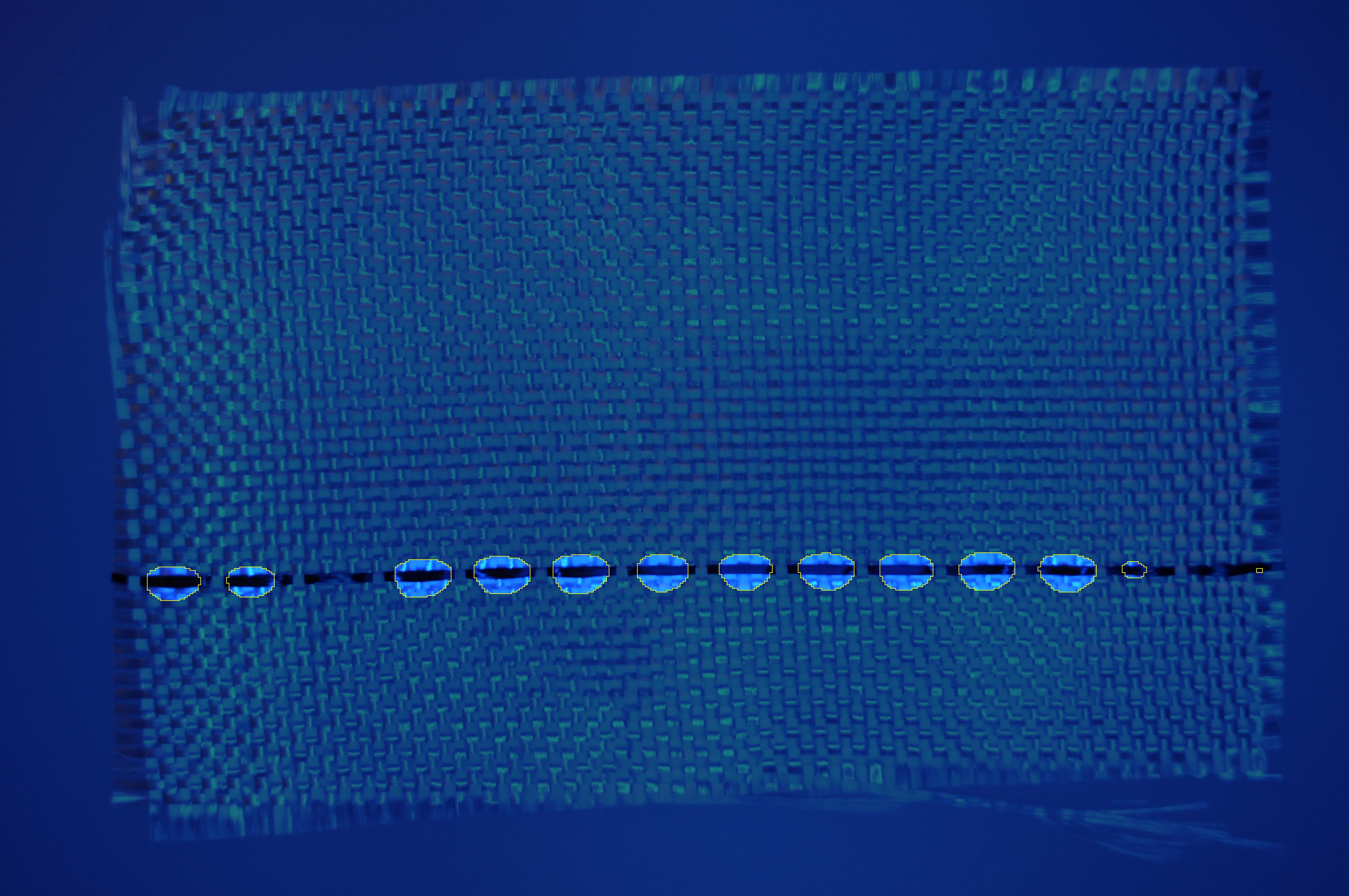}
        \includegraphics[width=0.24\linewidth]{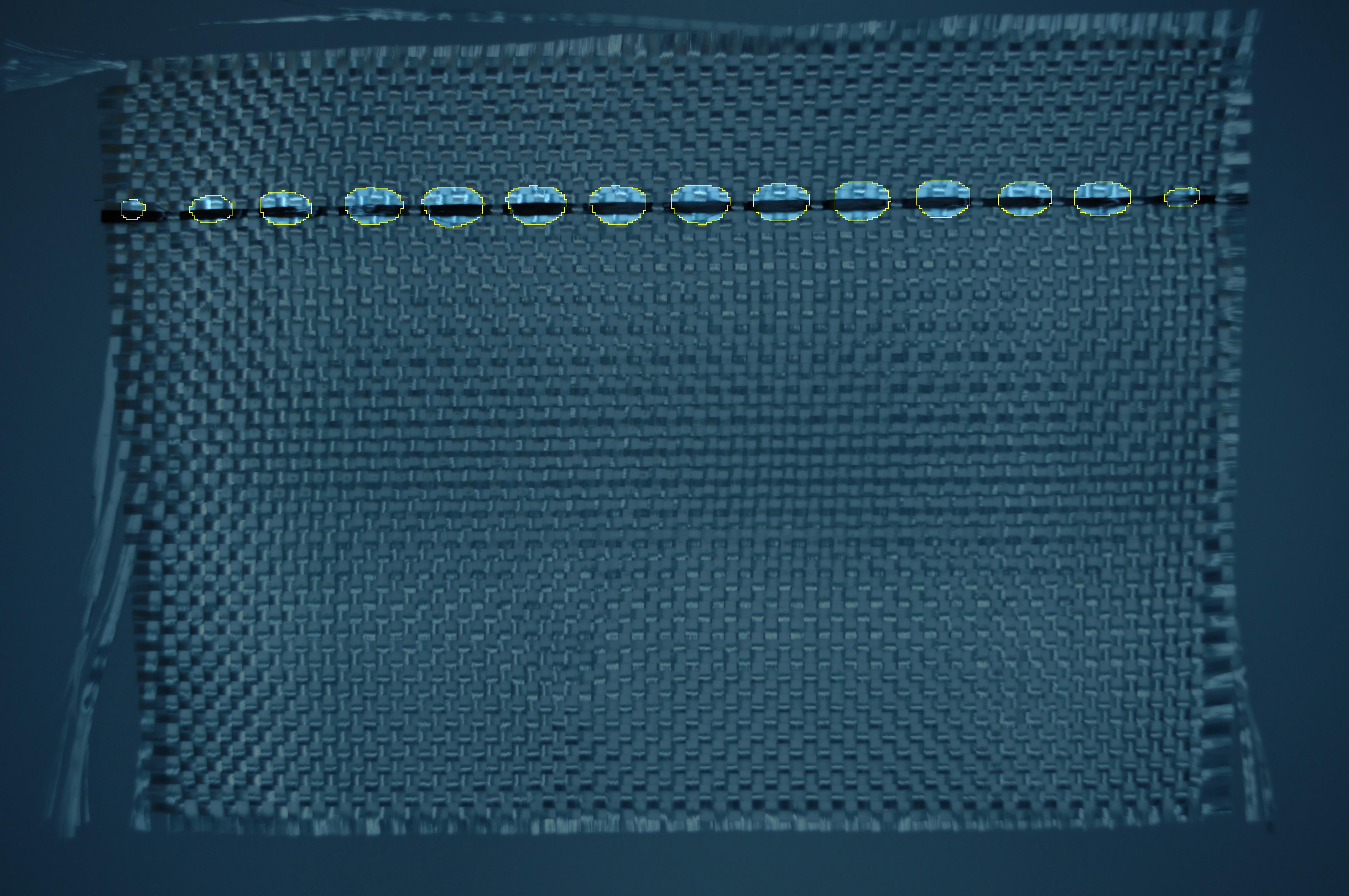}
        \caption{Concept 15, with a scaled importance score of 1.00.}
        \label{subfig:TT_C15}
    \end{subfigure}

    \begin{subfigure}{\textwidth}
        \centering
        \includegraphics[width=0.24\linewidth]{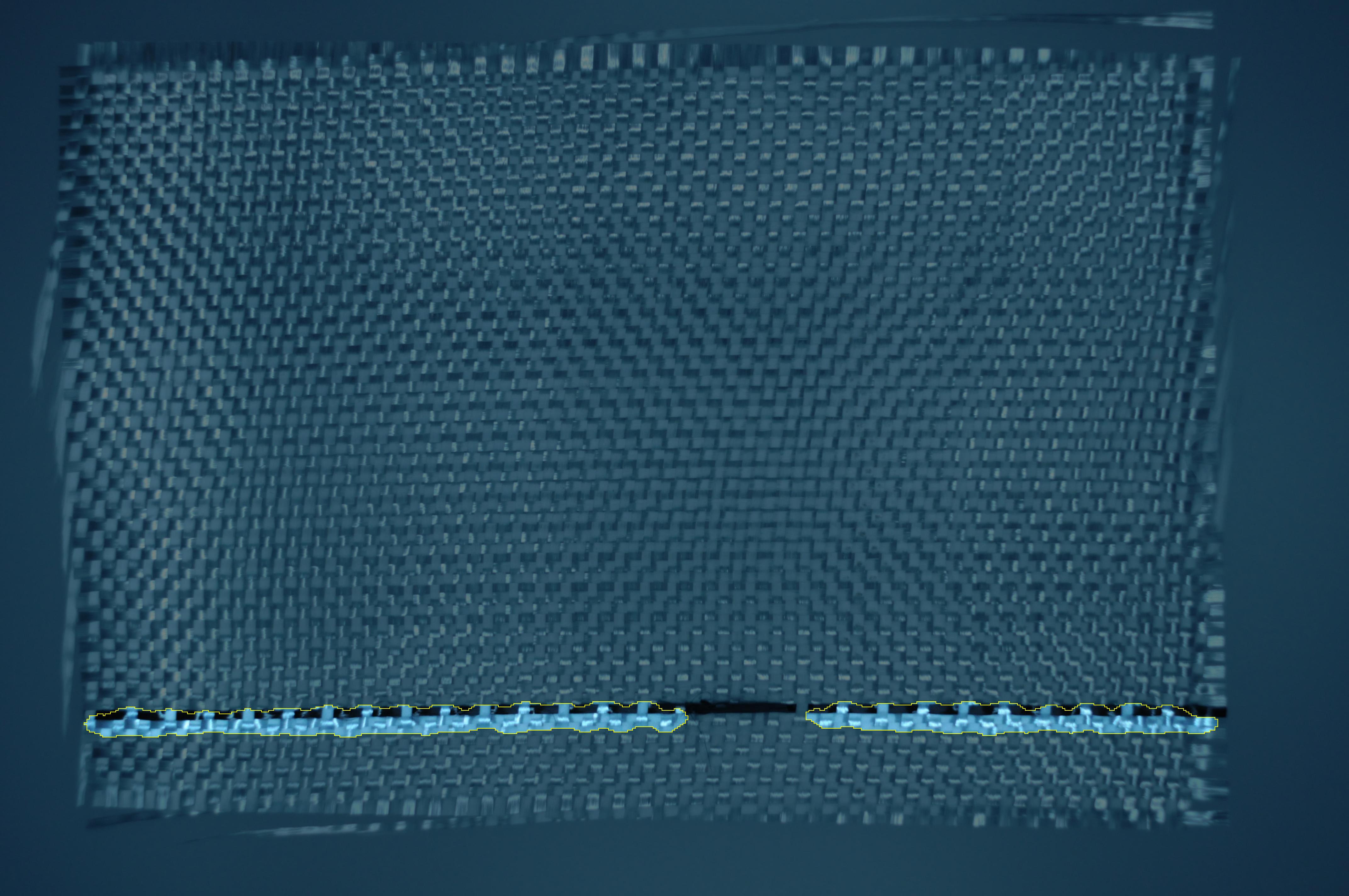}
        \includegraphics[width=0.24\linewidth]{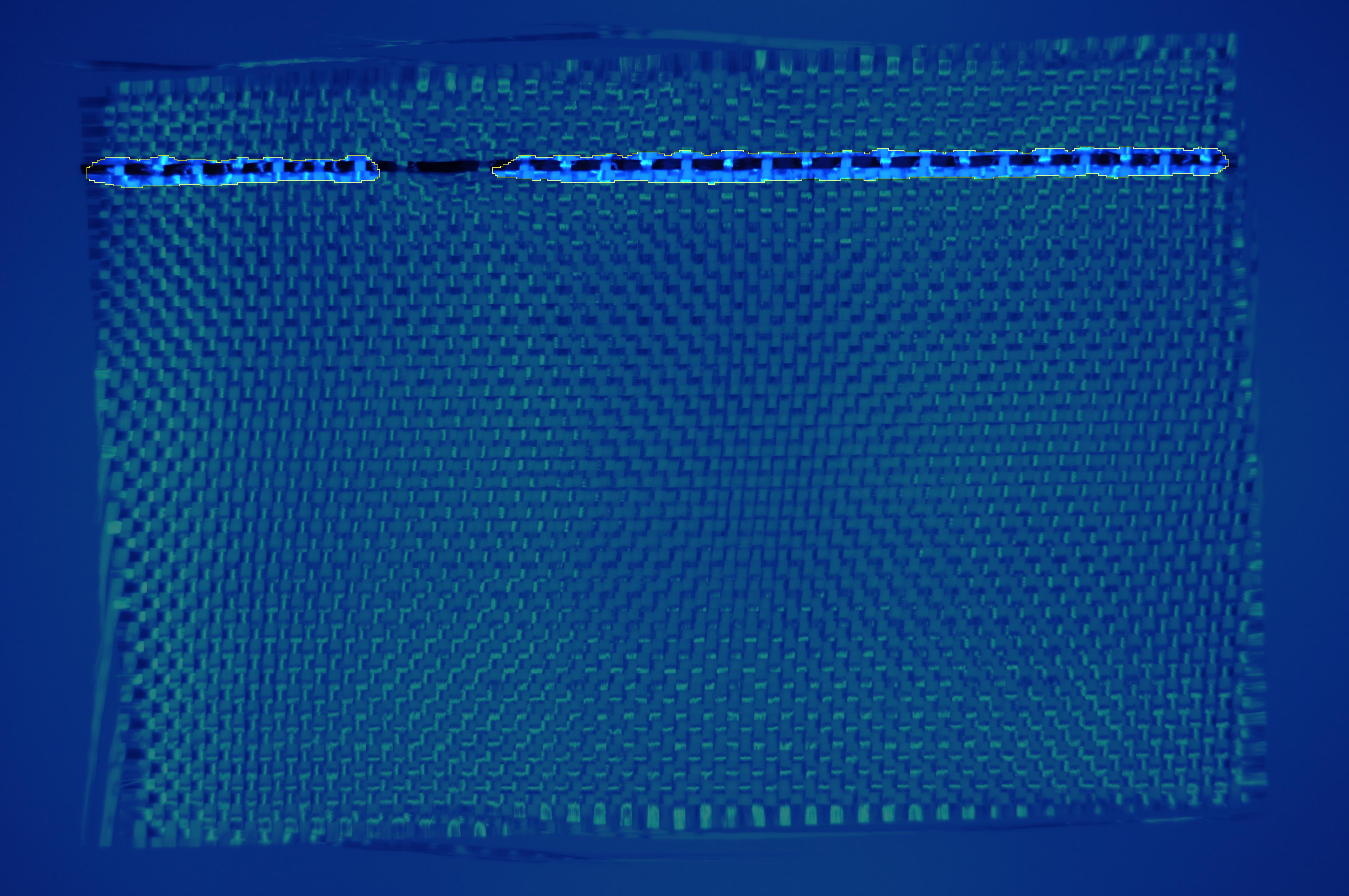}
        \includegraphics[width=0.24\linewidth]{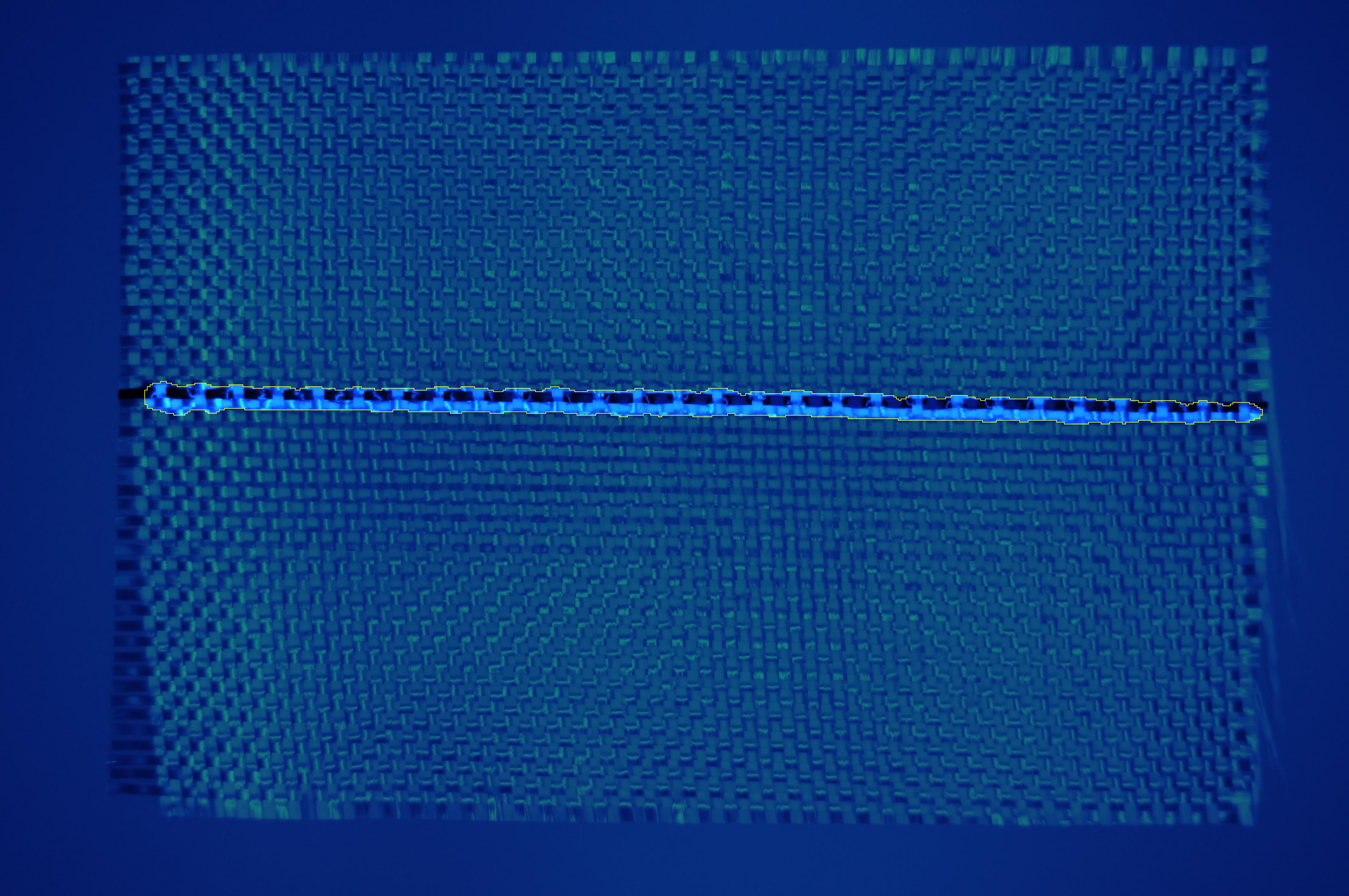}
        \includegraphics[width=0.24\linewidth]{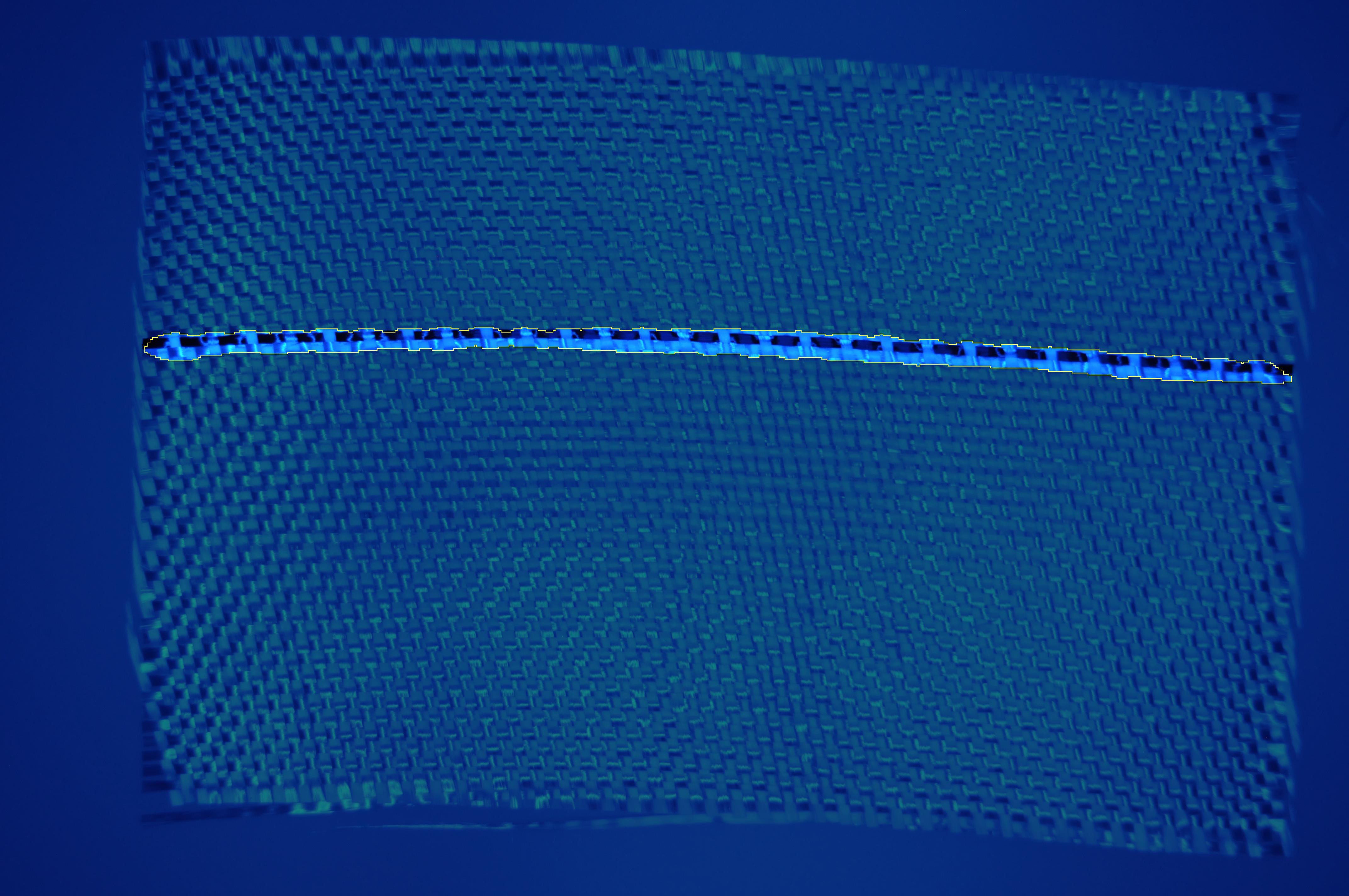}
        \caption{Concept 17, with a scaled importance score of 0.06.}
        \label{subfig:TT_C17}
    \end{subfigure}
    
    \begin{subfigure}{\textwidth}
        \centering
        \includegraphics[width=0.24\linewidth]{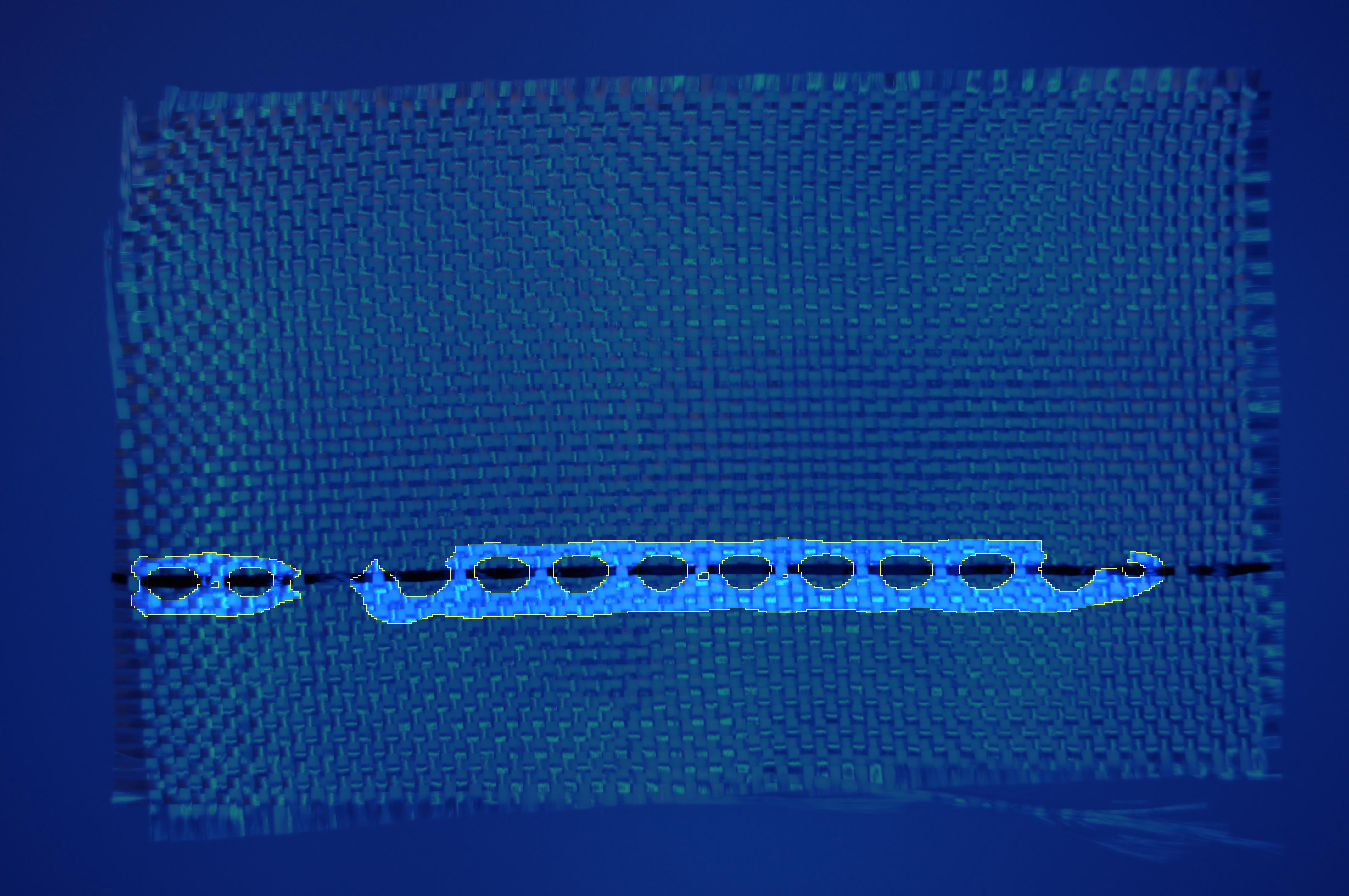}
        \includegraphics[width=0.24\linewidth]{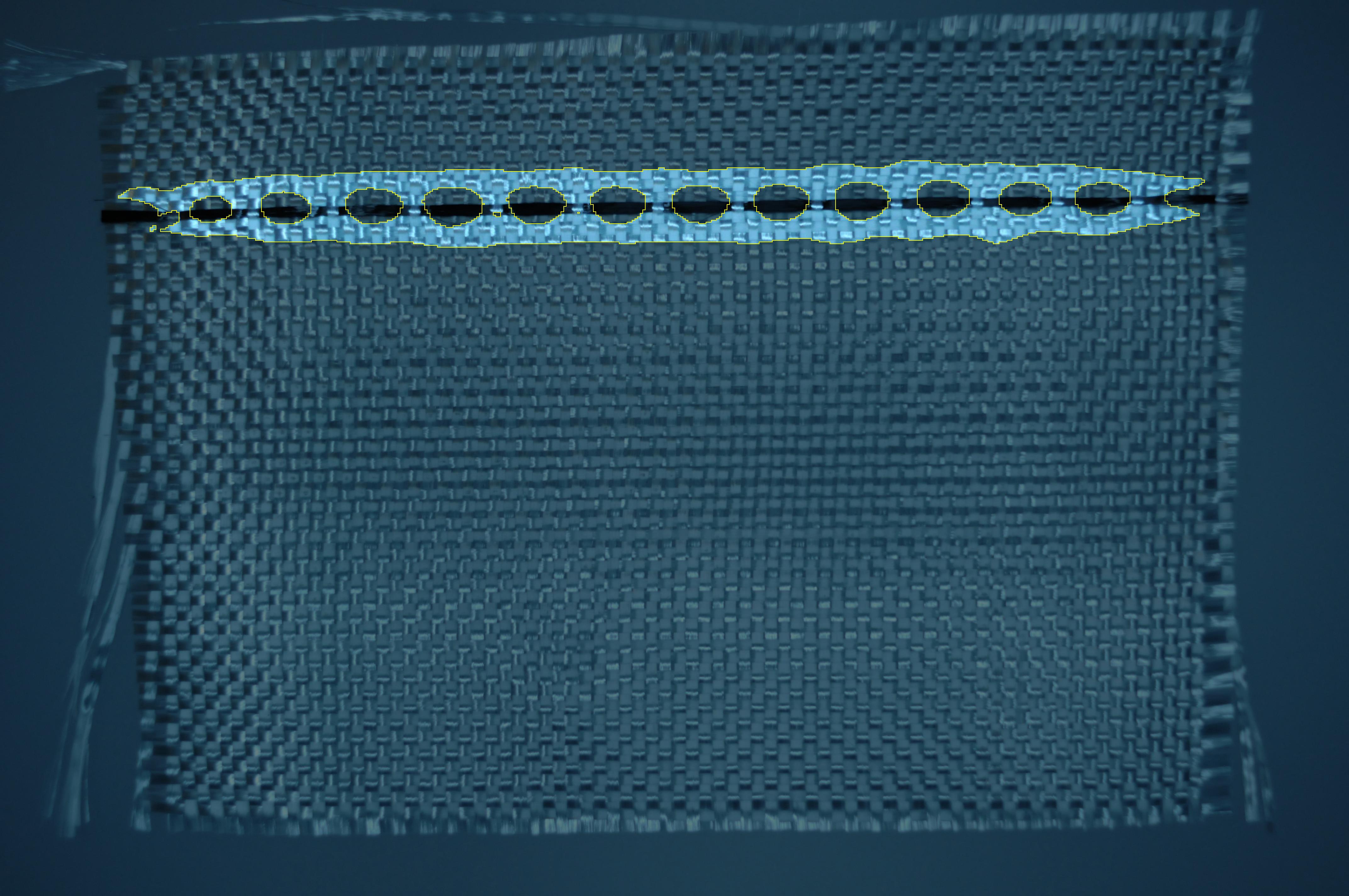}
        \includegraphics[width=0.24\linewidth]{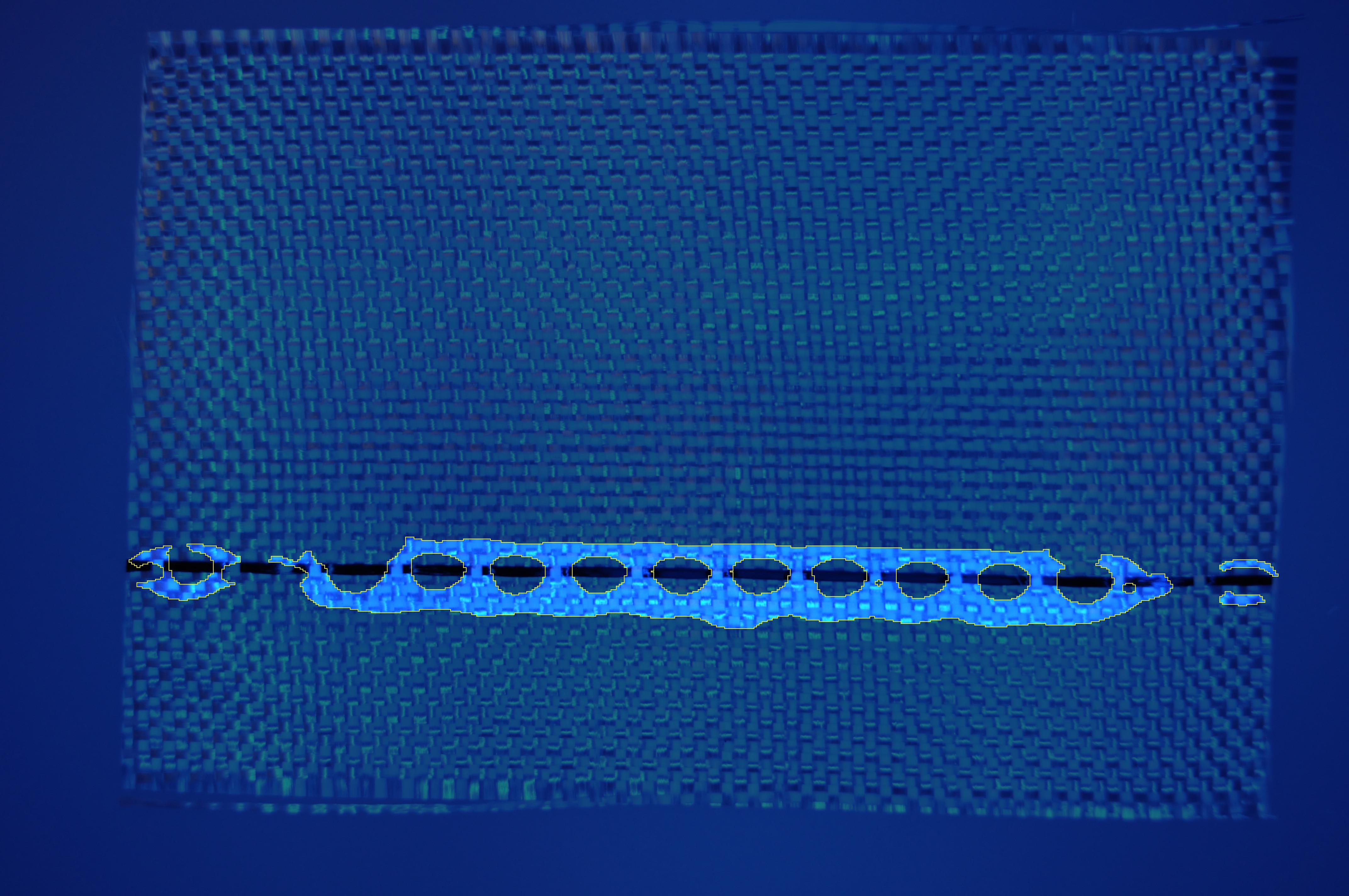}
        \includegraphics[width=0.24\linewidth]{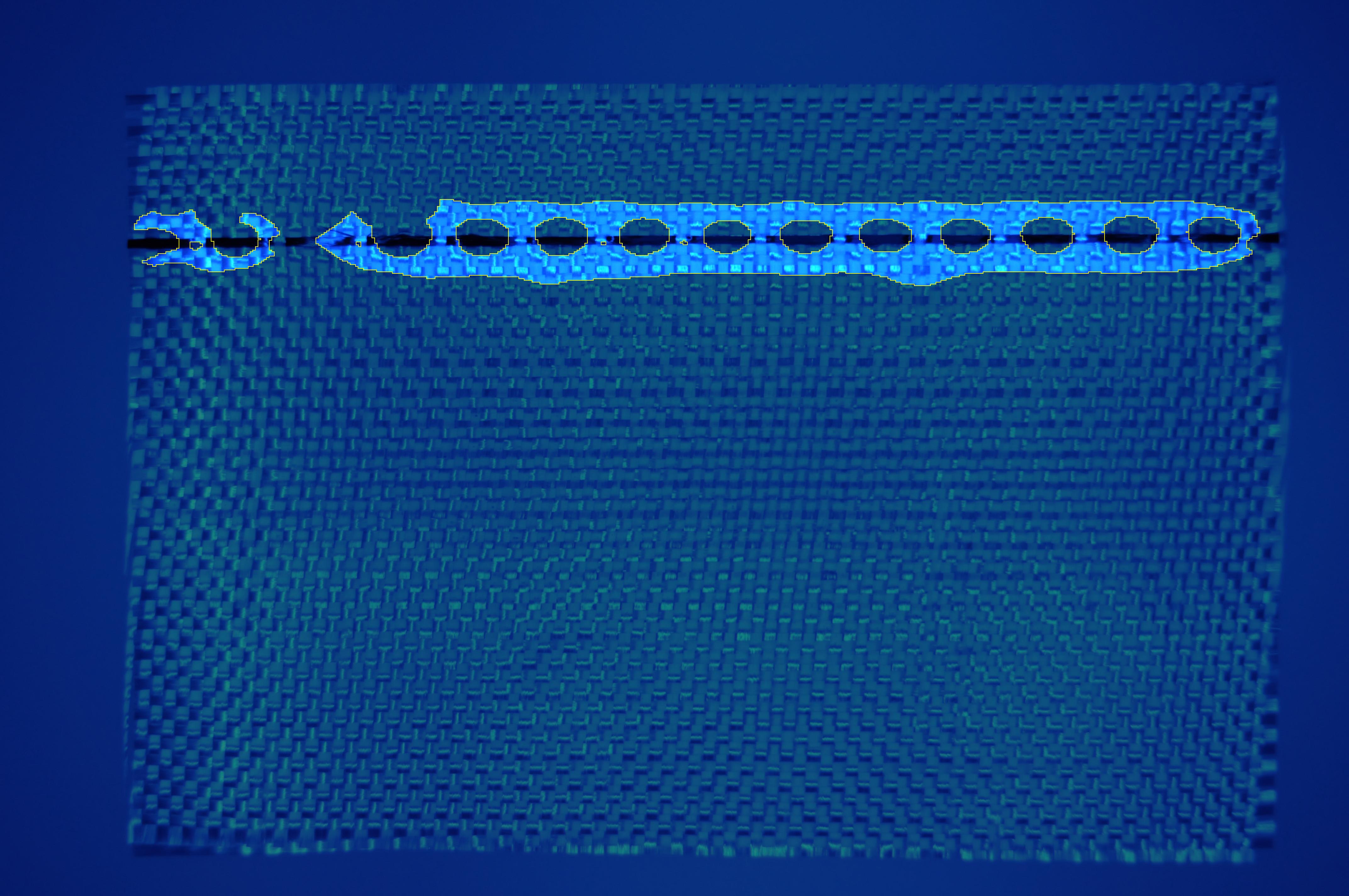}
        \caption{Concept 19, with a scaled importance score of 0.23.}
        \label{subfig:TT_C19}
    \end{subfigure}
    
    \caption{Sample images of top extracted concepts. All five top concepts related to the visual cues used by human experts. Concepts 13 in Subfigures \ref{subfig:TT_C13} related to punctures present in class 2. Concept 12 in Subfigure \ref{subfig:TT_C12} refer to thread breakages from class 4. Concepts 15, 17, and 19 in Subfigures \ref{subfig:TT_C15}, \ref{subfig:TT_C17}, and \ref{subfig:TT_C19} relate to the visual cues of missing thread bindings, 11 mm bindings, or 22 mm bindings respectively.
    }
    \label{fig:TT_C}
\end{figure}

Thus, to verify if the model was working as intended, we executed our concept extraction method to generate a global explanation of its prediction process. Our algorithm was executed using mini batches of 8 images to obtain 20 concepts.
Afterward, the resulting concepts were inspected, taking into account their importance score, and their example sets.
The top five concepts with the highest importance scores are shown in Figure \ref{fig:TT_C}, other concepts represent different subregions of the images with importance scores lower than 0.01, thus deemed as unimportant.
Subsequently, we provided a human expert with the obtained concepts, and analyzed whether they coincided with the visual cues he used during an inspection process.

The main observation found from inspecting the concepts was that they corresponded with the visual cues that an expert would use for performing the quality control task.
Two concepts (punctures \ref{subfig:TT_C13} and thread breakage \ref{subfig:TT_C12}) related directly to the key visual cues of classes 2 and 4 respectively.
Although the other visual cues are entangled between multiple classes, the model learned their concepts accordingly.
Specifically, the visual cues in the different ``binding'' classes, are shared. This is, the same binding thread appears in all three classes, yet, what changes is the spacing or the lack of binding threads. The extracted concepts, directly relate to missing thread bindings (concept 15), binding threads with a spacing of 11 mm (concept 17), and binding threads with a spacing of 22 mm (concept 19).
The resulting concepts, provide evidence to the human experts that the model performs the task using the intended regions. 
Moreover, it also learned which specific features appeared across classes and disentangled them in distinctive representations.

\begin{figure}[!htb]
    \centering
    \begin{subfigure}{0.49\textwidth}
        \centering
        \includegraphics[width=0.49\linewidth]{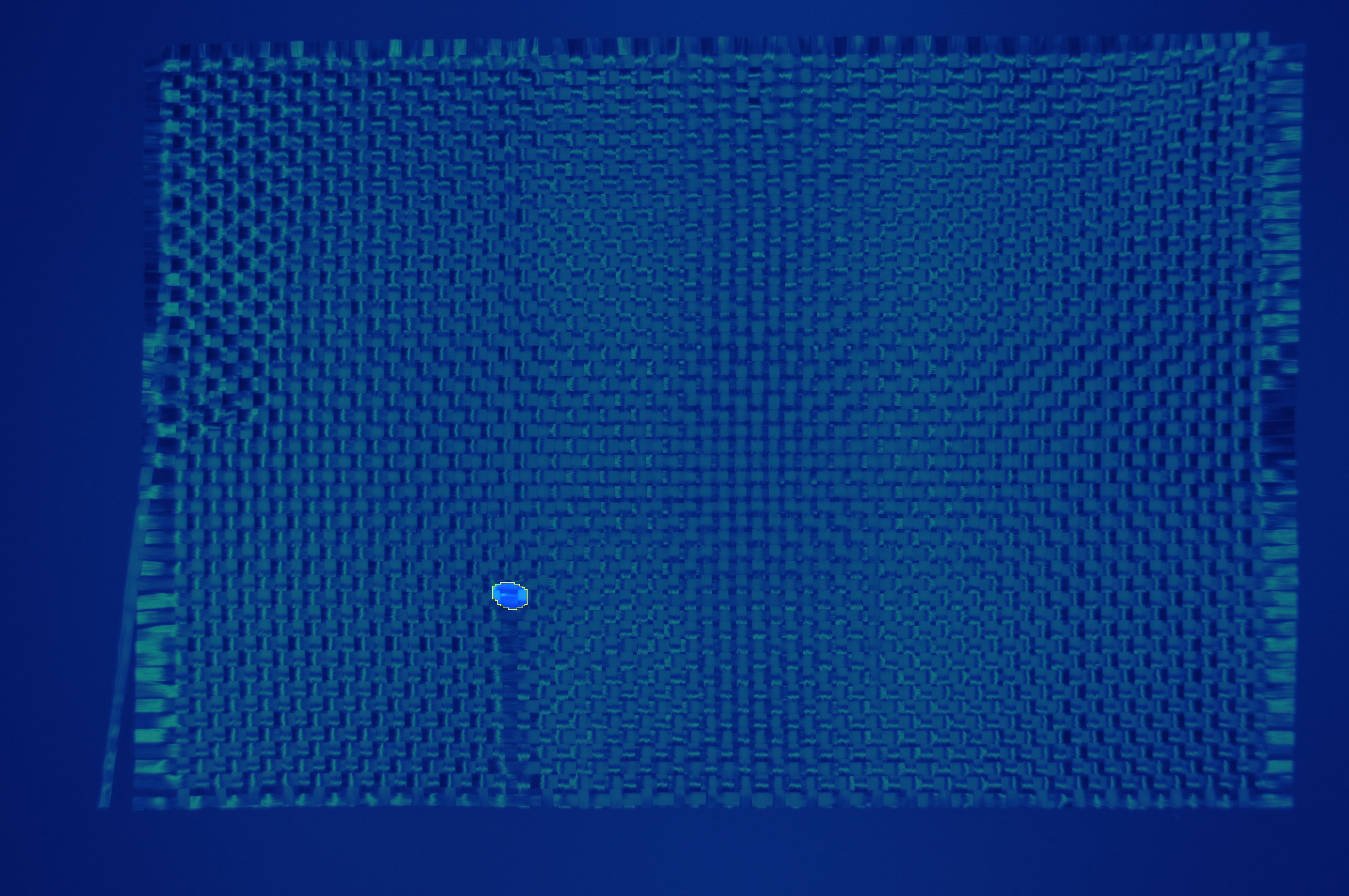}
        \includegraphics[width=0.49\linewidth]{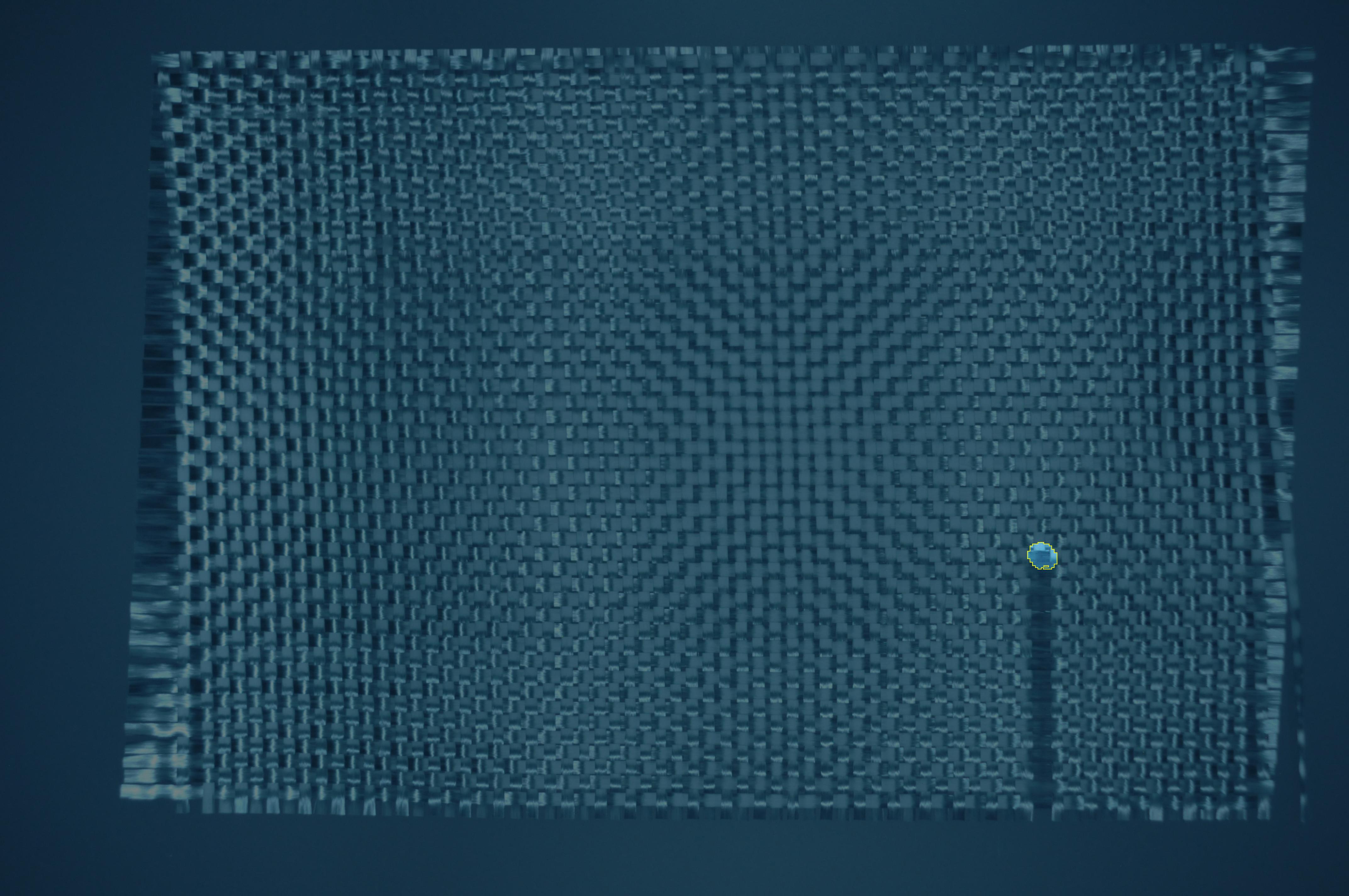}
        \caption{mixed samples on class 2}
        \label{subfig:TT_M13}
    \end{subfigure}
    \begin{subfigure}{0.49\textwidth}
        \centering
        \includegraphics[width=0.49\linewidth]{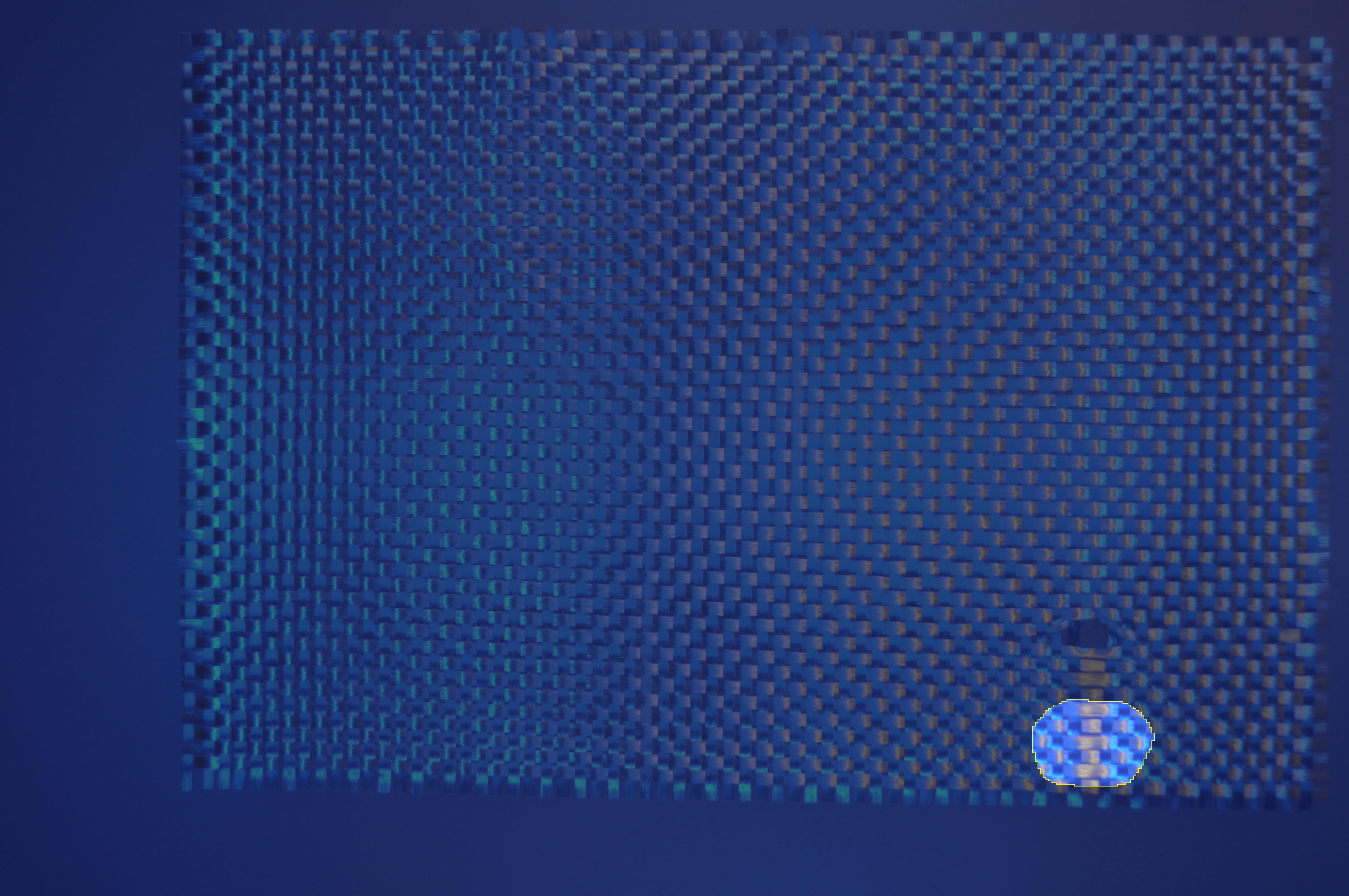}
        \includegraphics[width=0.49\linewidth]{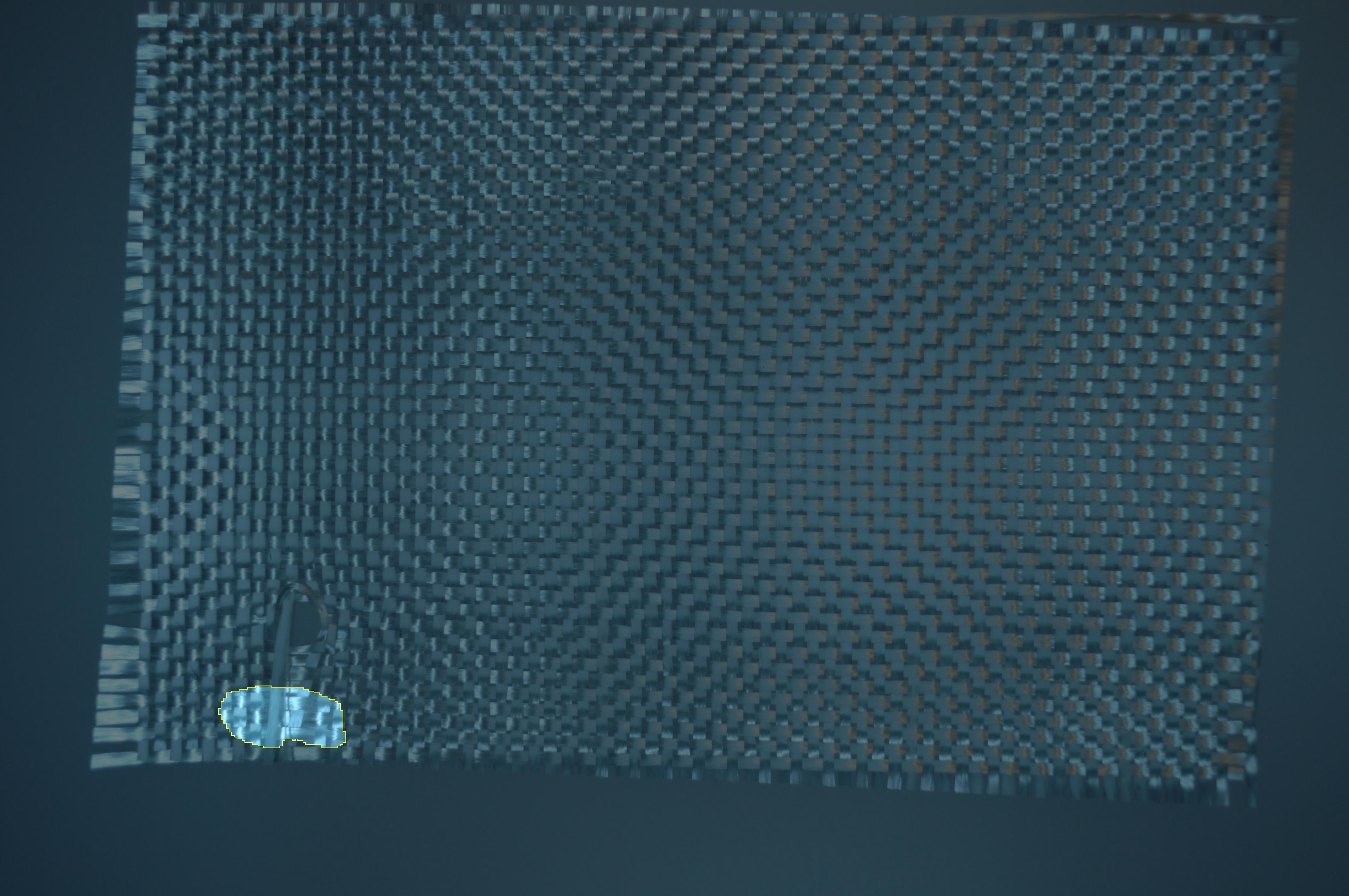}
        \caption{mixed samples on class 0}
        \label{subfig:TT_M12}
    \end{subfigure}

    \caption{Sample images from concepts found in non-related classes. Subfigure \ref{subfig:TT_M13} shows detections of concept 13 associated with punctures, but appearing in thread breakage images. Conversely, Subfigure \ref{subfig:TT_M12} shows detections of concept 12 related to thread breakages appearing adjacent to punctures.
    } 
    \label{fig:TT_M}
\end{figure}

In addition to generating a better understanding of what the model learned, and where does it react to the different patterns, CE techniques also allow for a better understanding of the original dataset.
Within the example sets of multiple concepts, mixed samples were observed. This is, concepts related to the visual feature of a single class, were detected in other images.
As shown in Figure \ref{fig:TT_M}, regions that the model encode as punctures were detected at the beginning of thread breakages. Similarly, regions that the model encodes as thread breakages were detected adjacent to puncture.
From a practical perspective, this confirms general knowledge that thread breakages can be caused by punctures, and thus, both visual cues can indeed appear together.
Nonetheless, as a result of this observation, experts can further discuss edge cases in the dataset, ensuring that they are labelled correctly.

Concept based explanations, provided a medium for experts to understand the prediction process of the model, increasing their trust.
Additionally, the visual evaluation of models, also highlighted that they could be used as a source of knowledge for better decision-making during production.

\subsection{Carbon Fiber Quality Control} \label{subsec:Carbon Fiber Quality Control}

This section presets a use case related to carbon fiber reinforcement quality control.
The carbon fiber (CF) use case relates to a quality control process posterior to a reinforcement procedure.
In general, carbon fiber textiles are a popular reinforcing material for plastics, due to its high stiffness, which makes it possible to craft a lightweight, yet powerful composite. 
The main application of fibre-reinforced composites are safety-critical domains like aviation or mobility.
In these domains, it is crucial to protect passengers' lives at all costs, which is why high quality levels are demanded per legislation.

Quality control of textiles and textile reinforced plastics is mainly executed in a manual manner, as existing intelligent quality control systems often fail silently when faced with changing errors appearances or different textile structures. 
This results in a tedious and tiresome visual inspection for workers in quality control, leading to exhaustion and increasing error proneness over time worked.
In this context, it is imperative for any trained CNN, to not only make accurate predictions, but also to make them for the right reasons.
Moreover, to increase trust from human experts, it is a necessity to have a better understanding of a model's capabilities and limitations in a more detailed manner.


\begin{figure}[!htb]
    \centering
    \begin{subfigure}{\textwidth}
        \centering
        \includegraphics[width=0.24\linewidth]{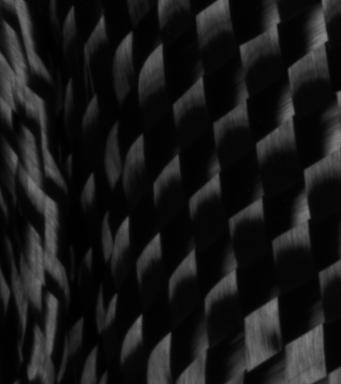}
        \includegraphics[width=0.24\linewidth]{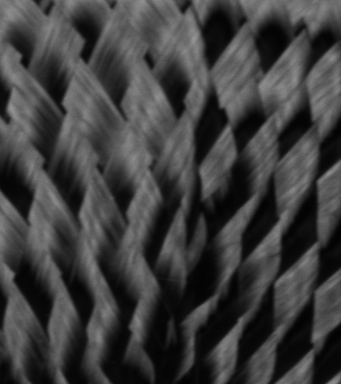}
        \includegraphics[width=0.24\linewidth]{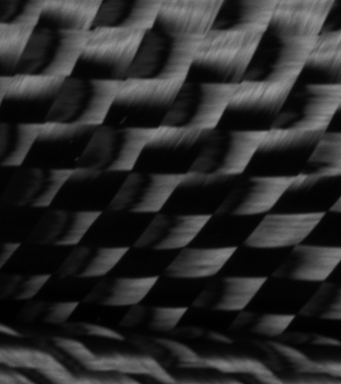}
        \includegraphics[width=0.24\linewidth]{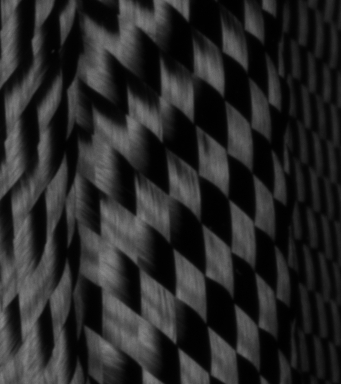}
        \caption{sample images from class 0: folded textile.}
        \label{subfig:CF_0}
    \end{subfigure}

    \begin{subfigure}{\textwidth}
        \centering
        \includegraphics[width=0.24\linewidth]{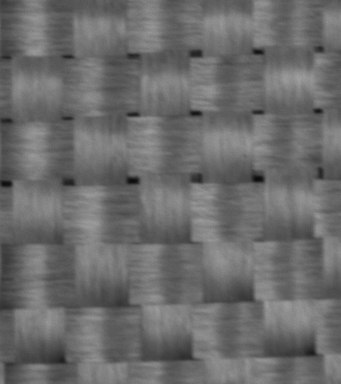}
        \includegraphics[width=0.24\linewidth]{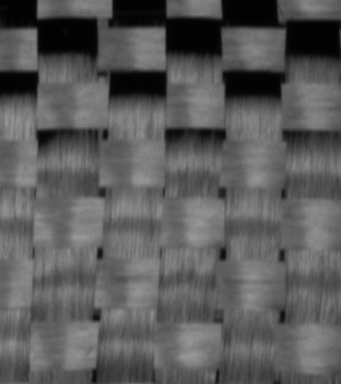}
        \includegraphics[width=0.24\linewidth]{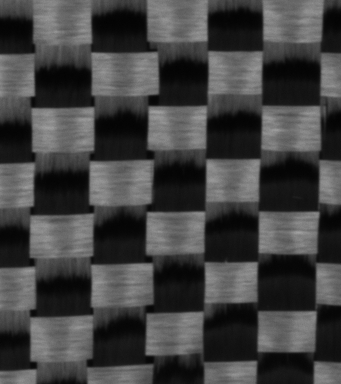}
        \includegraphics[width=0.24\linewidth]{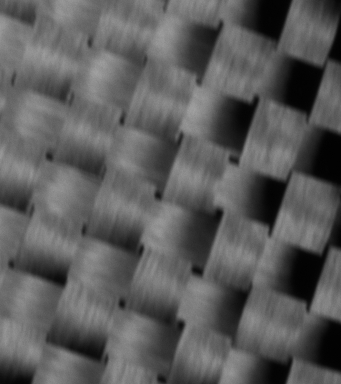}
        \caption{sample images from class 1: textile with gaps.}
        \label{subfig:CF_1}
    \end{subfigure}

    \begin{subfigure}{\textwidth}
        \centering
        \includegraphics[width=0.24\linewidth]{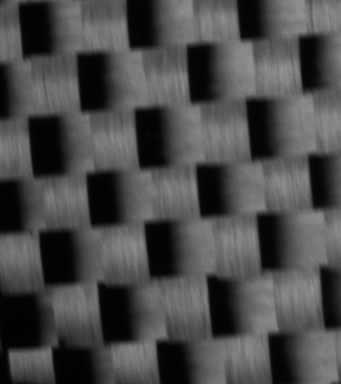}
        \includegraphics[width=0.24\linewidth]{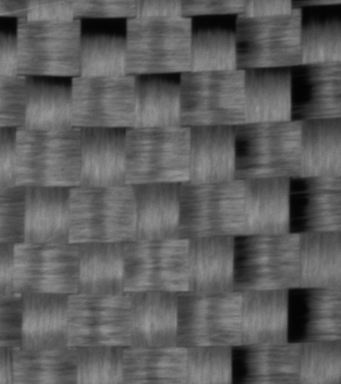}
        \includegraphics[width=0.24\linewidth]{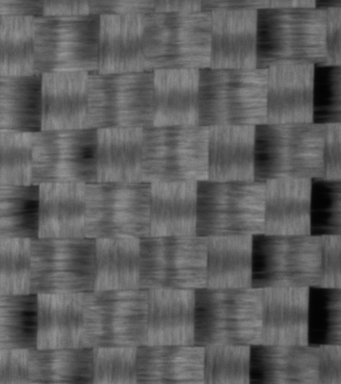}
        \includegraphics[width=0.24\linewidth]{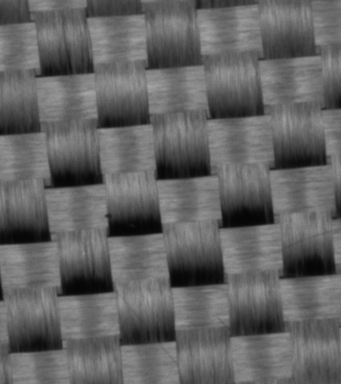}
        \caption{sample images from class 2: normal textile}
        \label{subfig:CF_2}
    \end{subfigure}

    \caption{Carbon fiber quality control dataset. The images on Figure \ref{subfig:CF_0}, depict patches of a textile with folded regions. Similarly, the images on Figure \ref{subfig:CF_1}, contain patches where gaps exist between the textile threads. Finally, the images on Figure \ref{subfig:CF_2}, represent normal patches of textile. Both folded regions and gaps have adverse impact in the mechanical properties of a final product.}
    \label{fig:CF}
\end{figure}

To represent a practical quality control problem of carbon fiber reinforcement, a dataset was generated. The dataset was obtained using a commonly used carbon plain weave with a grammage of $200\,g/m^{2}$. 
First, the weave was cut to $300\,\times \,300\, mm^{2}$ samples using a CNC cutter table. Then, two sample sheets were placed on top of each other to form a stack.
Next, in between the two textile layers, a binder (EPIKOTE EP 05311 from Hexion Inc.) with a grammage of $5\,g/m^{2}$ was applied to maintain shape stability after the three-dimensional forming process. The forming process was executed on an asymmetric form set (positive and negative) mounted on a press, shaping 25 stacks three-dimensionally.
Due to its asymmetric characteristic (curvatures and edges), the form was prone to different types of errors. 
Subsequently, using a high resolution camera (Apodius HP-C-V3D) mounted on a robotic arm (Hexagon ROMER Absolute Arm), 500 images with a resolution of 2048$\times$1536 pixels, were taken of the textiles' surfaces in three-dimensional shape. 
Afterwards, the surface scan images were cropped to 341$\times$384 pixels patches and transformed to grayscale.
Each patch was then labeled according to one of three classes, normal textile, gap, or fold, as seen in Figure \ref{fig:CF}.
Finally, the patches were manually inspected and images that were blurred, out of focus, or had bad contrast were sorted out. 
\footnote{The dataset is available under the link: \url{https://doi.org/10.5281/zenodo.7970490}.}


Similar to the previous use case, the quality control process was formulated as an image classification problem. A Densenet121 model \cite{DBLP:conf/cvpr/HuangLMW17} was used and trained until convergence with a train-test split of 0.8. The training procedure was maintained, using a learning rate of 0.1 and a reduce on plateau scheduler.
Given the original size of the patches, the images were resized to 224$\times$224 pixels, applying random rotations and photometric data augmentations.
After training, the model achieved perfect accuracy in the testing set. Which, given the complexity of the task, raises the question of whether such minimal visual cues as gaps were learned by the model, or other biases such as the average intensity of the pixels could be used to perform the classification.

\begin{figure}[!htb]
    \centering
    \begin{subfigure}{0.49\textwidth}
        \centering
        \includegraphics[width=0.49\linewidth]{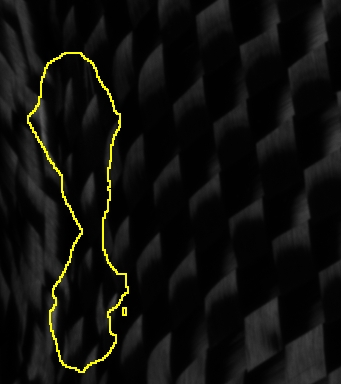}
        \includegraphics[width=0.49\linewidth]{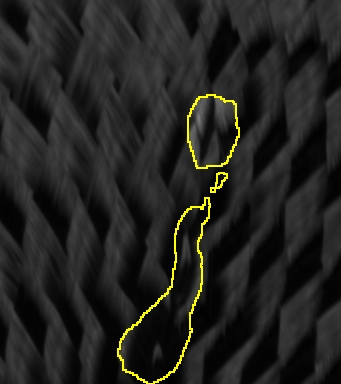}
        \caption{Concept 5, with a scaled importance score of 1.0.}
        \label{subfig:CF_C5}
    \end{subfigure}
    \begin{subfigure}{0.49\textwidth}
        \centering
        \includegraphics[width=0.49\linewidth]{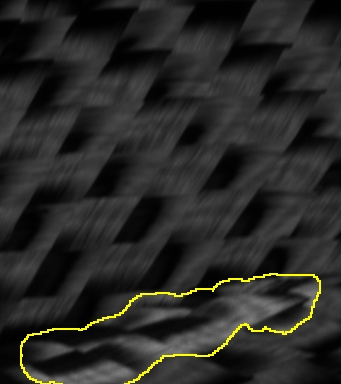}
        \includegraphics[width=0.49\linewidth]{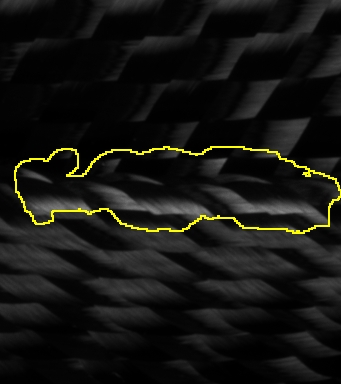}
        \caption{Concept 6, with a scaled importance score of 0.55.}
        \label{subfig:CF_C6}
    \end{subfigure}

    \begin{subfigure}{\textwidth}
        \centering
        \includegraphics[width=0.24\linewidth]{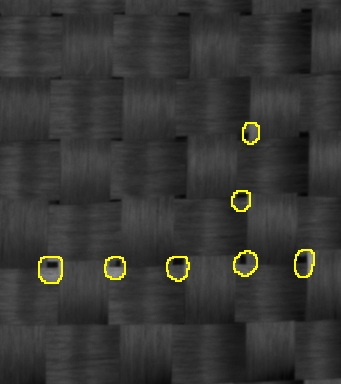}
        \includegraphics[width=0.24\linewidth]{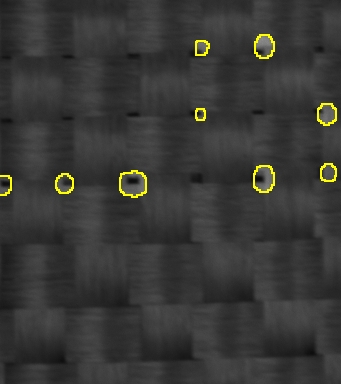}
        \includegraphics[width=0.24\linewidth]{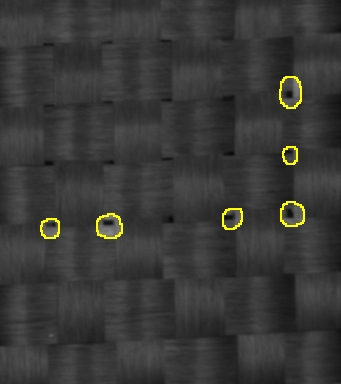}
        \includegraphics[width=0.24\linewidth]{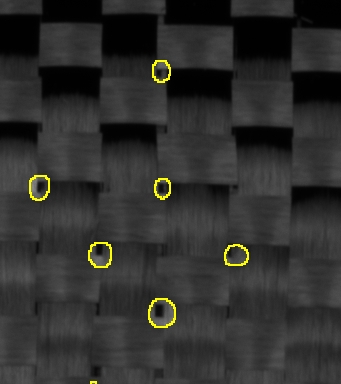}
        \caption{Concept 8, with a scaled importance score of 0.14.}
        \label{subfig:CF_C8}
    \end{subfigure}

    \caption{Top concepts extracted from Densenet121, trained on the carbon fiber quality control dataset. From the top concepts, the two most important ones (concept 5 and 6), relate to the visual cues present in folded regions of the textiles. In addition, concept 8, refers to gaps, or the darker regions present in the intersection point between fibers. Other unrelated concepts had scaled importance scores lower than 0.03. Indicating that visual cues learned by the CNN coincide with those used by a human expert while performing the visual quality control.}
    \label{fig:CF_C}
\end{figure}

Thus, our concept extraction method was used to provide human experts with the means for better understanding the prediction process of the model.
Our method was executed to mine for 10 concepts using mini-batches of 8 images.
Afterwards, the example sets of the concepts and their respective importance scores were presented to the human expert.
The three concepts with the highest importance scores are shown in Figure \ref{fig:CF_C}, where the two most important concepts (5 and 6) relate to folded regions of the textiles in different directions, and the third one (8) relates to gaps in the textiles. Other concepts had importance scores below 0.3, with a minimal contribution towards the prediction process of the model.

As a key finding, the obtained concepts aligned with the visual cues that human experts would use to perform a similar task.
Particularly, it was possible for human experts to verify that the gap regions were indeed used as a key visual cue for differentiating the types of errors.
Moreover, although the model has distinctive representations for folded regions on different angles, it was possible to verify that what mattered to the model was indeed the center of the fold, and not the intensity of the region itself.
This provided insights into the generalization capabilities of the model, increasing experts trust.

Furthermore, the capabilities of the CE technique to localize concepts in new samples, provide the experts with knowledge regarding the correctness of the predictions.
This is specially important in cases where textiles differ from the original dataset, or when a specific prediction is challenged.



\begin{figure}[!htb]
    \centering
    \begin{subfigure}{\textwidth}
        \centering
        \includegraphics[width=0.24\linewidth]{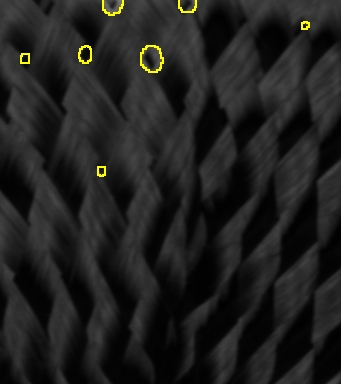}
        \includegraphics[width=0.24\linewidth]{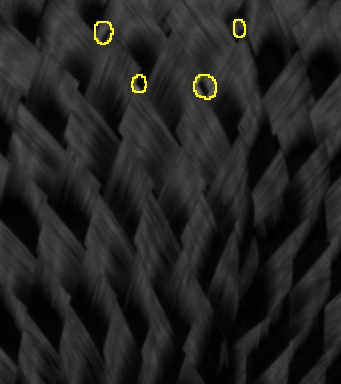}
        \includegraphics[width=0.24\linewidth]{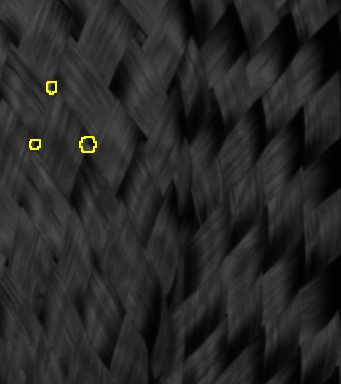}
        \includegraphics[width=0.24\linewidth]{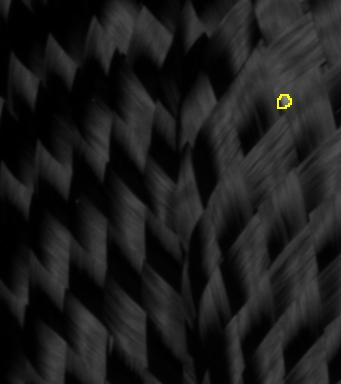}
        \caption{Detections of concept 8 related to gaps, in images from class 0 labelled as folded.}
        \label{subfig:CF_M80}
    \end{subfigure}

    \begin{subfigure}{\textwidth}
        \centering
        \includegraphics[width=0.24\linewidth]{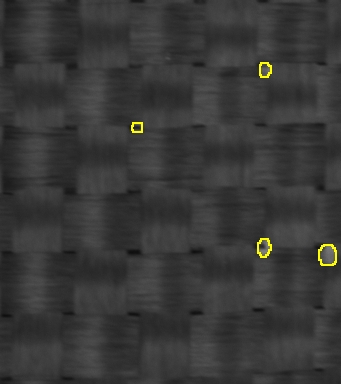}
        \includegraphics[width=0.24\linewidth]{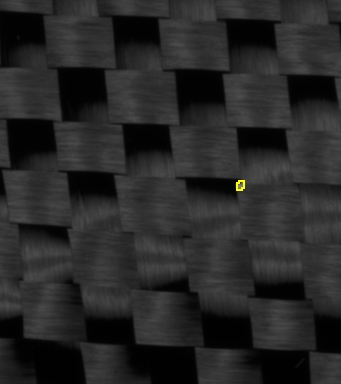}
        \includegraphics[width=0.24\linewidth]{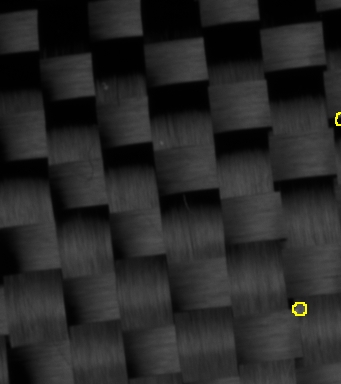}
        \includegraphics[width=0.24\linewidth]{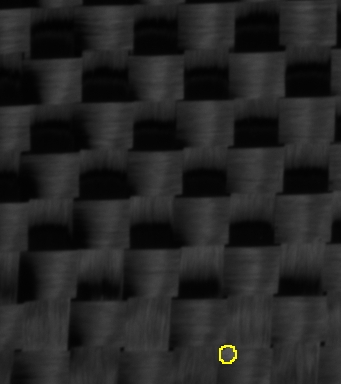}
        \caption{Detections of concept 8 related to gaps, in images from class 2 labelled as normal.}
        \label{subfig:CF_M82}
    \end{subfigure}

    \caption{Samples off cross appearance of visual cues. These images depict edge cases on both folded and normal classes, where visual cues related to gaps (concept 8) are detected. These edge cases provide valuable feedback for experts to decide if any mislabelling has ocurred.}
    \label{fig:CF_M}
\end{figure}

Similar to before, the analysis provides insights not only on the decision process of the model, but also on edge cases found in the dataset.
In this case, regions related to concept 8 (associated with gaps) were found in samples of the other classes, seen in Figure \ref{fig:CF_M}.
In one case, textiles which were bent on specific ways, forced some gaps while deforming the textile threads. This confirmed that, although the visual cues of folded textiles are given priority by the model, it still detects the regions containing gaps.
In other cases, it was detected that some images initially labelled as normal, contained gaps of different sizes.
This detections allow a dual refinement of both, models and dataset, while at the same time, allowing human experts to contribute in the trainng process of the models.



Finally, the obtained concepts, provide experts with a tool for understanding the model and other future predictions.
This proves valuable in cases where experts want models to adapt to other textiles (e.g. glass, aramide, basalt, flax), and want to validate whether the original visual cues are still detected for the new types of materials.



\subsection{Photovoltaic panels maintenance} \label{subsec:Photovoltaic panels maintenance}

The final use case relates to the inspection processes in photovoltaic (PV) power plants.
Within the context of the industry 4.0, efforts in pursuing low-carbon energy have boosted recent adoption of large scale photovoltaic (PV) systems \cite{deitsch2019automatic}. 
Nonetheless, the adoption of PV systems can be complex, as their viability highly depends on a robust and scalable monitoring of the photovoltaic modules \cite{li2017visible}. These solar modules have multiple physical safeguards such as a glass lamination and metal frames. Yet, the modules are still vulnerable to mechanical damages, obstructions, as well as installation and manufacturing errors \cite{li2017visible}. 
In an effort to implement large scale inspection of PV systems, recent approaches have used unmanned aerial vehicles (UAV) for the acquisition of high resolution images of the installed solar modules. 
In this context, deep learning visual inspection systems provide an alternative for the large scale monitoring of PV modules.
Nonetheless, Visual identification of defects is still challenging, even for experts \cite{li2017visible,deitsch2019automatic}.

\begin{figure}[!htb]
    \centering
    \begin{subfigure}{0.49\textwidth}
        \centering
        \includegraphics[width=0.49\linewidth]{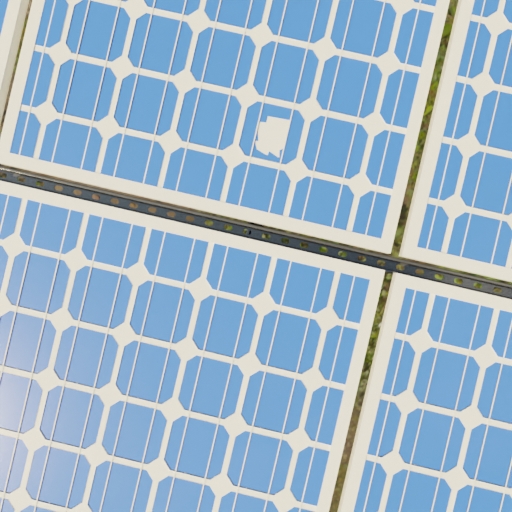}
        \includegraphics[width=0.49\linewidth]{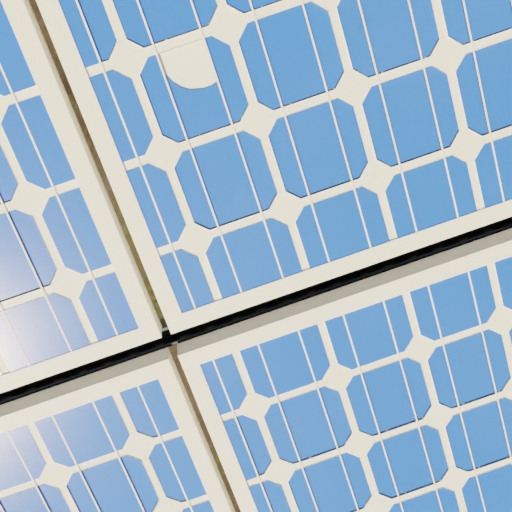}
        \caption{Sample images from class 0: containing broken panels.}
        \label{subfig:PV_0}
    \end{subfigure}
    \begin{subfigure}{0.49\textwidth}
        \centering
        \includegraphics[width=0.49\linewidth]{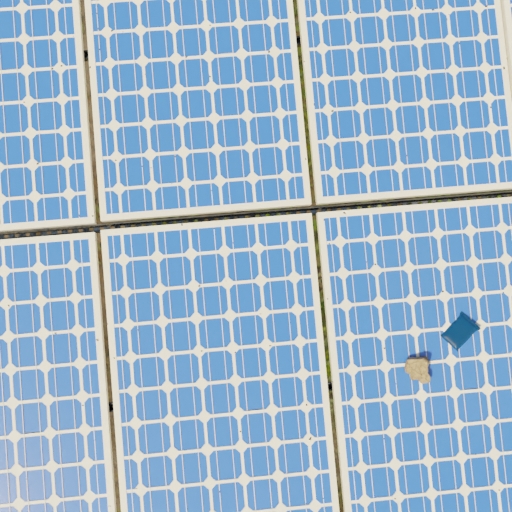}
        \includegraphics[width=0.49\linewidth]{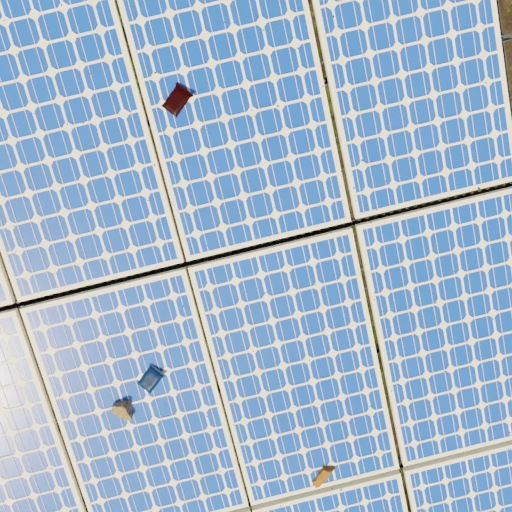}
        \caption{Sample images from class 1: containing foreign objects over the panels.}
        \label{subfig:PV_1}
    \end{subfigure}

    \begin{subfigure}{0.49\textwidth}
        \centering
        \includegraphics[width=0.49\linewidth]{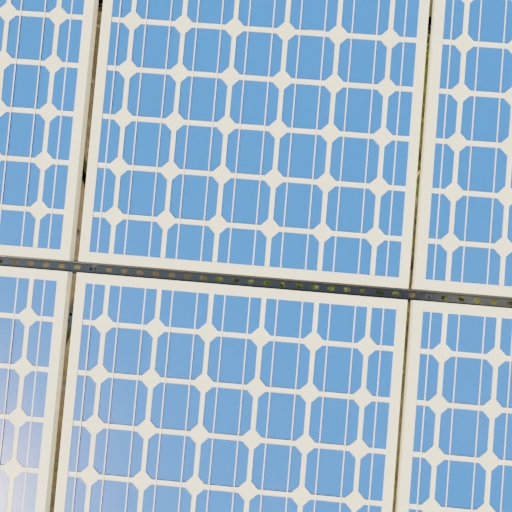}
        \includegraphics[width=0.49\linewidth]{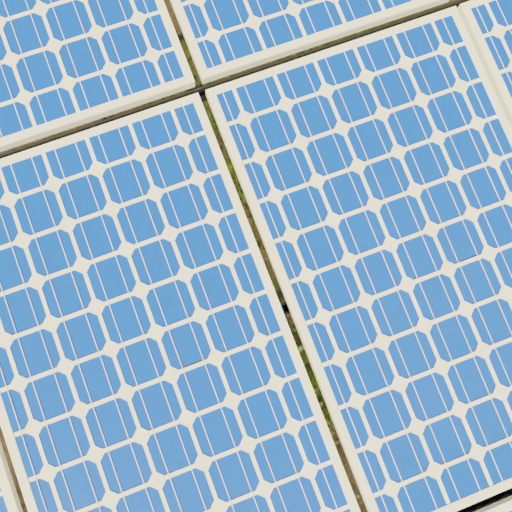}
        \caption{Sample images from class 2: containing normal panels.}
        \label{subfig:PV_2}
    \end{subfigure}
    \begin{subfigure}{0.49\textwidth}
        \centering
        \includegraphics[width=0.49\linewidth]{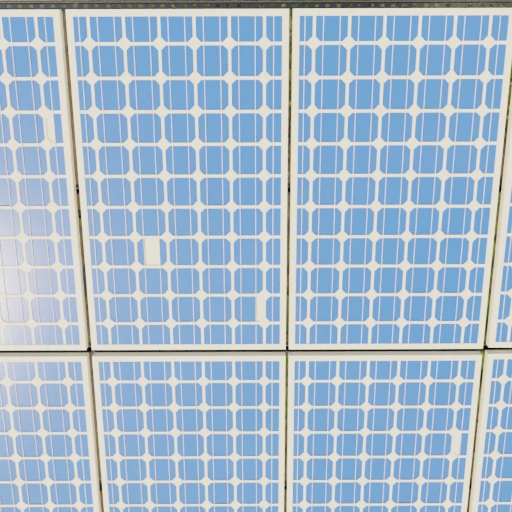}
        \includegraphics[width=0.49\linewidth]{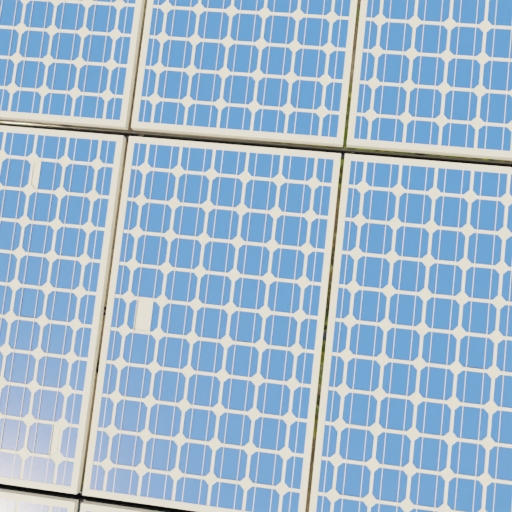}
        \caption{Sample images from class 3: containing panels with complete missing sections.}
        \label{subfig:PV_3}
    \end{subfigure}

    \caption{Photovoltaic module dataset, depicting images of solar panels, emulating an automatic inspection in a large scale panel system \cite{DBLP:journals/corr/abs-2211-13968}. The images in Figures \ref{subfig:PV_0} and \ref{subfig:PV_3} contain panels where one of the subsections is partially broken or fully missing. The images from Figure \ref{subfig:PV_1}, depicts images where foreign objects have landed over the panels. Finally, the images on Figure \ref{subfig:PV_2} contain normal solar panels without any defect.}
    \label{fig:PV}
\end{figure}

In this context, we represent a practical case using the dataset provided by Bao et al. \cite{DBLP:journals/corr/abs-2211-13968}, mimicking a real world photovoltaic power plant, where images containing multiple types of defects were provided. 
As seen in Figure \ref{fig:PV}, the used dataset contains PV panels images which are either defect free, contain broken panels, foreign objects, or missing panels. The dataset is composed of 5000 images from different perspectives of 512$\times$512 pixels.
This dataset, represents the common case of having to inspect the solar panels installed in a large scale PV system though a set of images obtained using UAVs. 
It is common for these inspections to be performed manually, as the classification of which panels contain defects can be challenging, and the predictions from automated models can be unreliable.
In this context, it is critical to understand not only the capabilities of any used model, but also to have more information on why a prediction was made, and where was an issue detect.
This becomes a first step towards understanding why a model is performing unreliably, and towards making the necessary changes to retrain a working system.

This task of visual inspection was formulated as an image classification problem. A similar architecture as before was used (Densenet121 \cite{DBLP:conf/cvpr/HuangLMW17}), following the same training process as the previous use cases. The images were resized to 224$\times$224 pixels, and random rotations as well as color jitter were added as data augmentations.
This procedure highlights the flexibility of CNN techniques, and their power to learn patterns from different types of data.
Regardless, it also highlights the importance of understanding why predictions are being made, as a high accuracy score does not necessary mean that the model is performing the task as intended.

\begin{figure}[!htb]
    \centering
    \begin{subfigure}{\textwidth}
        \centering
        \includegraphics[width=0.24\linewidth]{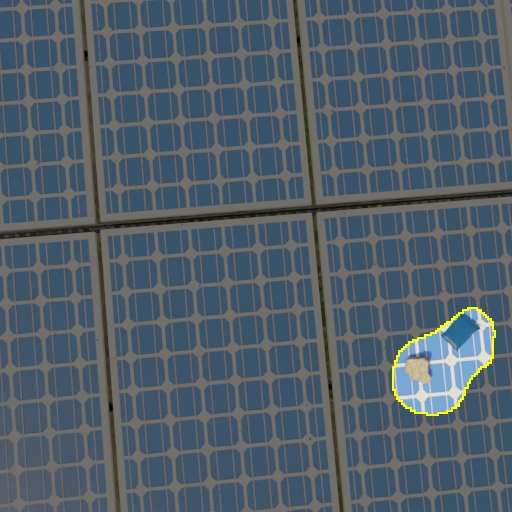}
        \includegraphics[width=0.24\linewidth]{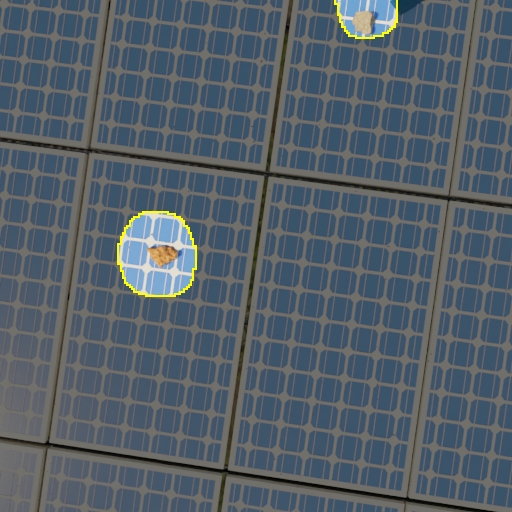}
        \includegraphics[width=0.24\linewidth]{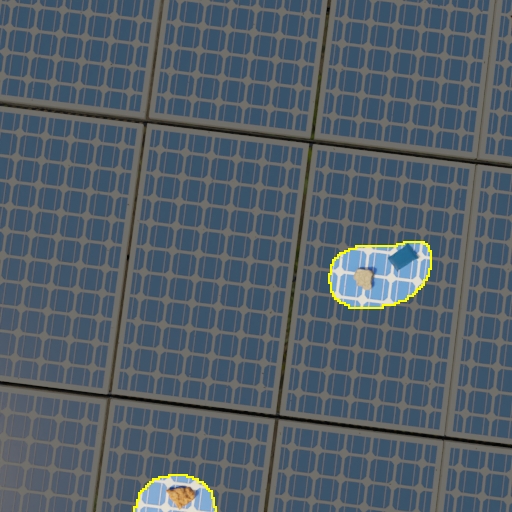}
        \includegraphics[width=0.24\linewidth]{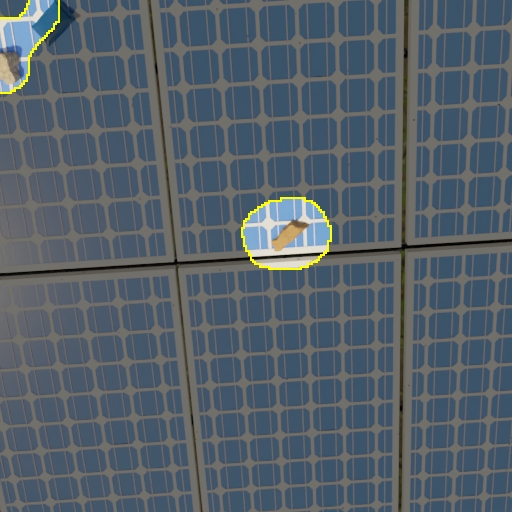}
        \caption{Concept 7, with a scaled importance score of 1.00.}
        \label{subfig:PV_C7}
    \end{subfigure}

    \begin{subfigure}{\textwidth}
        \centering
        \includegraphics[width=0.24\linewidth]{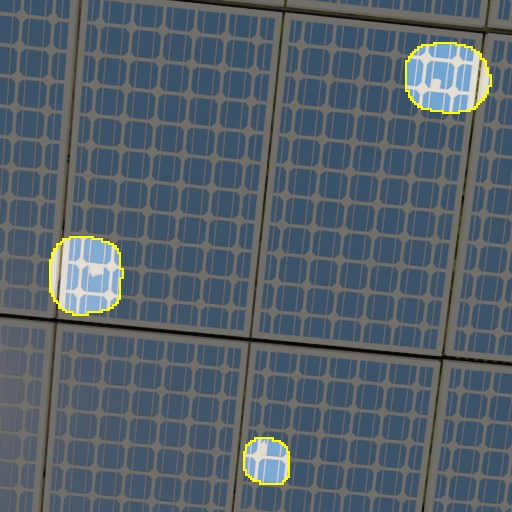}
        \includegraphics[width=0.24\linewidth]{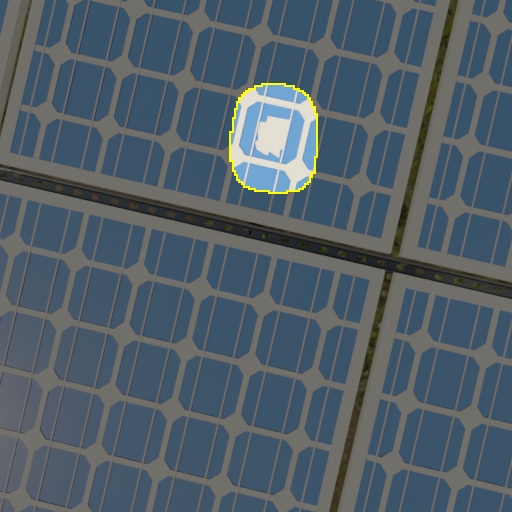}
        \includegraphics[width=0.24\linewidth]{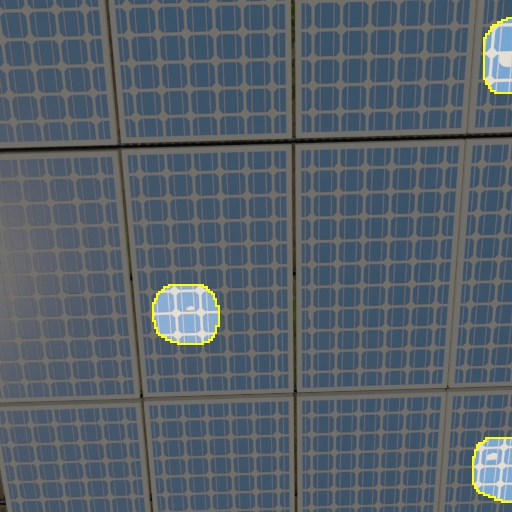}
        \includegraphics[width=0.24\linewidth]{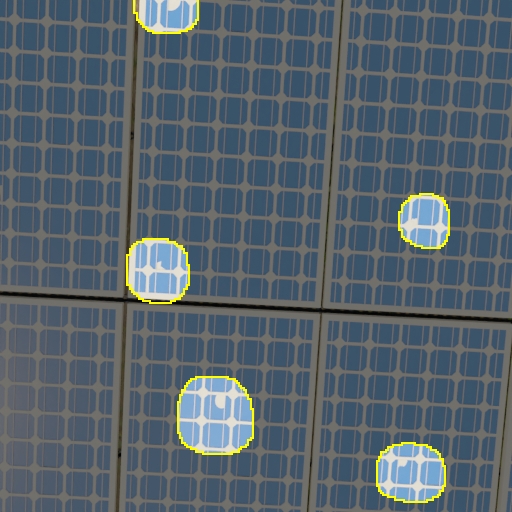}
        \caption{Concept 6, with a scaled importance score of 0.42.}
        \label{subfig:PV_C6}
    \end{subfigure}

    \begin{subfigure}{\textwidth}
        \centering
        \includegraphics[width=0.24\linewidth]{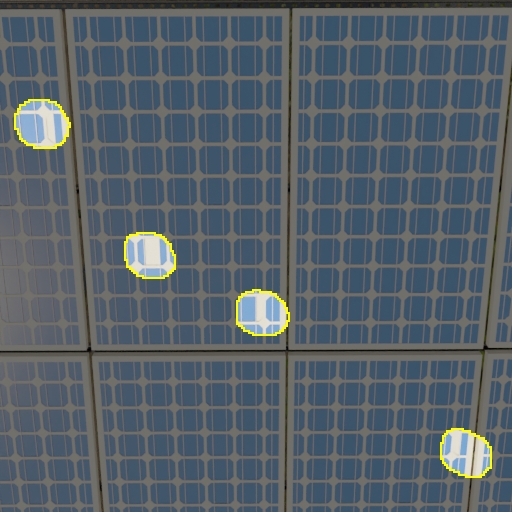}
        \includegraphics[width=0.24\linewidth]{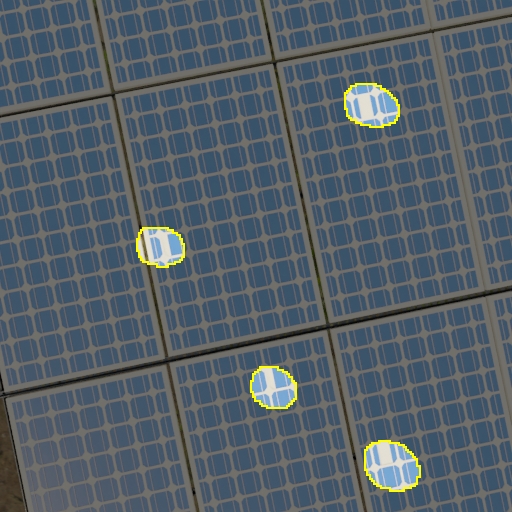}
        \includegraphics[width=0.24\linewidth]{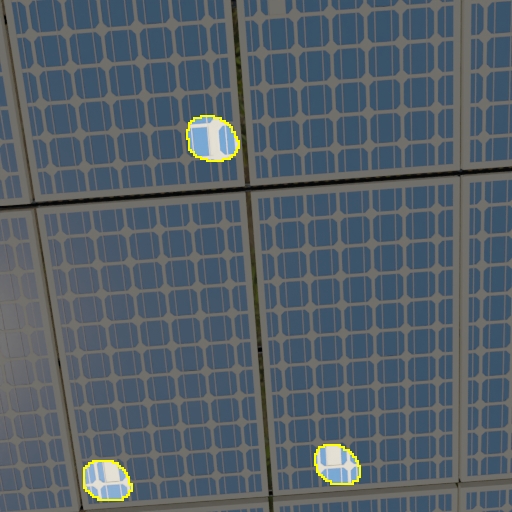}
        \includegraphics[width=0.24\linewidth]{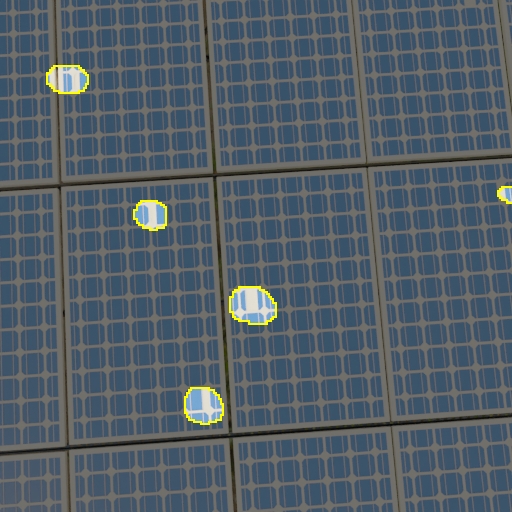}
        \caption{Concept 9, with a scaled importance score of 0.28.}
        \label{subfig:PV_C9}
    \end{subfigure}

    \caption{Top concepts extracted from a Densenet121, trained on the Photovoltaic module dataset. The three most important concepts are directly aligned with the labelled classes, containing either missing panels (concept 7), broken panels (concept 6), or the foreign objects (concept 9). Other concepts had scaled importance scores lower than 0.04. These explanations provide extra information complimenting the initial predictions, with an indication of what was detected and where.}
    \label{fig:PV_C}
\end{figure}

Thus, after training the model until convergence, we used our concept extraction technique for explaining the inner workings of the model.
The main objective for analyzing this model is not only to provide mediums for understanding the models' behavior, but also for highlighting, for specific predictions, where are the defects detected, facilitating human inspectors assets management.
In this regard, our concept extraction technique was executed to extract 10 clusters, using mini batches of 8 images. From this process, the three main concepts are shown in Figure \ref{fig:PV_C}, as other concepts were scored with importance lower than 0.04, contributing minimally to the prediction of any class.
In highlight, the resulting concepts align directly with the main visual cues relate to each class, giving a basic confirmation that the image classification task was solved using the desired visual cues.

In more detail, the resulting explanation assured the human expert, that the network was able to differentiate foreign objects, broken panels, and missing panels. As seen in Figure \ref{fig:PV_C}, not only was the network capable of classifying each class correctly, but internally, within its latent space, it has the capabilities of a coarse segmentation of defects from raw images.
This is specially significant, given the similarities between broken and missing panels, where the difference lays in the ``completeness'' or not of the missing section.
Thus, though the concept extraction process, human experts can have an assurance that, not only is the model performing the detections as they intend, but that for any new images, the prediction can be extended to localize where the defects are. This detection, is the results of the network abstraction, without the need of any extra label or the training of other object detection algorithms.  


As a final remark, global explanations allowed the better understanding of the three use cases, providing human experts with the key visual cues used in the models' prediction.
Moreover, the explanations allowed the processing of raw data to generate knowledge, by localizing concepts for any new prediction. 
\section{Conclusions} \label{sec:Conclusions}

This manuscript explores the usage of concept-based explanations in the context of the industry 4.0. To this end, we discuss the relation between the industry 4.0 and explainability techniques, we state the connection of explanations, digital shadows, explainability methods and digital twins. Then, we base our study on the concept extraction method ECLAD, modifying its concept scoring process and making it more scalable. Finally, we apply our CE method on industry related data and demonstrate the viability of its usage in this context.

First, we discuss the connection between concept based explanations and the framework of the Industry 4.0.
Specifically, that concept extraction methods can be part of digital twins, and the generated explanations are part of digital shadows.
More so, that high level concepts provide domain experts with means to assess alignment. Moreover, further concept localization can provide valuable insights during manufacturing operations.

Second, we modify the concept extraction method ECLAD to improve its scalability. We propose a new concept scoring procedure, based on a novel CNN wrapping function, and the aggregation of the gradients at the input of the model. With our method, we decrease the number of evaluations in the concept scoring phase by a factor of $n_k \times n_c$.

Third, we explore the application of our method in the context of tailored textiles, carbon fiber quality control, and maintenance of photovoltaic modules. In the three use cases, after training and explaining models, the obtained concepts aligned with the visual cues pertinent to each task.
We show how models trained to differentiate fiber glass errors, learn the visual cues of punctures, thread breakages, and binding issues. Similarly, models trained on carbon fiber errors, learn the folding regions of textiles, as well as abnormally big gaps. In addition, we show how models trained for detecting photovoltaic panel issues, learned the visual cues related to foreign objects, broken panels, and missing panels.
Through empirical results, we show that concept extraction methods provide usable explanations in the context of industrial data. These explanations can be used to better understand the prediction process of models and to ensure the models' alignment with expert knowledge.

The current work shows promising results on the usage of concept-based explanations for industrial applications, opening multiple possible research directions. 
On a practical perspective, it opens the opportunity for using high-level concepts, as understandable mediums in industrial applications for detecting domain shifts, anomalies, class differences, and in general, for an analysis of possible biases and data leakages present in datasets. 
On an abstract perspective, it hints towards a structured integration of explanation methods in industry 4.0 frameworks and processes. Finally, it opens a discussion about measuring the value of these explanations, by estimating the increase of trust generated, or the possibility of detecting different kinds of biases present in data. 

\section*{Acknowledgements} \label{sec:Acknowledgements}

Funded by the Deutsche Forschungsgemeinschaft (DFG, German Research Foundation) under Germany’s Excellence Strategy–EXC-2023 Internet of Production–390621612.

This preprint has not undergone peer review (when applicable) or any post-submission improvements or corrections. The Version of Record of this contribution is published in the proceedings of ``The World Conference on eXplainable Artificial Intelligence".
%
%
%
\newpage
\bibliographystyle{splncs04}
\bibliography{bibliography}

\begin{thebibliography}{10}
\providecommand{\url}[1]{\texttt{#1}}
\providecommand{\urlprefix}{URL }
\providecommand{\doi}[1]{https://doi.org/#1}

\bibitem{DBLP:conf/nips/AdebayoGMGHK18}
Adebayo, J., Gilmer, J., Muelly, M., Goodfellow, I.J., Hardt, M., Kim, B.:
  Sanity checks for saliency maps. In: Bengio, S., Wallach, H.M., Larochelle,
  H., Grauman, K., Cesa{-}Bianchi, N., Garnett, R. (eds.) NeurIPS. pp.
  9525--9536 (2018)

\bibitem{DBLP:journals/tii/AhmedJP22}
Ahmed, I., Jeon, G., Piccialli, F.: From artificial intelligence to explainable
  artificial intelligence in industry 4.0: {A} survey on what, how, and where.
  {IEEE} Trans. Ind. Informatics  \textbf{18}(8),  5031--5042 (2022)

\bibitem{DBLP:journals/inffus/ArrietaRSBTBGGM20}
Arrieta, A.B., Rodr{\'{\i}}guez, N.D., Ser, J.D., Bennetot, A., Tabik, S.,
  Barbado, A., Garc{\'{\i}}a, S., Gil{-}Lopez, S., Molina, D., Benjamins, R.,
  Chatila, R., Herrera, F.: Explainable artificial intelligence {(XAI):}
  concepts, taxonomies, opportunities and challenges toward responsible {AI}.
  Inf. Fusion  \textbf{58},  82--115 (2020)

\bibitem{DBLP:journals/corr/abs-2211-13968}
Bao, T., Chen, J., Li, W., Wang, X., Fei, J., Wu, L., Zhao, R., Zheng, Y.:
  {MIAD:} {A} maintenance inspection dataset for unsupervised anomaly
  detection. CoRR  \textbf{abs/2211.13968} (2022)

\bibitem{becker2021conceptual}
Becker, F., Bibow, P., Dalibor, M., Gannouni, A., Hahn, V., Hopmann, C., Jarke,
  M., Koren, I., Kr{\"o}ger, M., Lipp, J., et~al.: A conceptual model for
  digital shadows in industry and its application. In: Conceptual Modeling:
  40th International Conference, ER 2021, Virtual Event, October 18--21, 2021,
  Proceedings 40. pp. 271--281. Springer (2021)

\bibitem{DBLP:conf/caise/BibowDHMRSSW20}
Bibow, P., Dalibor, M., Hopmann, C., Mainz, B., Rumpe, B., Schmalzing, D.,
  Schmitz, M., Wortmann, A.: Model-driven development of a digital twin for
  injection molding. In: Dustdar, S., Yu, E., Salinesi, C., Rieu, D., Pant, V.
  (eds.) Advanced Information Systems Engineering - 32nd International
  Conference, CAiSE 2020, Proceedings. LNCS, vol. 12127, pp. 85--100. Springer
  (2020)

\bibitem{Brillowski:848836}
Brillowski, F.S., Becker, M., Vermeulen, M., Trauth, D., Bergs, T., Greb, C.,
  Gries, T.: {E}xplainable {AI} for error detection in composites : knowledge
  discovery in artificial neural networks. In: SAMPE EUROPE Conference and
  Exhibition 2021. SAMPE EUROPE Conference and Exhibition, Baden/Zürich
  (Switzerland), 29 Oct 2021 - 30 Oct 2021 (Oct 2021),
  \url{https://publications.rwth-aachen.de/record/848836}

\bibitem{DBLP:journals/corr/abs-2102-11848}
Brito, L.C., Susto, G.A., Brito, J.N., Duarte, M.A.V.: An explainable
  artificial intelligence approach for unsupervised fault detection and
  diagnosis in rotating machinery. CoRR  \textbf{abs/2102.11848} (2021)

\bibitem{DBLP:journals/jair/BurkartH21}
Burkart, N., Huber, M.F.: A survey on the explainability of supervised machine
  learning. J. Artif. Intell. Res.  \textbf{70},  245--317 (2021)

\bibitem{chowdhury2022xai}
Chowdhury, D., Sinha, A., Das, D.: Xai-3dp: Diagnosis and understanding faults
  of 3-d printer with explainable ensemble ai. IEEE Sensors Letters
  \textbf{7}(1), ~1--4 (2022)

\bibitem{DBLP:journals/corr/abs-2006-11371}
Das, A., Rad, P.: Opportunities and challenges in explainable artificial
  intelligence {(XAI):} {A} survey. CoRR  \textbf{abs/2006.11371} (2020)

\bibitem{deitsch2019automatic}
Deitsch, S., Christlein, V., Berger, S., Buerhop-Lutz, C., Maier, A., Gallwitz,
  F., Riess, C.: Automatic classification of defective photovoltaic module
  cells in electroluminescence images. Solar Energy  \textbf{185},  455--468
  (2019)

\bibitem{DINDeutschesInstitutfurNormierunge.V..1987}
{DIN Deutsches Institut f{\"u}r Normierung e.V.}: DIN~65147: Kohlenstoffasern
  Gewebe aus Kohlenstofffilamentgarn. {beuth Verlag}, Berlin (1987)

\bibitem{DINDeutschesInstitutfurNormierunge.V..1999}
{DIN Deutsches Institut f{\"u}r Normierung e.V.}: DIN 65673: Luft- und
  Raumfahrt Faserverst{\"a}rkte Kunststoffe. {beuth Verlag}, Berlin (1999)

\bibitem{DBLP:journals/ijinfoman/DuanED19}
Duan, Y., Edwards, J.S., Dwivedi, Y.K.: Artificial intelligence for decision
  making in the era of big data - evolution, challenges and research agenda.
  Int. J. Inf. Manag.  \textbf{48},  63--71 (2019)

\bibitem{Duboust.2017}
Duboust, N., Ghadbeigi, H., Pinna, C., Ayvar-Soberanis, S., Collis, A., Scaife,
  R., Kerrigan, K.: An optical method for measuring surface roughness of
  machined carbon fibre-reinforced plastic composites. Journal of Composite
  Materials  \textbf{51}(3),  289--302 (2017)

\bibitem{gamble2021determining}
Gamble, P., Jaroensri, R., Wang, H., Tan, F., Moran, M., Brown, T.,
  Flament-Auvigne, I., Rakha, E.A., Toss, M., Dabbs, D.J., et~al.: Determining
  breast cancer biomarker status and associated morphological features using
  deep learning. Communications medicine  \textbf{1}(1), ~14 (2021)

\bibitem{DBLP:journals/natmi/GeirhosJMZBBW20}
Geirhos, R., Jacobsen, J., Michaelis, C., Zemel, R.S., Brendel, W., Bethge, M.,
  Wichmann, F.A.: Shortcut learning in deep neural networks. Nat. Mach. Intell.
   \textbf{2}(11),  665--673 (2020)

\bibitem{Gholizadeh.2016}
Gholizadeh, S.: A review of non-destructive testing methods of composite
  materials. Procedia Structural Integrity  \textbf{1}(2),  50--57 (2016)

\bibitem{DBLP:conf/nips/GhorbaniWZK19}
Ghorbani, A., Wexler, J., Zou, J.Y., Kim, B.: Towards automatic concept-based
  explanations. In: Wallach, H.M., Larochelle, H., Beygelzimer, A.,
  d'Alch{\'{e}}{-}Buc, F., Fox, E.B., Garnett, R. (eds.) NeurIPS. pp.
  9273--9282 (2019)

\bibitem{DBLP:journals/corr/abs-1907-07165}
Goyal, Y., Shalit, U., Kim, B.: Explaining classifiers with causal concept
  effect (cace). CoRR  \textbf{abs/1907.07165} (2019)

\bibitem{DBLP:journals/corr/abs-1904-04520}
Graziani, M., Andrearczyk, V., M{\"{u}}ller, H.: Regression concept vectors for
  bidirectional explanations in histopathology. CoRR  \textbf{abs/1904.04520}
  (2019)

\bibitem{DBLP:conf/ickii/HongLLKH20}
Hong, C.W., Lee, C., Lee, K., Ko, M., Hur, K.: Explainable artificial
  intelligence for the remaining useful life prognosis of the turbofan engines.
  In: ICKII. pp. 144--147. {IEEE} (2020)

\bibitem{DBLP:conf/cvpr/HuangLMW17}
Huang, G., Liu, Z., van~der Maaten, L., Weinberger, K.Q.: Densely connected
  convolutional networks. In: CVPR. pp. 2261--2269. {IEEE} Computer Society
  (2017)

\bibitem{islam2022systematic}
Islam, M.R., Ahmed, M.U., Barua, S., Begum, S.: A systematic review of
  explainable artificial intelligence in terms of different application domains
  and tasks. Applied Sciences  \textbf{12}(3), ~1353 (2022)

\bibitem{DBLP:conf/ijcnn/KamakshiGK21}
Kamakshi, V., Gupta, U., Krishnan, N.C.: {PACE:} posthoc architecture-agnostic
  concept extractor for explaining cnns. In: IJCNN. pp.~1--8. {IEEE} (2021)

\bibitem{DBLP:conf/icml/KimWGCWVS18}
Kim, B., Wattenberg, M., Gilmer, J., Cai, C.J., Wexler, J., Vi{\'{e}}gas, F.B.,
  Sayres, R.: Interpretability beyond feature attribution: Quantitative testing
  with concept activation vectors {(TCAV)}. In: Dy, J.G., Krause, A. (eds.)
  ICML. Proceedings of Machine Learning Research, vol.~80, pp. 2673--2682.
  {PMLR} (2018)

\bibitem{DBLP:journals/tai/KumarSGKK21}
Kumar, A., Sehgal, K., Garg, P., Kamakshi, V., Krishnan, N.C.: {MACE:} model
  agnostic concept extractor for explaining image classification networks.
  {IEEE} Trans. Artif. Intell.  \textbf{2}(6),  574--583 (2021)

\bibitem{li2017visible}
Li, X., Yang, Q., Chen, Z., Luo, X., Yan, W.: Visible defects detection based
  on uav-based inspection in large-scale photovoltaic systems. IET Renewable
  Power Generation  \textbf{11}(10),  1234--1244 (2017)

\bibitem{DBLP:conf/nips/LundbergL17}
Lundberg, S.M., Lee, S.: A unified approach to interpreting model predictions.
  In: Guyon, I., von Luxburg, U., Bengio, S., Wallach, H.M., Fergus, R.,
  Vishwanathan, S.V.N., Garnett, R. (eds.) NeurIPS. pp. 4765--4774 (2017)

\bibitem{DBLP:journals/sensors/MeasMKTLLPB22}
Meas, M., Machlev, R., Kose, A., Tepljakov, A., Loo, L., Levron, Y., Petlenkov,
  E., Belikov, J.: Explainability and transparency of classifiers for
  air-handling unit faults using explainable artificial intelligence {(XAI)}.
  Sensors  \textbf{22}(17), ~6338 (2022)

\bibitem{Mueller.2022}
Mueller, K., Greb, C.: Machine vision: Error detection and classification of
  tailored textiles using neural networks. In: Andersen, A.L., Andersen, R.,
  Brunoe, T.D., Larsen, M.S.S., Nielsen, K., Napoleone, A., Kjeldgaard, S.
  (eds.) Towards Sustainable Customization: Bridging Smart Products and
  Manufacturing Systems, pp. 595--602. Lecture Notes in Mechanical Engineering,
  {Springer International Publishing}, Cham (2022)

\bibitem{DBLP:journals/corr/abs-2206-04531}
Posada{-}Moreno, A.F., Surya, N., Trimpe, S.: {ECLAD:} extracting concepts with
  local aggregated descriptors. CoRR  \textbf{abs/2206.04531} (2022)

\bibitem{DBLP:conf/kdd/Ribeiro0G16}
Ribeiro, M.T., Singh, S., Guestrin, C.: "why should {I} trust you?": Explaining
  the predictions of any classifier. In: Krishnapuram, B., Shah, M., Smola,
  A.J., Aggarwal, C.C., Shen, D., Rastogi, R. (eds.) SIGKDD. pp. 1135--1144.
  {ACM} (2016)

\bibitem{saranya2023systematic}
Saranya, A., Subhashini, R.: A systematic review of explainable artificial
  intelligence models and applications: Recent developments and future trends.
  Decision Analytics Journal p. 100230 (2023)

\bibitem{DBLP:conf/icmla/MouchawehR22}
{Sayed Mouchaweh}, M., Rajaoarisoa, L.H.: Explainable decision support tool for
  iot predictive maintenance within the context of industry 4.0. In: Wani,
  M.A., Kantardzic, M.M., Palade, V., Neagu, D., Yang, L., Chan, K.Y. (eds.)
  ICMLA. pp. 1492--1497. {IEEE} (2022)

\bibitem{DBLP:conf/www/Sculley10}
Sculley, D.: Web-scale k-means clustering. In: Rappa, M., Jones, P., Freire,
  J., Chakrabarti, S. (eds.) Proceedings of the 19th International Conference
  on World Wide Web, {WWW} 2010, Raleigh, North Carolina, USA, April 26-30,
  2010. pp. 1177--1178. {ACM} (2010)

\bibitem{DBLP:conf/iccv/SelvarajuCDVPB17}
Selvaraju, R.R., Cogswell, M., Das, A., Vedantam, R., Parikh, D., Batra, D.:
  Grad-cam: Visual explanations from deep networks via gradient-based
  localization. In: ICCV. pp. 618--626. {IEEE} Computer Society (2017)

\bibitem{DBLP:journals/mansci/SenonerNF22}
Senoner, J., Netland, T.H., Feuerriegel, S.: Using explainable artificial
  intelligence to improve process quality: Evidence from semiconductor
  manufacturing. Manag. Sci.  \textbf{68}(8),  5704--5723 (2022)

\bibitem{DBLP:conf/fuzzIEEE/SerradillaZCAOZ20}
Serradilla, O., Zugasti, E., Cernuda, C., Aranburu, A., de~Okariz, J.R.,
  Zurutuza, U.: Interpreting remaining useful life estimations combining
  explainable artificial intelligence and domain knowledge in industrial
  machinery. In: FUZZ-IEEE. pp.~1--8. {IEEE} (2020)

\bibitem{DBLP:journals/kais/StrumbeljK14}
Strumbelj, E., Kononenko, I.: Explaining prediction models and individual
  predictions with feature contributions. Knowl. Inf. Syst.  \textbf{41}(3),
  647--665 (2014)

\bibitem{DBLP:journals/access/SunHTLJL20}
Sun, K.H., Huh, H., Tama, B.A., Lee, S.Y., Jung, J.H., Lee, S.: Vision-based
  fault diagnostics using explainable deep learning with class activation maps.
  {IEEE} Access  \textbf{8},  129169--129179 (2020)

\bibitem{Uthemann.2017}
Uthemann, C., Jacobsen, L., Gries, T.: Cost efficiency through load-optimised
  and semi-impregnated prepregs. Lightweight Design worldwide  \textbf{10}(6),
  18--21 (2017)

\bibitem{DBLP:journals/candie/WangLWT21}
Wang, J., Lim, M.K., Wang, C., Tseng, M.: The evolution of the internet of
  things (iot) over the past 20 years. Comput. Ind. Eng.  \textbf{155},  107174
  (2021)

\bibitem{Witten.2022}
Witten, E., Mathes, V.: Der europ{\"a}ische markt f{\"u}r faserverst{\"a}rkte
  kunststoffe / composites 2021: Marktentwicklungen, trends, herausforderungen
  und ausblicke (2022),
  \url{www.avk-tv.de/files/20220503\_avk\_marktbericht\_2022\_final.pdf}

\bibitem{DBLP:conf/nips/YehKALPR20}
Yeh, C., Kim, B., Arik, S.{\"{O}}., Li, C., Pfister, T., Ravikumar, P.: On
  completeness-aware concept-based explanations in deep neural networks. In:
  Larochelle, H., Ranzato, M., Hadsell, R., Balcan, M., Lin, H. (eds.) NeurIPS
  (2020)

\bibitem{DBLP:journals/corr/abs-2110-14297}
Yona, G., Greenfeld, D.: Revisiting sanity checks for saliency maps. CoRR
  \textbf{abs/2110.14297} (2021)

\bibitem{DBLP:journals/access/ZhangHDYT22}
Zhang, Z., Hamadi, H.M.N.A., Damiani, E., Yeun, C.Y., Taher, F.: Explainable
  artificial intelligence applications in cyber security: State-of-the-art in
  research. {IEEE} Access  \textbf{10},  93104--93139 (2022)

\bibitem{DBLP:conf/cvpr/ZhouKLOT16}
Zhou, B., Khosla, A., Lapedriza, {\`{A}}., Oliva, A., Torralba, A.: Learning
  deep features for discriminative localization. In: CVPR. pp. 2921--2929.
  {IEEE} Computer Society (2016)

\end{thebibliography}
\end{document}